%% file: main.tex
\documentclass{article}
\PassOptionsToPackage{numbers, compress}{natbib}

\usepackage[preprint]{neurips_2021}

\usepackage[utf8]{inputenc} 
\usepackage[T1]{fontenc}    
\usepackage{hyperref}       
\usepackage{url}            
\usepackage{booktabs}       
\usepackage{amsfonts}       
\usepackage{nicefrac}       
\usepackage{microtype}      
\usepackage{xcolor}         
\usepackage{natbib}
\usepackage{amssymb}
\usepackage{pifont}

\usepackage{microtype}
\usepackage{graphicx}
\usepackage{subfigure}
\usepackage{bbm}
\usepackage{booktabs} 
\usepackage{amsmath,amsfonts,amsthm,amssymb}
\usepackage{mathrsfs}

\newtheorem{theorem}{Theorem}
\newtheorem{corollary}{Corollary}

\newtheorem{proposition}{Proposition}

\newtheorem{remark}{Remark}
\usepackage{setspace}
\usepackage{graphicx}
\usepackage{subfigure}
\usepackage{hyperref}
\newcommand*{\Scale}[2][4]{\scalebox{#1}{$#2$}}%

\usepackage{xcolor}
\newcount\comments  
\newcommand{\genComment}[2]{\ifnum\comments=1{\textcolor{#1}{\textsf{\footnotesize #2}}}\fi}

\usepackage{wrapfig}
\usepackage{setspace}
\usepackage{graphicx}
\usepackage{subfigure}
\usepackage{booktabs} 

\usepackage{hyperref}
\usepackage{algorithmic}
\usepackage{algorithm}

\title{Safe Exploration by Solving Early Terminated MDP}

\author{%
Hao Sun\textsuperscript{1}\thanks{sh018@ie.cuhk.edu.hk},
Ziping Xu\textsuperscript{2},
Meng Fang\textsuperscript{3},
Zhenghao Peng\textsuperscript{1},
Jiadong Guo\textsuperscript{4},
Bo Dai\textsuperscript{5},
Bolei Zhou\textsuperscript{1} \\
\textsuperscript{1}CUHK,
\textsuperscript{2}University of Michigan,
\textsuperscript{3}Tencent,
\textsuperscript{3}HKUST,
\textsuperscript{5}NTU
}
\begin{document}

\maketitle

\input{tex/abstract}
\input{tex/introduction}
\input{tex/preliminaries}

\input{tex/relatedwork}
\input{tex/method}

\input{tex/experiment}

\input{tex/conclusion}

\newpage
\bibliography{example_paper}
\bibliographystyle{icml2021}
\newpage
\input{tex/appendix}


\end{document}

%% file: tex/abstract.tex
\begin{abstract}
Safe exploration is crucial for the real-world application of reinforcement learning (RL). Previous works consider the safe exploration problem as Constrained Markov Decision Process (CMDP), where the policies are being optimized under constraints. However, when encountering any potential dangers, human tends to stop immediately and rarely learns to behave safely in danger. Motivated by human learning, we introduce a new approach to address safe RL problems under the framework of Early Terminated MDP (ET-MDP). We first define the ET-MDP as an unconstrained MDP with the same optimal value function as its corresponding CMDP. An off-policy algorithm based on context models is then proposed to solve the ET-MDP, which thereby solves the corresponding CMDP with better asymptotic performance and improved learning efficiency. Experiments on various CMDP tasks show a substantial improvement over previous methods that directly solve CMDP.
\end{abstract}

%% file: tex/introduction.tex
\section{Introduction}
\label{submission}

While reinforcement learning (RL) achieves great success in solving challenging decision making problems, several critical issues need to be addressed before it can be adopted in real-world applications. RL Safety, including safe optimization and safe exploration, is one of them. The learning paradigm of RL is composed of exploration and exploitation with experiences from trial-and-error~\cite{sutton1998introduction}. Thus the RL agents need to attempt a wide range of states and actions to better estimate their values, some of which are harmful that may lead to major damage. 

To tackle the RL safety problem, \citet{altman1999constrained} defines the Constrained Markov Decision Processes (CMDPs), where the policy optimization of standard RL algorithms should be executed in a constraint-satisfied policy class.
Many deep RL approaches for CMDPs are proposed after the rising of deep neural network function approximators: those works mainly focus on the optimization of the CMDP tasks, $i.e.$, how to effectively convert a CMDP task into a solvable form. \citet{achiam2017constrained} extend the trust region methods~\cite{schulman2015trust} into the context of CMDPs and guarantees the monotonicity of policy improvement; the Lagrangian methods, barrier (interior point) methods used in normal constrained optimization tasks and Lyapunov methods are extended to solve the CMDPs based on their MDP counterparts in ~\citet{chow2017risk, taylor2020learning, liu2020ipo,cheng2019end, perkins2002lyapunov,chow2018lyapunov, sikchi2020lyapunov}; another approach is based on safety-critic, where an additional critic is learned beside the primal critic for rewards to predict the cost of possible behaviors~\cite{zhang2020cautious,bharadhwaj2020conservative,srinivasan2020learning}.


Almost all those previous works on CMDP are derived from on-policy methods except for ~\citet{srinivasan2020learning}, which leverages an off-policy critic for the reward-related critic. Different from the previous works, our proposed safe-RL algorithm depends purely on off-policy algorithms which are known to achieve better sample efficiency.
It is worth mentioning that although we mainly focus on off-policy methods in this work, in principle, the proposed framework can be also combined with other on-policy algorithms to solve safe-RL tasks.

In this work, we provide both theoretical analyses and empirical experiments to show that CMDPs can be efficiently solved through their early terminated counterparts, namely ET-MDPs, which terminate an episode whenever the constraints are violated. We first show under deterministic and tabular MDPs, early termination can filter out invalid episodes and improve sample complexity. We go further to explore if the same improvement holds for more complex cases in practice. 

The challenge is that it is not suitable to directly apply the conventional RL algorithms like TD3~\cite{fujimoto2018addressing} to the ET-MDP tasks. The issue lies on the problem of \textbf{limited state visitation} during learning. Intuitively, if we terminate a training episode whenever the agent violates the constraints, the agent's behavior will be limited in a relatively small state space compared to the exploration without the constraints. Thereby, the learning efficiency will be limited~\cite{agarwal2019theory}. In order to solve the limited state visitation problem in ET-MDP, we adopt the 
idea of context models, previously introduced in Meta-RL literature ~\cite{fakoor2019meta,rakelly2019efficient} to improve the generality of policies across different training tasks. In our setting of ET-MDP,  a context variable is learned to improve the generality of the learned policy over different states, thus it enables our policy to perform safely over different states within one task.


We evaluate our method on a range of CMDP environments, including both the deterministic and the stochastic environments with different types of constraints. These environments are the diagnostic 2D-Maze navigation task with different levels, stochastic navigation environments from the Safety-Gym~\cite{safety_gym_Ray2019}, the PointGather~\cite{achiam2017constrained}, and MuJoCo Locomotion environment with constraints~\cite{1606.01540}. 
The proposed method shows a remarkably improved performance in terms of both learning efficiency and asymptotic performance under constraints.


%% file: tex/preliminaries.tex
\section{Preliminaries}
\label{sect:prel}

\textbf{Constrained Markov Decision Process.} The standard formulation of Constrained RL is the Constrained Markov Decision Process (CMDP), where an agent interacts with the environment under certain constraints. Here we consider the deterministic CMDP with a fixed horizon $H\in\mathbb{N}^+$ denoted by a tuple $(\mathcal{S},\mathcal{A}, H,r,c,C,\mathcal{T})$, where $\mathcal{S}$ and $\mathcal{A}$ are the state and action space; $r, c:\mathcal{S}\times\mathcal{A} \to \mathbb{R}$ denote the reward function and cost function;
$C\in \mathbb{R}^+$ is the upper bound on the permitted expected cumulative cost; 
$\mathcal{T}: \mathcal{S} \times \mathcal{A} \mapsto \mathcal{S}$ denotes the transition function.
 
We use $\Pi$ to denote the stationary policy class, where $\Pi = \{\pi: \mathcal{S}\times\mathcal{A}\to[0,1],\sum_a\pi(a|s) = 1\}$.
An algorithm for CMDP is to find $\pi^*\in\Pi$ as the result of the following optimization problem,
\begin{equation}
\label{cmdp}
\begin{array}{ll}
  \max_{\pi\in\Pi} \mathbb{E}_{\tau\sim\pi, \mathcal{T}}[\sum_{t=1}^H  r_t], 
   \quad \text{s.t.} \quad \mathbb{E}_{\tau\sim\pi,\mathcal{T}}[\sum_{t=1}^H c_t] \le C,
\end{array}
\end{equation}
where the expectation is taken over the trajectory $\tau = (s_1, a_1, r_1, \dots, s_H, a_H, r_H)$ generated by policy $\pi$ under the environment $\mathcal{T}$.

\textbf{Lagrangian Method.} The Lagrangian method relaxes the problem Eqn.(\ref{cmdp}) to an unconstrained optimization problem with a penalty term
\begin{equation}
\label{eq:lagrangian}
\pi^{*}=\max_{\pi\in\Pi}\min_{\lambda\ge0} \mathbb{E}_{\tau\sim\pi,\mathcal{T}}[\sum_{t=1}^H  r_t-\lambda c_t] + \lambda C ,
\end{equation}
where $\lambda\ge0$ is known as the Lagrangian multiplier. Suppose the policy $\pi$ is parameterized by $\theta$, i.e., $\pi = \pi_\theta$, the optimization over $\theta$ and $\lambda$ can be conducted iteratively through policy gradient ascent and stochastic gradient descent respectively according to Eqn.(\ref{eq:lagrangian}). \citet{chow2018lyapunov} points out that one of the possible defects of the Lagrangian methods is the violation of constraints during training, which is successfully solved by our proposed method.

\textbf{Constrained Policy Optimization.} \citet{achiam2017constrained} proposes the Constrained Policy Optimization (CPO), an analytical way to solve CMDP through trust region optimization. Specifically, CPO develops an approximation of Eqn.(\ref{cmdp}) by replacing the objective and constraints with surrogate functions~\citep{achiam2017constrained,schulman2015trust} and provides theoretical analysis on the worst case performance as well as constraint violation. In CPO, the policy is updated as:
\begin{equation}
\label{cpo}
\begin{array}{ll}
\pi_{k+1}=\arg\max_{\pi\in\Pi} \mathbb{E}[A_{r, 1}^{\pi_k}(s,a)], 
\quad \text{s.t.}  \widetilde{J}_c(\pi_k)\le C, \quad \bar{D}_{KL}(\pi||\pi_k)\le \delta,
\end{array}
\end{equation}
wherein $\widetilde{J}_c(\pi_k) = \mathbb{E}_{\tau\sim\pi_k, \mathcal{T}}[\sum_{t=1}^H c_t] +  \mathbb{E}_{s,a}[A_{c, 1}^{\pi_k}(s,a)]$, $k=0,1,...,K$. Here $A_{r, i}^{\pi_k}(s,a)$ and $A_{c, i}^{\pi_k}(s,a)$ denote the advantage functions of reward and cost at step $i$ respectively. CPO is closely-connected to the $\theta$-projection approach of \citet{chow2018lyapunov}. 
The close relationship between CPO and the family of trust region algorithms makes it difficult to implement and to extend to other existing RL algorithms. On contrast, our proposed approach is highly flexible and is easy to implement to a various algorithms in nature.

\textbf{Context Models for Meta-RL.} Meta-RL aims to learn a good inductive bias of policy that can be quickly generalized to previously unseen tasks. In the meta-training phase, several tasks $\mathcal{D}_{train} = \{D_{(k)}\}_{k=1}^K$ are sampled from a task distribution $\mathcal{D}_{meta}$. In the meta-testing phase, $\mathcal{D}_{test} = \{D_{(k)}\}_{k=K+1}^N$ are sampled from the same task distribution.

Although the meta-optimization approaches~\cite{finn2017model,nichol2018first} have been successfully applied to various image classification tasks, their performance is relatively limited in RL tasks~\cite{fakoor2019meta}. Recent advance of the context approach meta-RL~\cite{rakelly2019efficient} learns a latent representation of the task and construct a context model through recurrent networks~\cite{gers1999learning,cho2014learning}. In this work, we follow~\cite{fakoor2019meta} to use the simplest form of meta-training, i.e., the multi-task objective:
\begin{equation}
    \hat{\theta}_{meta} = \arg \max_\theta  \frac{1}{n} \sum_{k=1}^n \mathbb{E} [\ell^{(k)}(\theta)],
\end{equation}
where $\ell^{(k)}(\theta)$ denotes the objective evaluated on the $k$-th task $D_{(k)}$.

%% file: tex/relatedwork.tex
\section{Related Work}
Learning RL policy under safety constraints~\citep{garcia2015comprehensive,amodei2016concrete,safety_gym_Ray2019} becomes an important topic in RL community due to the safety concerns of RL in real-world applications. For example, ~\citet{richter2019open} applies RL to the simulated surgical robot. ~\citet{kendall2019learning} implements RL algorithm in the autonomous driving scenario. In those applications, the safety of the learned policy is critical and the policy should be optimized under some safety constraints.
The common practice for this problem is to involve human interventions ~\cite{saunders2018trial} or correction of the output action~\cite{dalal2018safe,van2020online} under uncertain conditions.



In previous works, ~\citet{saunders2018trial} proposes HIRL, a scheme for safe RL requiring extra manual efforts to intervene the agent when it produces actions that lead to catastrophic outcomes. ~\citet{dalal2018safe} equips the policy network with a safety layer that can modulate the output action as an absolute safe one. However, the linear layer is incompetent to capture the dynamics of complex environments and it requires pre-training, which brings extra computation and risks to constraint violations.
\citet{achiam2017constrained} proposes the Constrained Policy Optimization (CPO), which is an analytical solution to solve CMDP through trust region optimization. 

The close relationship between CPO and the family of trust region methods~\cite{schulman2015trust} makes it difficult to implement and extend to other existing RL algorithms. 
Our context-based ETMDP approach, on the contrary, is highly flexible and can be implemented on top of various algorithms such as PPO~\cite{schulman2017proximal}, TRPO~\cite{schulman2015trust} and TD3~\cite{fujimoto2018addressing}.
Another straightforward approach to the soft constraint problem is the Lagrangian method~\cite{safety_gym_Ray2019}, which relaxes the hard-constrained optimization problem to an unconstrained one with an auxiliary penalty term.
An interesting result reported in~\cite{safety_gym_Ray2019} is that the approximation errors in CPO prevent it from fully satisfying the constraint. In contrast, simple Lagrangian method can find constraint-satisfying policies that attain nontrivial returns.

Note that while previous works have discussed the effectiveness of applying early termination~\cite{hamalainen2019visualizing} and absorbing state~\cite{geibel2005risk} in constrained RL to further improve the performance of their proposed methods~\cite{wachi2020safe}, we show in our work such an early-terminated MDP approach can be formulated in a more principled way. We also verify that it is capable enough to work in isolation to solve constrained RL tasks, with proper learning algorithms we will introduce in the following section. 

%% file: tex/method.tex
\section{Method}
In this work, we propose to handle constraints in the most straightforward way: an early termination is triggered whenever the learning policy violates the constraints. 
Such an early termination is previously used as a trick to improve the sample efficiency of solving regular MDPs~\cite{wang2019benchmarking}:
terminating bad trajectories accelerates the learning process since the policy space to search is reduced and the time horizon is shortened. Moreover, we do not need to learn to proceed after violations as an \textit{ideal} policy should never break the constraints.

We first introduce two types of constraints in CMDPs in Section~\ref{method_classification}. We will show that the early-termination trick used in locomotion tasks is indeed an intuitive approach for solving CMDP with, what we call, loose constraints. In Section~\ref{sec_etmdp}, we define ET-MDP as the foundation of our proposed method. ET-MDP enables the algorithms previously designed for MDP to solve CMDP tasks. We provide discussion on some practical issues in Section~\ref{sec_prac_issues} and introduce our algorithm to solve ET-MDP efficiently in Section~\ref{sec_context_td3}.

\subsection{Constraint Types}
\label{method_classification}
To show the relationship between normal MDPs and CMDPs as well as better illustrate the inspiration that links CMDPs with early termination, we first unify CMDP and MDP formulation in the loose-constrained cases: 
MDPs can be regarded as loose-constrained CMDPs when the constraints do not change their optimal solution. Thus those CMDPs can be solved by the same policy trained from its early termination counterparts. We then extend similar idea to the other case where constraints are tight.
We start with the definition of the learning objective. If we denote a policy set satisfying the constraints $C$ as
\begin{equation}
\label{eq_constrained_class}
    \Pi^c = \{\text{any policy } \pi: \sum_{t=1}^H c(s_t,\pi(s_t))\le C \},
\end{equation}
then the learning objective of Eqn.(\ref{cmdp}) becomes $\max_{\pi\in\Pi^c} \mathbb{E}_{\tau\sim\pi, \mathcal{T}}[\sum_{t=1}^H  r_t]$. The two types of constraints differ in whether the optimal policy lies in the constrained policy class (Eqn.(\ref{eq_constrained_class})) or not.
\paragraph{Loose Constraints}
In model-free RL, early termination is often used as a default environment setting to accelerate learning~\cite{duan2016benchmarking,wang2019benchmarking}. In such problems, early termination is usually applied when the agent reaches some undesired state, e.g., the center of mass get lower than some certain threshold. We call this kind of constraints loose ones, because the solution of the CMDP is the same as the MDP without constraints:
\begin{equation}
    \pi^*=\arg\max_{\pi\in\Pi} \mathbb{E}_{\tau\sim\pi, \mathcal{T}}[\sum_{t=1}^H  r_t] \in \Pi^c.
\end{equation}
In such CMDPs, considering the constraints or not won't change the final policy as the optimal solution will learn to not break the constraints automatically. Those loose constraints are shown to be able to accelerate learning in~\cite{pham2018constrained}.
\paragraph{Tight Constraints}
In other cases such as navigation in a space with barriers or lava, the barriers or lava can be regarded as constraints and will clearly change the optimal solution to navigate to the goal point compared with the environment of an empty space where there is no constraint applied:
\begin{equation}
    \pi^*=\arg\max_{\pi\in\Pi} \mathbb{E}_{\tau\sim\pi, \mathcal{T}}[\sum_{t=1}^H  r_t] \notin \Pi^c.
\end{equation}
In such CMDPs, learning to solve the MDP without the constraints can not lead to a satisfying policy for the CMDP as feasible behaviors of the agent must take the constraints into consideration. 



Based on such insights, we investigate the approach to solve the CMDPs with their early-termination (ET) counterparts, namely the ET-MDPs. We show the major challenge may come from the limited state visitation problem, which will be further illustrated in detail in Section~\ref{sec_context_td3}.

\subsection{Early Terminated MDP (ET-MDP)}
\label{sec_etmdp}
For any CMDP $(\mathcal{S}, \mathcal{A}, H, r, c, C, \mathcal{T})$, its ET-MDP is defined as a new unconstrained MDP $(\mathcal{S} \cup \{s_e\}, \mathcal{A}, H, r', \mathcal{T}')$, where $s_e$ is the absorbing state after termination. Generally speaking, ET-MDP has a history-dependent transition dynamic. In order to have a regular MDP, one can introduce an extra dimension to state space recording the cumulative costs denoted by $b_t = \sum_{\tau = 1}^{t} c_{\tau}$. Though $b_t$ takes values from a large set, the transition dynamic that involves in $b_t$ is known to the agent:
$\mathcal{T}'(s, b, a) = \mathcal{T}(s, a) \mathbbm{1}(b \leq C) + \mathbbm{1}(s = s_e, b > C)$. The reward function becomes $r'(s, b, a) = r(s, a)\mathbbm{1}(b \leq C) + r_{e}\mathbbm{1}(b > C)$ for some $r_e \in \mathbb{R}$. Since we are searching for a policy for the original CMDP, we still consider policies that are stationary with respect to $b$, $i.e$. $\pi(s, b) \equiv \pi(s)$. 


\begin{proposition}
\label{prop_1}
For sufficient small $r_e$, the optimal policy of ET-MDP coincidences with $\pi^*$ of the original CMDP. (Proof is given by Appendix)
\end{proposition}
Proposition 1 indicates that CMDP can be solved with their ET-MDP correspondence as long as the termination reward $r_{e}$ is small enough, which can be easily implemented in practice.
Intuitively, as an early-terminated episode is shorter than the original one, one should be able to save samples by solving ET-MDP. In the following part, we show the benefits of solving CMDP through its ET-MDP.
Now that we consider a special case, where $c(s, a) = \mathbbm{1}(\mathcal{T}(s, a) \in \mathcal{S}_c)$ for some $\mathcal{S}_c \subset \mathcal{S}$. Here the violation is caused by the entrance to some invalid states. We also assume that $\mathcal{S}_c$ is an absorbing class. This is an important case we consider in our experiments, as exploring invalid space is unnecessary and early termination can save samples.

To fairly show the benefits, we introduce a performance measure called regret, the difference between the total rewards of the optimal policy and the rewards received by the running algorithm $\mathcal{L}$:
$$
    R_T(\mathcal{L}) = \sum_{k = 1}^{\lfloor T/H \rfloor} (V^c_{\pi^*} - V_{\pi_k}^c),
$$
where $V_{\pi}^c$ is the expected value function under policy $\pi$ and $\pi_k$ is the policy chosen for episode $k$. Regrets for deterministic MDPs can be lower and upper bounded. 
\begin{theorem}[Theorem 3 in \cite{wen2013efficient}]
Any reinforcement learning algorithm $\mathcal{L}$ that takes input a state space, action space, a horizon, there exists an MDP, such that the regret $\sup_T R_T(\mathcal{L}) \geq 2H|\mathcal{S}||\mathcal{A}|$. There exists an algorithm $\mathcal{L}$, such that for any MDP, the regret $\sup_T R_T(\mathcal{L}) \leq 2H|\mathcal{S}||\mathcal{A}|$.
\end{theorem}
The above lower bound applies to the CMDP as one can construct an MDP with a extreme loose constraint such that all the policies are valid. The above upper bound applies to our ET-MDP since in this special case, $c(s, a)$ is either $0$ or $1$ and the termination happens whenever a cost of $1$ is received, which makes it unnecessary to record the cumulative cost and our ET-MDP is a regular MDP with state space $(\mathcal{S} \setminus \mathcal{S}_c) \cup \{s_e\}$.
\begin{corollary}
\label{cor:1}
There exists a algorithm $\mathcal{L}_{ET}$ ET-MDP such that for any algorithm $\mathcal{L}_c$ for the original CMDP, the ratio 
$$
\frac{\sup_T R_T(\mathcal{L}_c)}{\sup_T R_T(\mathcal{L}_{ET})} \geq \frac{|\mathcal{S}|}{|\mathcal{S}| - |\mathcal{S}_c| + 1}.
$$
\end{corollary}

\begin{remark}[ET-MDPs reduce sample complexity]
\label{remark:2}
The above analysis ignored the fact that ET-MDP does not have to finish the whole $H$ steps for each episode, which means that when an algorithm is actually running, the above dependence on $H$ can be also decreased depending on the actual cutoffs.
\end{remark}

Corollary \ref{cor:1} shows that for tasks with a large invalid space, solving ET-MDP is more efficient. It provides the insight to apply similar methods to more complicated tasks like CMDPs for continuous control. As for the case where the termination trick is applied to the loose constrained tasks such as the Walker2d, Hopper and Humanoid in the MuJoCo Locomotion Suite~\cite{wang2019benchmarking}, we don't need to collect samples from infeasible regions. Thus we need to further investigate whether solving the ET-MDP can be a practically effective way to solve CMDPs.
\subsection{Practical Issues of ET-MDP}
\label{sec_prac_issues}
\subsubsection{Budget Tasks}
\label{sec_budget}
In general, there are two different empirical settings in CMDP. The first is the case where there is a \textit{budget} of behavior costs. Behaviors with some cost is not preferable but is permitted to some extent. Henceforth, to satisfy the constraints in Eqn.(\ref{cmdp}), the historical information of cumulative cost should be taken into consideration in making every-step decision. To achieve this, the primal state space $\mathcal{S}$ must be extended to $\mathcal{S}_{ext} = \mathcal{S}\oplus \mathcal{S}_{budget} \oplus \mathcal{S}_{time}$, where $\mathcal{S}_{budget}$ indicates the budget left in the episode, and $\mathcal{S}_{time}$ provides information on the number of time steps left in the episode~\cite{pardo2018time}.

Previous works under this setting include the Safety-Gym~\cite{safety_gym_Ray2019} and the PointGather environment~\cite{achiam2017constrained}, where the budget is a fixed positive integer that indicates how many times the agent can reach a certain type of states.

\subsubsection{Binary Tasks}

On the other hand, there are cases where the safety is considered to be extremely important that the constraints should never be broken in the deployment time of a learned policy. We call this kind of setting the binary CMDPs, where the binary indicates classifying a trajectory as safe or not safe. This is the relatively simple case and the constraints in Eqn.(\ref{cmdp}) can be simplified as $\sum_{t=0}^{\infty} c_t \le 0$, where $c_t = c > 0$ if the constraints are broken and $c_t = 0$ otherwise. Besides, no more effort is needed to ensure the decision making process Markovian.

For example, navigation in a space with lava belongs to such a setting. Another example is the MuJoCo Locomotion Suites, where a hopper, walker or humanoid simulator is required to move forward as fast as possible, and through the whole time the agent should never fall down. To sum up, this setting can be applied as long as there exists a solution that a task can be accomplished without stepping into any positive-cost regions.

\subsubsection{Empirical Tightened Approximation}
\label{sec_tightened_appx}
In this work, we propose to use a strict version of budget tasks: we show in experiments that considering a budget task as a binary task can be a practical approximation by not permitting the agent to reach any risky or costly regions. Although there can be problems in some certain environments where the CMDP can not be solved if the budget is too small, we will show in most of the standard benchmarks, such an approximated solution has good empirical performance in the next section. Analysis and example on the exception cases where such an implementation fails are provided in Appendix~\ref{counterexample}.

\begin{wrapfigure}{l}{8cm}
\centering
\includegraphics[width=0.5\columnwidth]{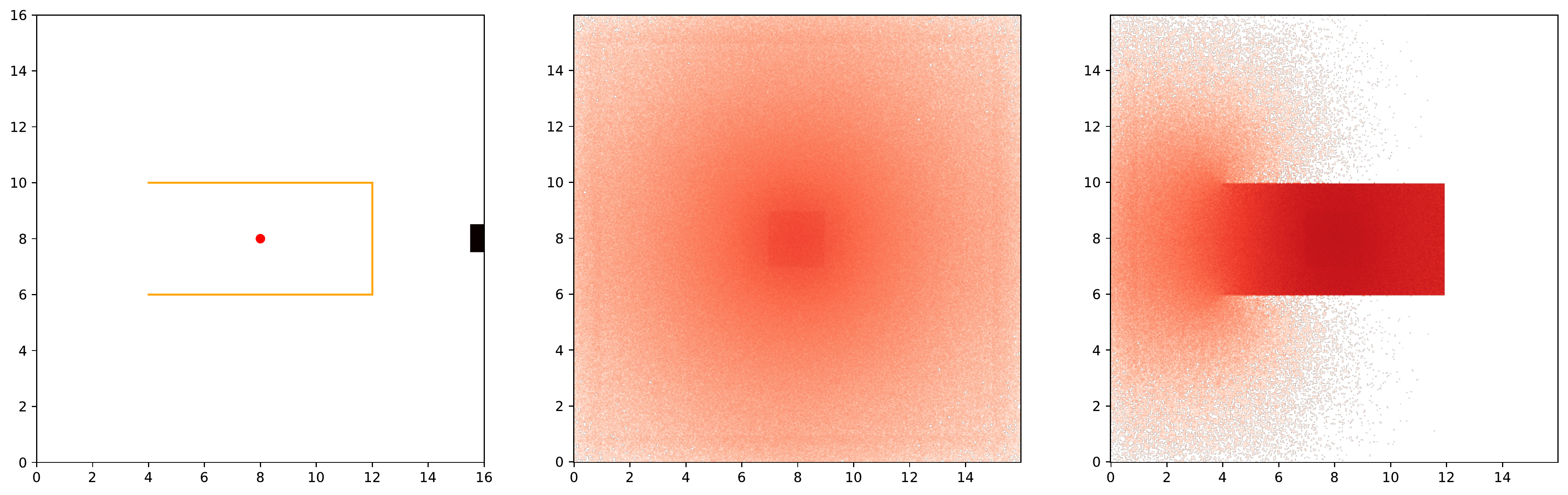}
\caption{The difference in state visitation frequency of MDPs and ET-MDPs in a diagnostic 2-D navigation environment. \textbf{Left}: the environment, an agent starts from the central red point in each episode and yellow lines denote lava, i.e., danger zone; \textbf{Middle}: the state visitation frequency of a random agent in a MDP by ignoring the lava; \textbf{Right}: the state visitation frequency of a random agent with lava in a ET-MDP. The limited state visitation in ET-MDP is the major challenge for existing RL algorithms.}
\label{fig_context_illu}
\vskip -0.2in
\end{wrapfigure}

\subsection{Solving ET-MDP with Context Models}
\label{sec_context_td3}

In the previous section, we have shown CMDPs can be solved with their ET-MDP counterparts, and the next step is to find suitable solvers for the ET-MDP task. Intuitively, solving ET-MDP is similar to solving normal MDPs as there are no constraints that should be taken into consideration. Any prevailing algorithm can be applied as ET-MDP solver, such as TD3~\cite{fujimoto2018addressing}, SAC~\cite{haarnoja2018soft}, PPO~\cite{schulman2017proximal}, TRPO~\cite{schulman2015trust}, Evolution Strategies (ES)~\cite{salimans2017evolution} and so on.

However, since ET-MDP is different from normal MDPs as there are possibly lots of terminated states $\mathcal{S}_{end}$, algorithms designed for normal MDP tasks are easy to get trapped in limited states, leading to relatively low learning efficiency. Similar problem has been discussed and termed as distribution shift in ~\cite{agarwal2019theory}. 

Figure~\ref{fig_context_illu} illustrates the difference in state visitation frequency of normal MDP (middle) and ET-MDP (right) under random exploration in a 2-D navigation environment, where the central red point denotes the starting point and the constraints are shown as yellow boundaries in the left figure. As all of the constraint-violation states will lead to termination in ET-MDP, the generalization ability of the learned policy becomes extremely important. Intuitively, learning algorithms that can generalize better to previously unseen states will be more competent in such tasks.

To solve such a challenge, we propose to adopt context models proposed in Meta RL literature~\cite{fakoor2019meta}. 
While in previous work the context models concentrate on the generalization ability among a series of Meta RL tasks, we apply context models to solve a specified ET-MDP task and tackle the limited state visitation problem.  Context models in RL are shown to learn generalizable representations between \textit{tasks} in meta-RL. Thus we regard solving the ET-MDP task (e.g., avoid getting terminated and collect as many reward as possible) with different initial states as different tasks, the context models should be able to learn transferable representations over different \textit{initial states}, and generalize learned policies to previously unseen states to avoid being terminated. 

We use the Gated Recurrent Units~\cite{cho2014learning} to model context variables as generalizable representations to solve ET-MDPs. We follow~\cite{fakoor2019meta} to build the context model based on TD3~\cite{fujimoto2018addressing}. Here we use separated context networks in our work for training stability, $i.e.$, we use $\mathcal{C}_{w_a}$ for actor and $\mathcal{C}_{w_c}$ for critic, such that both the actor $\pi$ and critic $Q$ take an additional context variable as input:
\begin{align}
    \pi = & \pi(s,z_a),\\ 
    Q = & Q(s,a,z_c), 
\end{align}

where $z_a = \mathcal{C}_{w_a}(\mathcal{Z}'_L)$, $z_c = \mathcal{C}_{w_c}(\mathcal{Z}'_L)$ and $\mathcal{Z}'_L$ is the previous $L$ step historical transitions: $\mathcal{Z}'_L=\{s_{t-L}, a_{t-L}, r_{t-L}, ..., s_{t-1},a_{t-1},r_{t-1}\}$. If $t-L\le0$, we use zero state $\boldsymbol{0}_s$, zero action $\boldsymbol{0}_a$ and zero reward $\boldsymbol{0}_r$ instead.

The context models ($\mathcal{C}_{w_a}, \mathcal{C}_{w_c}$) are optimized through the gradient chain rule in the optimization of actor and critic networks, with the gradient of
\begin{align}
\label{eq_upd_cwa}
&\Scale[0.9]{\nabla_{w_a} Q_{w_1}(s,a,z_c)|_{a=\pi_\theta(s,z_a)}\nabla_{z_a} \pi_\theta (s,z_a)|_{z_a = \mathcal{C}_{w_a}(\mathcal{Z}'_L)}\nabla_{w_a}\mathcal{C}_{w_a}(\mathcal{Z}'_L)},\\
    &\nabla_{w_c} \textbf{TD}(Q(s,a,\mathcal{C}_{w_c}(\mathcal{Z}'_L)))
\end{align}
separately, where \textbf{TD} denotes the temporal difference error. Details of the proposed algorithm are provided in Algorithm~\ref{Algorithm2} in Appendix~\ref{detailed_algo}. In the next section we will demonstrate the superiority of the proposed methods in solving ET-MDPs. 

%% file: tex/experiment.tex
\section{Experiments}
\begin{figure}[t]
\vskip 0.2in
\begin{center}
\begin{minipage}[htbp]{1.0\linewidth}
			\centering
			\includegraphics[width=1.0\linewidth]{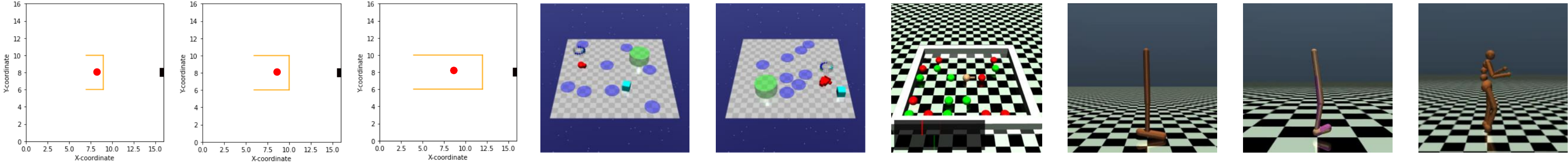}
		\end{minipage}%
\caption{Examples of the tested environments: The first three figures show the diagnostic 2D-Nav tasks with different constraint level; the following three figures show the budget tasks where agents control a point or a car to collect reward without hitting cost regions too many times; the last three figures show loose-constrained tasks where agents need to learn to move forward without falling.}
\label{fig_envs}
\end{center}
\vskip -0.2in
\end{figure}

We evaluate our proposed method on a diverse set of environments, including 1. loose constrained tasks (Hopper-Not-Fall, Walker-Not-Fall, Humanoid-Not-Fall), 2. static maze tasks with different levels, 3. stochastic navigation tasks (PointGoal1-v0, CarGoal1-v0), and 4. PointGather. The first two sets of environments are binary tasks while the other two sets of environments are budget tasks. Examples of environments are shown in Figure~\ref{fig_envs}.

We aim to validate the following claims in our experiments:
\begin{enumerate}
	\item CMDPs can be solved by solving their ET-MDP counterparts with the context-based TD3. For tight-constrained CMDPs, early termination can help improve learning efficiency (Section~\ref{exp_tight}), The tightened approximation in budget tasks can achieve satisfying empirical performance.
	\item While directly applying the standard RL algorithms like TD3 has the problem of limited state visitation, the context model mitigates the problem and improves the sample efficiency (Section~\ref{exp_state_visitation}).
	\item Context-based TD3 can further improve the performance on loose-constrained tasks (Section~\ref{exp_loose}).
\end{enumerate}

\begin{figure}[t]
\vskip 0.2in
\begin{center}
\begin{minipage}[htbp]{0.245\linewidth}
			\centering
			\includegraphics[width=1.0\linewidth]{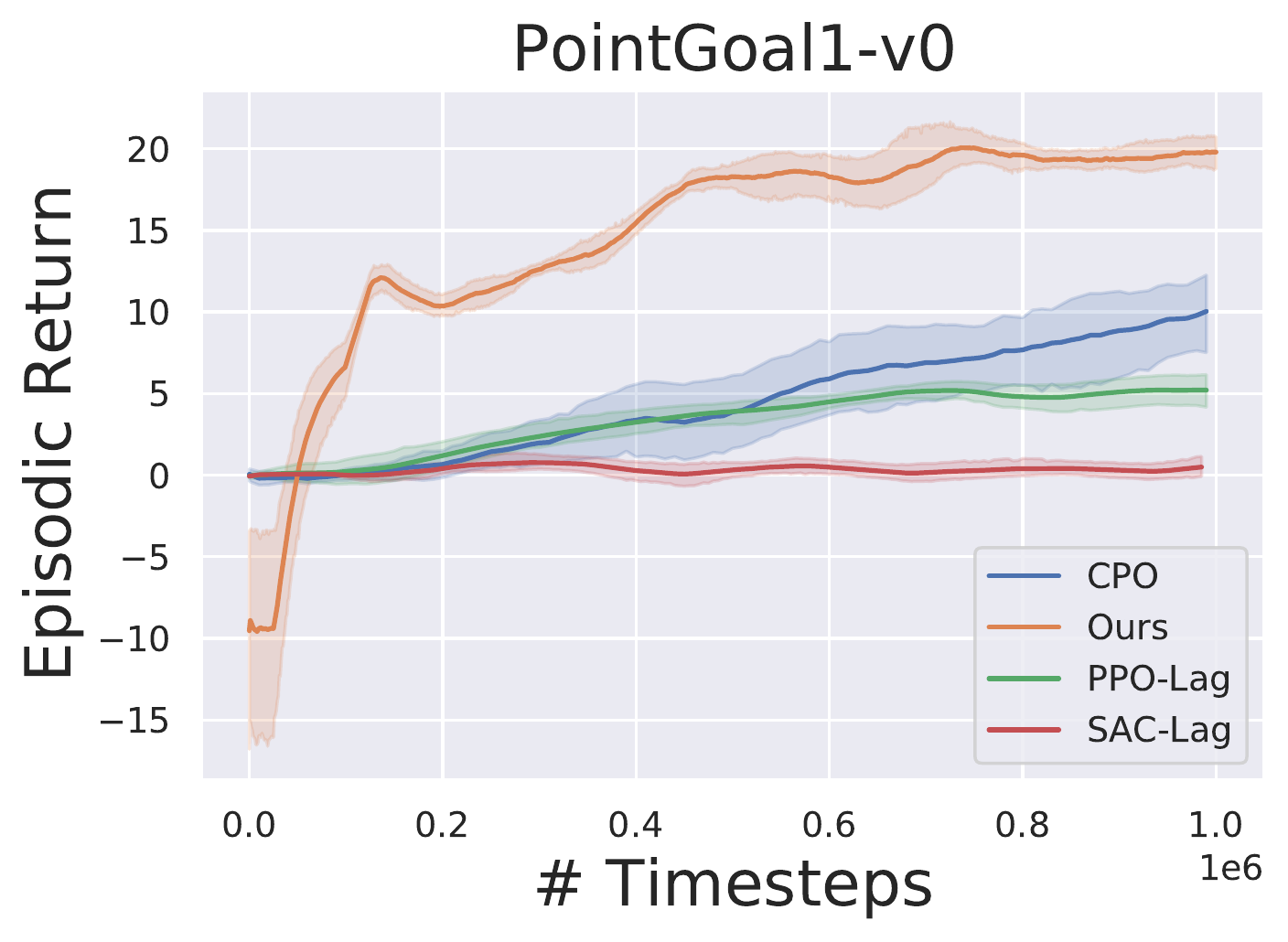}
		\end{minipage}%
		\begin{minipage}[htbp]{0.245\linewidth}
			\centering
			\includegraphics[width=1.0\linewidth]{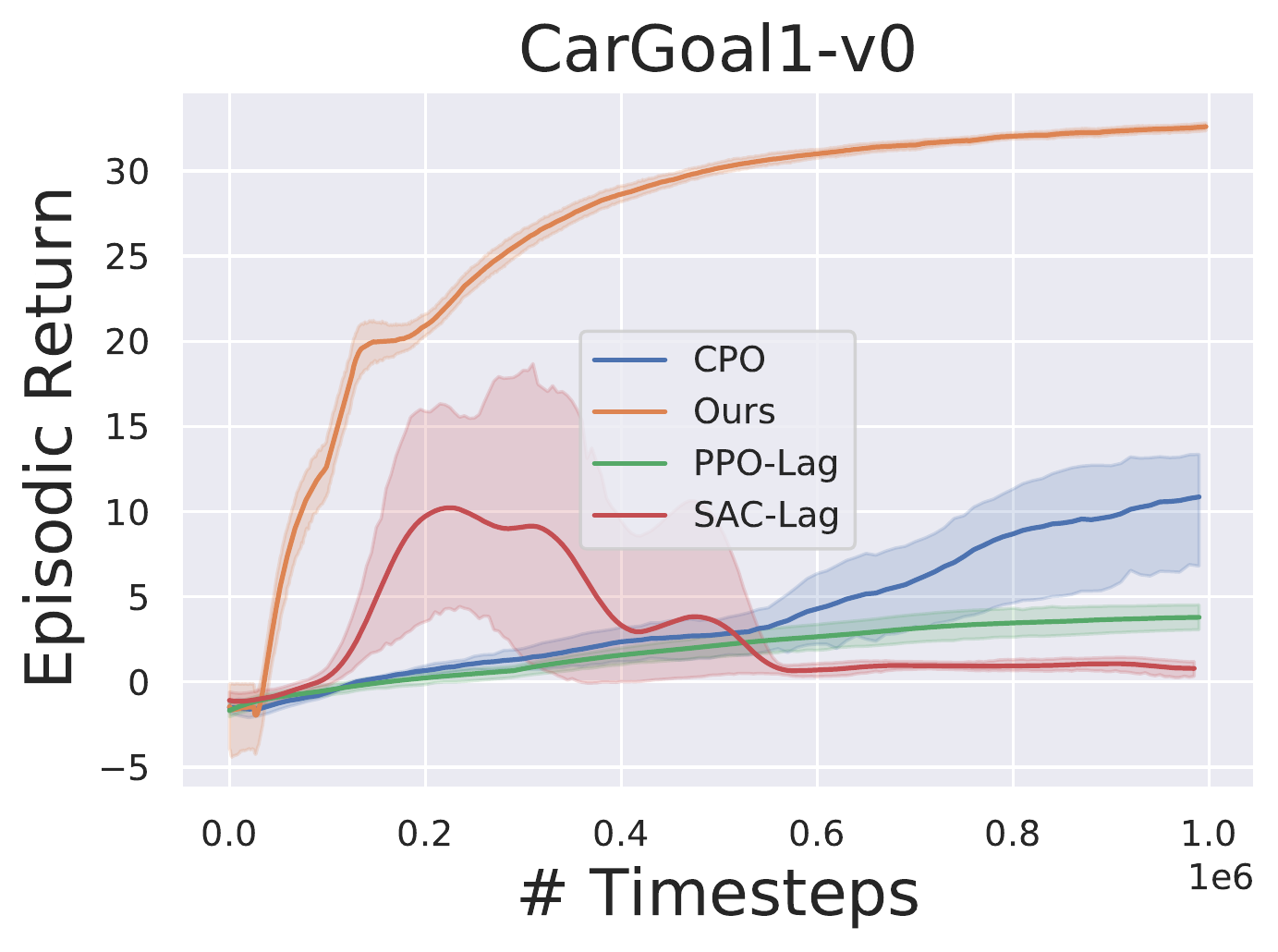}
		\end{minipage}%
		\begin{minipage}[htbp]{0.245\linewidth}
			\centering
			\includegraphics[width=1.0\linewidth]{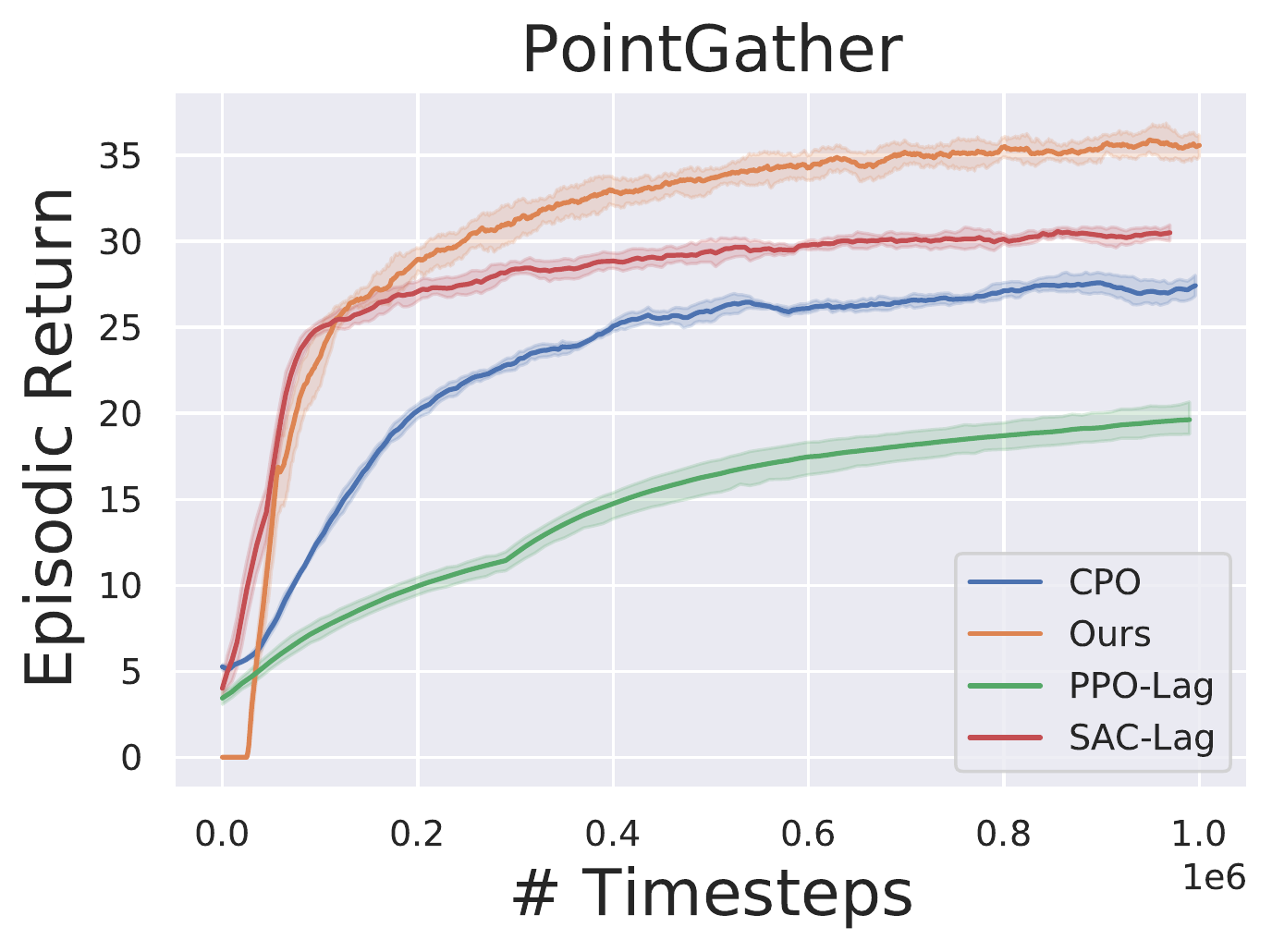}
		\end{minipage}%
		\begin{minipage}[htbp]{0.245\linewidth}
			\centering
			\includegraphics[width=1.0\linewidth]{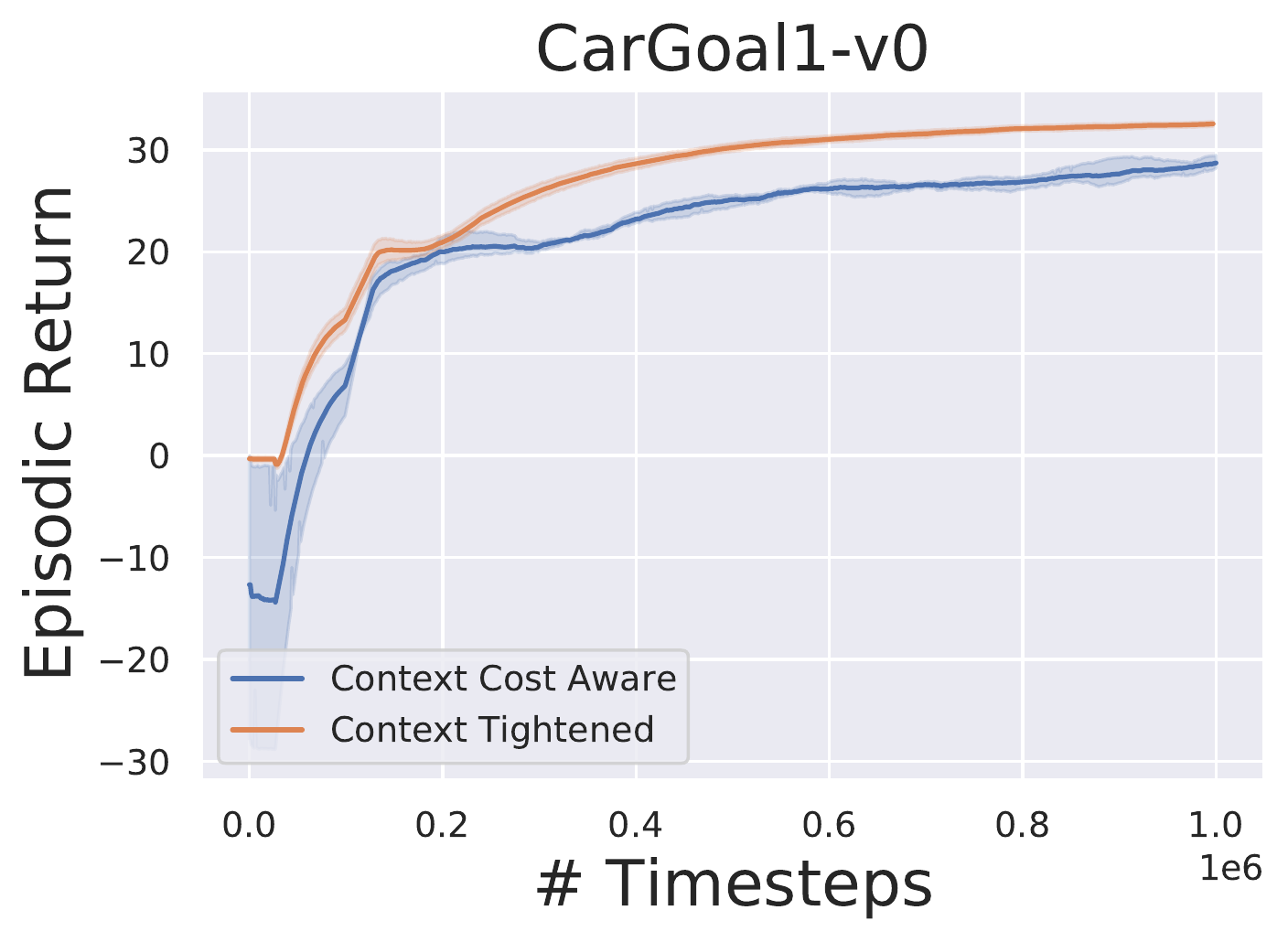}
		\end{minipage}\\%
		\begin{minipage}[htbp]{0.245\linewidth}
			\centering
			\includegraphics[width=1.0\linewidth]{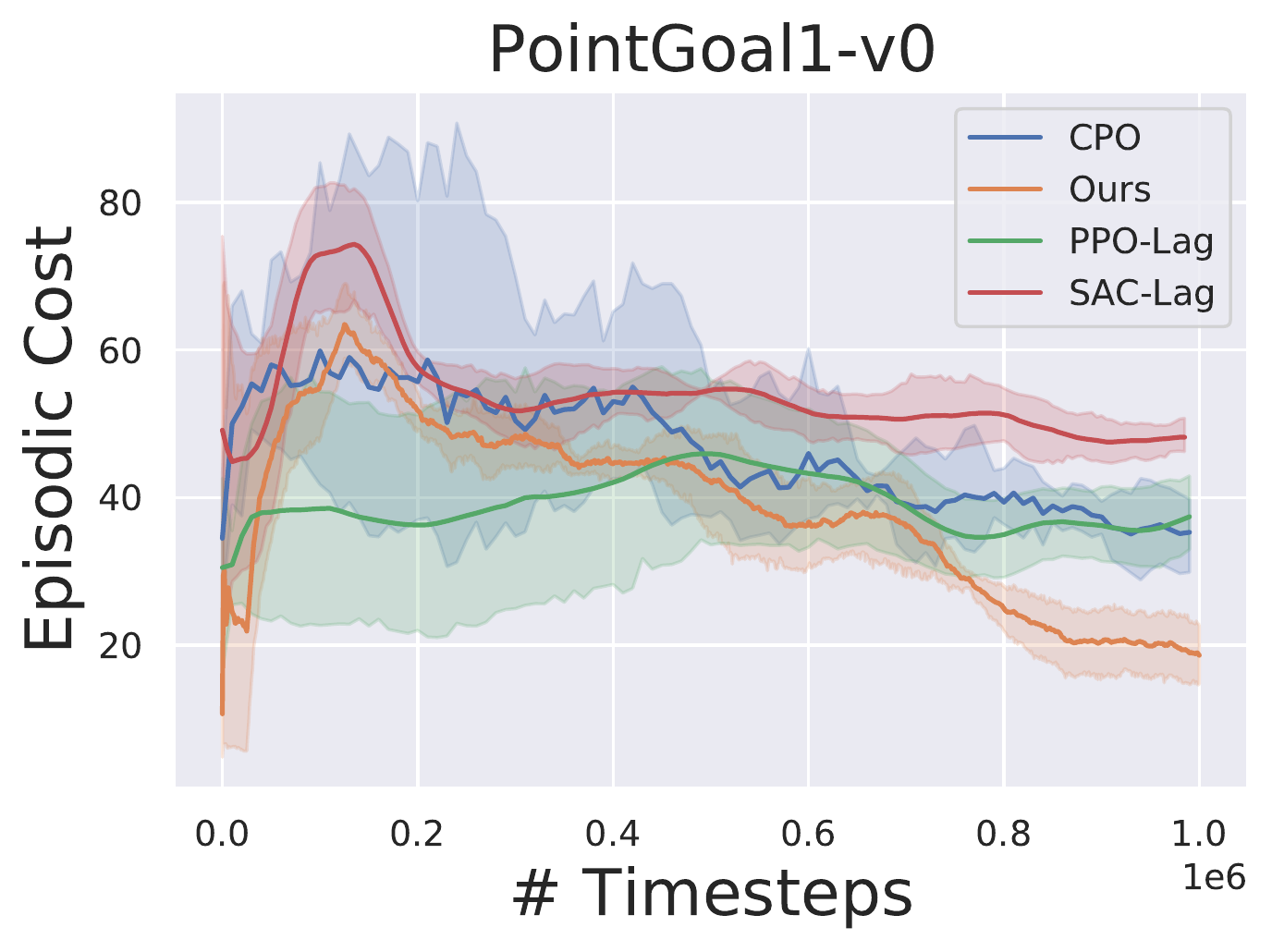}
		\end{minipage}%
		\begin{minipage}[htbp]{0.245\linewidth}
			\centering
			\includegraphics[width=1.0\linewidth]{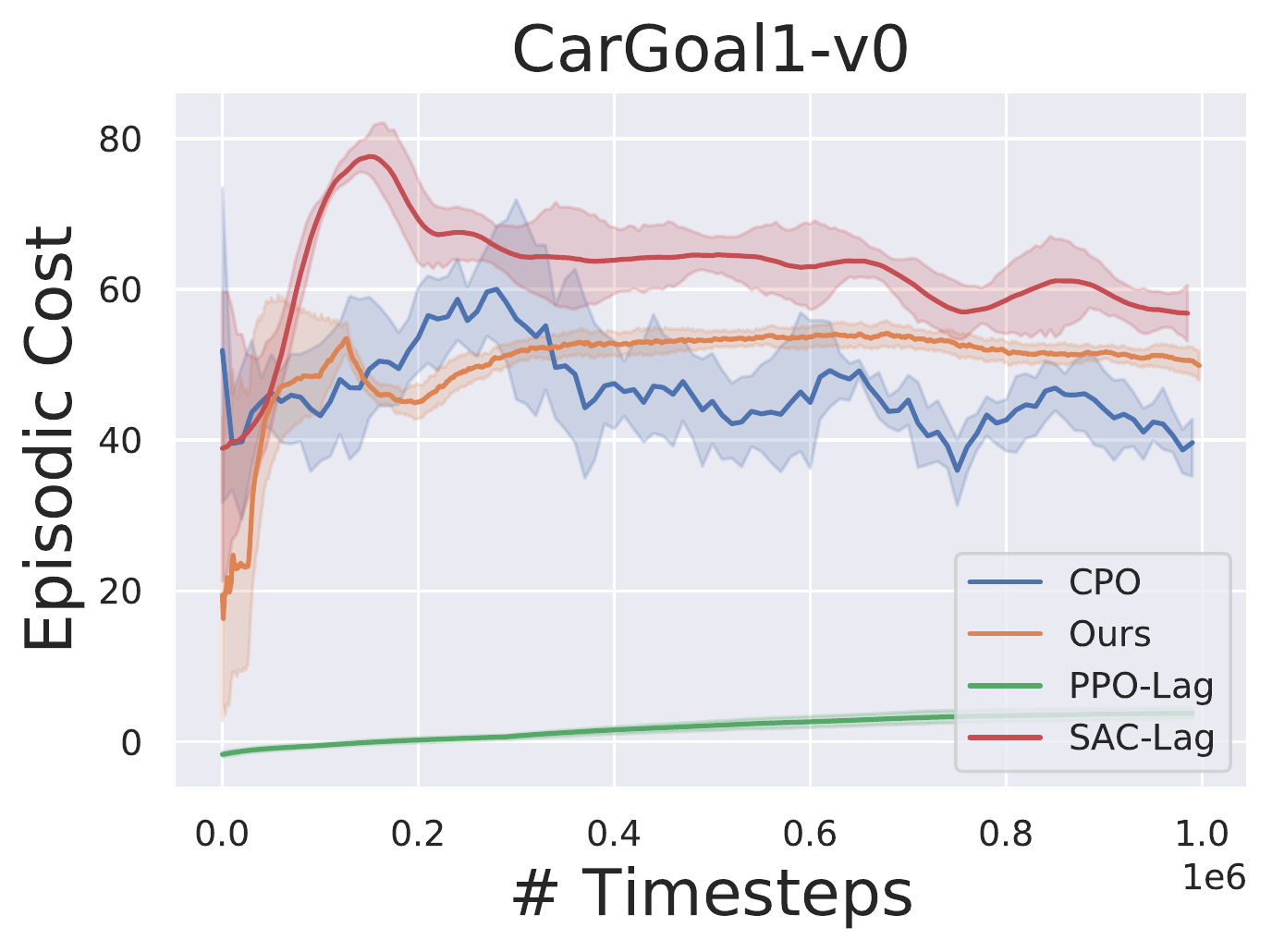}
		\end{minipage}%
		\begin{minipage}[htbp]{0.245\linewidth}
			\centering
			\includegraphics[width=1.0\linewidth]{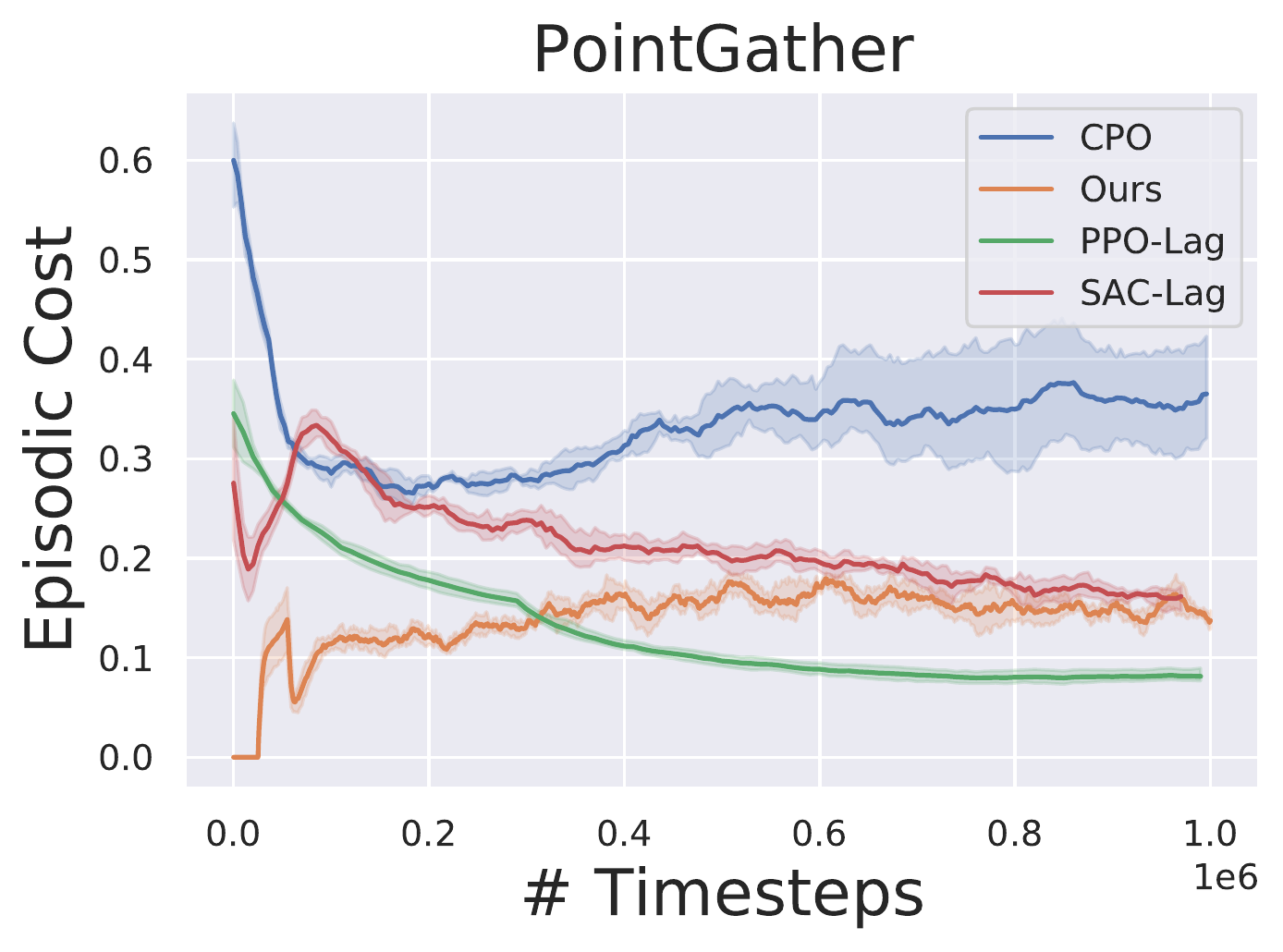}
		\end{minipage}%
		\begin{minipage}[htbp]{0.245\linewidth}
			\centering
			\includegraphics[width=1.0\linewidth]{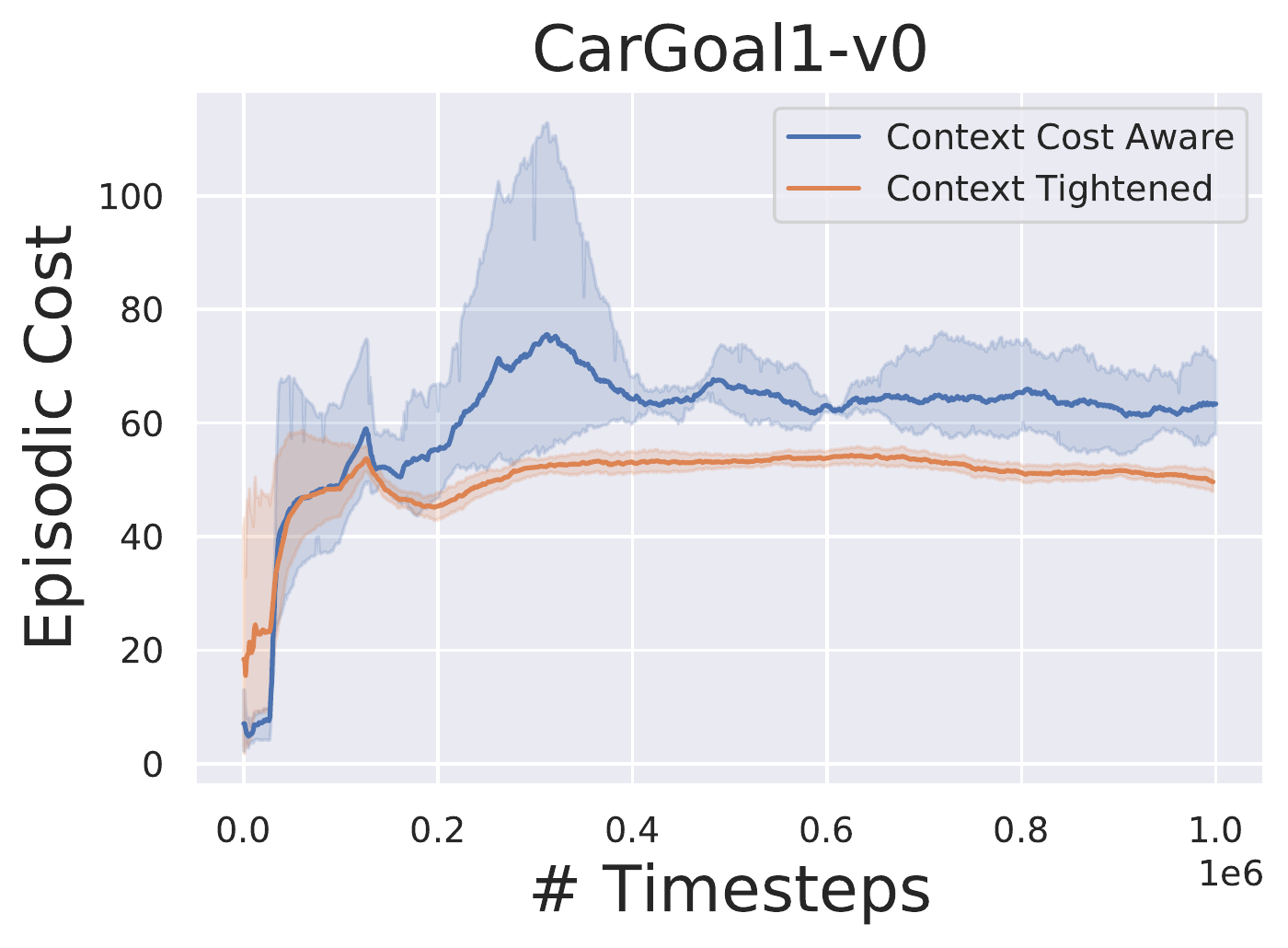}
		\end{minipage}%
\caption{Results on the three budget tasks. The first three columns show the rewards and the costs of different methods on the three environments respectively, while the last column shows the performance comparison between learning with extended state space and tightened approximation. As discussed in Sec.~\ref{sec_tightened_appx}} 
\label{fig_main_budget}
\end{center}
\vskip -0.2in
\end{figure}

\subsection{Tight Constraints}
\label{exp_tight}
In this section, we evaluate different methods on the environments where constraints do change the optimal solution. As termed in the previous section, those are tight constrained problems.
\subsubsection{Binary Tasks}
\begin{wraptable}{l}{5cm}
\caption{Success rate of different methods on the diagnostic environment.}
	\centering
	\small
	\begin{tabular}{lrr}
	\toprule
    Success Rate & Easy & Hard \\
    \midrule
     TD3 & 2/10 & 2/10 \\
    CPO & 4/10 & 0/10 \\
    PPO-Lag & 3/10 & 1/10 \\
     Ours & \textbf{10/10} & \textbf{8/10}\\
    \bottomrule
	\end{tabular}
	\vspace{-0.1in}
\end{wraptable}

We first experiment on a maze environment where an agent is asked to navigate to the goal point without stepping into the lava. The input of the agent is the coordinate of current state, and permitted action is limited to $[-1,1]$. We generate four different level of tasks. In all experiments the size of the maze is set to be $16$, and episodic length is set to be $32$, which is two times of the side length. In each episode, the agent is initialized in the center of the maze. Stepping into the target position which is located at middle of right edge will result in a $+30$ reward, and stay in the position will continuously receive that reward. A tiny punishment of $-0.1$ is applied for each timestep otherwise. 

\begin{wrapfigure}{l}{5cm}
\centering
\begin{minipage}[htbp]{1.0\linewidth}
\vspace{0.in}
			\centering
			\includegraphics[width=1.0\linewidth]{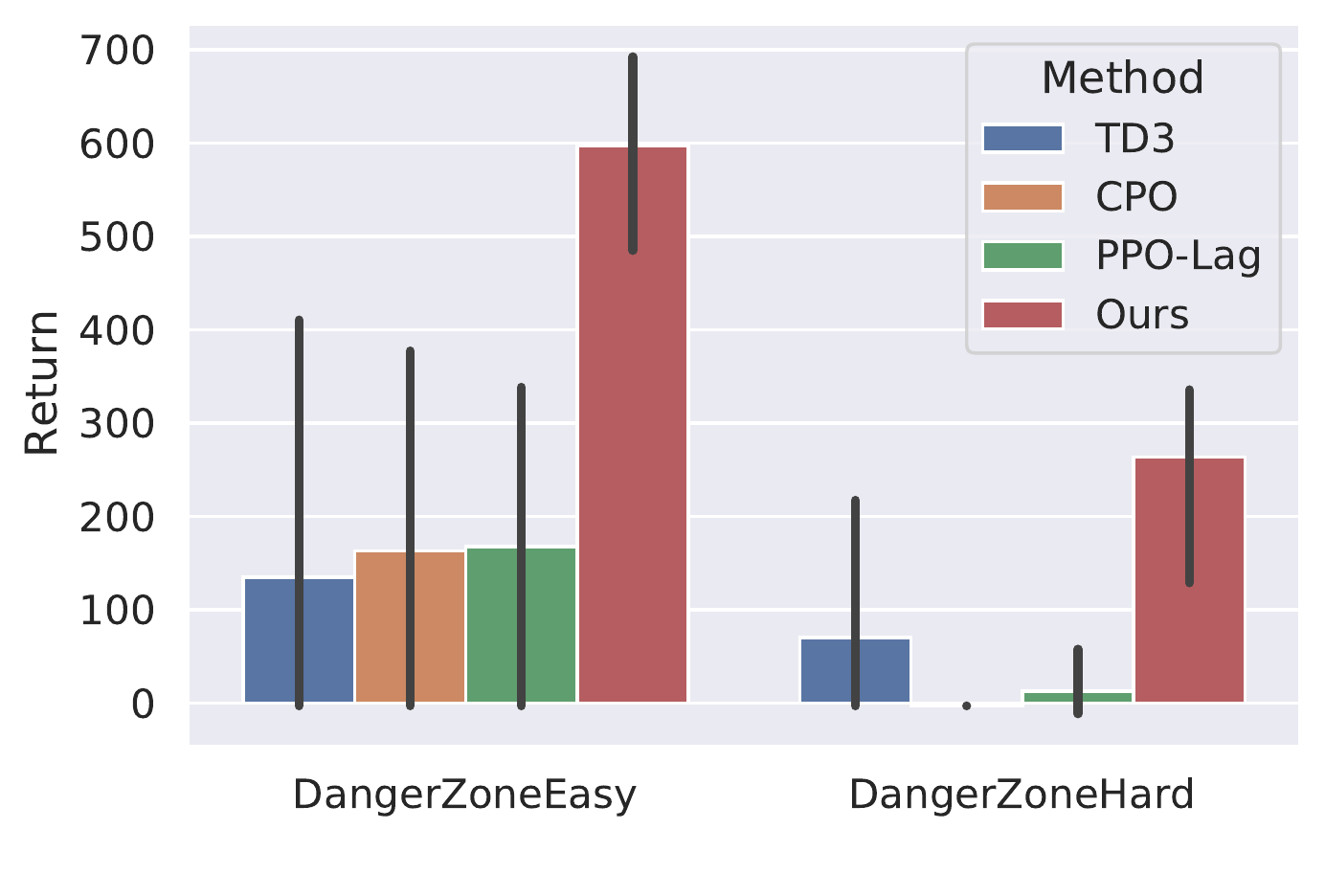}
		\end{minipage}%
\caption{Experiment Results on the diagnostic 2D navigation environment.}
\label{fig_dangerzone}
\vspace{-0.4in}
\end{wrapfigure}

In those tasks, the constraints are binary as the agent is not permitted to step into lave during navigation to the target point. Results shown in Figure~\ref{fig_main_maze} show the superior performance of ET-MDP with Context TD3 in solving those constrained MDP tasks with binary constraints. Our method enables the off-policy methods to be applied to constrained optimization with remarkable sample efficiency.


\subsubsection{Budget Tasks}
For the budget tasks, we experiment on PointGoal1-v0, CarGoal1-v0, and PointGather to show the performance of our proposed method. In PointGoal1-v0, a mass point navigates in a 2-D maze to collect reward while avoiding dangerous regions, which will lead to a $+1$ cost. The budget for the cost is set to be $+25$ in our experiments~\cite{safety_gym_Ray2019}. In CarGoal1-v0, a car replaces the mass point in the previous environment to attain the same objective and the threshold is increased to $+50$ as the task is more challenging~\cite{stooke2020responsive,bharadhwaj2020conservative}. In PointGather, a mass point is asked to collect apples while avoiding bombs which will lead to a $+1$ cost, and the cost budget is set to $0.1$~\cite{achiam2017constrained}, i.e., the agent is permitted to run into a bomb every ten games on average.

As we have shown in Section~\ref{sec_budget}, the previous information of cost should be taken as an additional input for policies to satisfy the Markov property in those environments. Another approach is to leverage tightened approximation in ET-MDP, where the budget tasks are converted to binary tasks. e.g., in all of those environments, the cost budget is set to $0$ and the episode will be terminated whenever a cost is encountered.

Figure~\ref{fig_main_budget} shows our experiment results with the binary approximation. In all experiments, ET-MDP with Context TD3 is able to reach the best asymptotic performance in terms of both high reward and low cost.
We conduct ablation studies on the approximation discussed in Section~\ref{sec_tightened_appx} to show how the empirical performance of such an approximation. We experiment on the car navigation environment to compare the performance of the primal CMDP with extended state space $\mathcal{S}_{ext} = \mathcal{S} \oplus \mathcal{S}_{budget}\oplus \mathcal{S}_{time}$ and with the tightened approximation. The last column of Figure~\ref{fig_main_budget} shows the experimental results we get: the tightened approximation leads to better constraints-satisfaction, $i.e.$, lower cost, while being able to achieve higher reward.

\subsection{Ablation study of context models under limited state visitation}
\label{exp_state_visitation}

We demonstrate the superiority of Context-TD3 model over vanilla TD3 in the diagnostic environment, where the tasks is no doubt an MDP, $i.e.$, the decision of the agent should be made only based on its present state and has no relevance to the historical information. We hence claim the improvement of Context-TD3 relies on better generalization ability rather than the memory mechanism proposed in previous works that also include recurrent networks in RL.

In this set of experiments, two different environments are generated to compare Context-TD3 and TD3. In the Random-Init environment, the initial position of the agent is uniformly distributed in the map while in the Fix-Init environment the initial position is fixed to the center of the map, which leads to a relatively limited state visitation.

The Context-TD3 performs much better than vanilla TD3 in the Fix-Init environment, showing the experiences collected in limited region can be better generalized to unseen states when context models are introduced.

\begin{figure}[t]
\vskip 0.2in
\begin{center}
\begin{minipage}[htbp]{0.245\linewidth}
			\centering
			\includegraphics[width=1.0\linewidth]{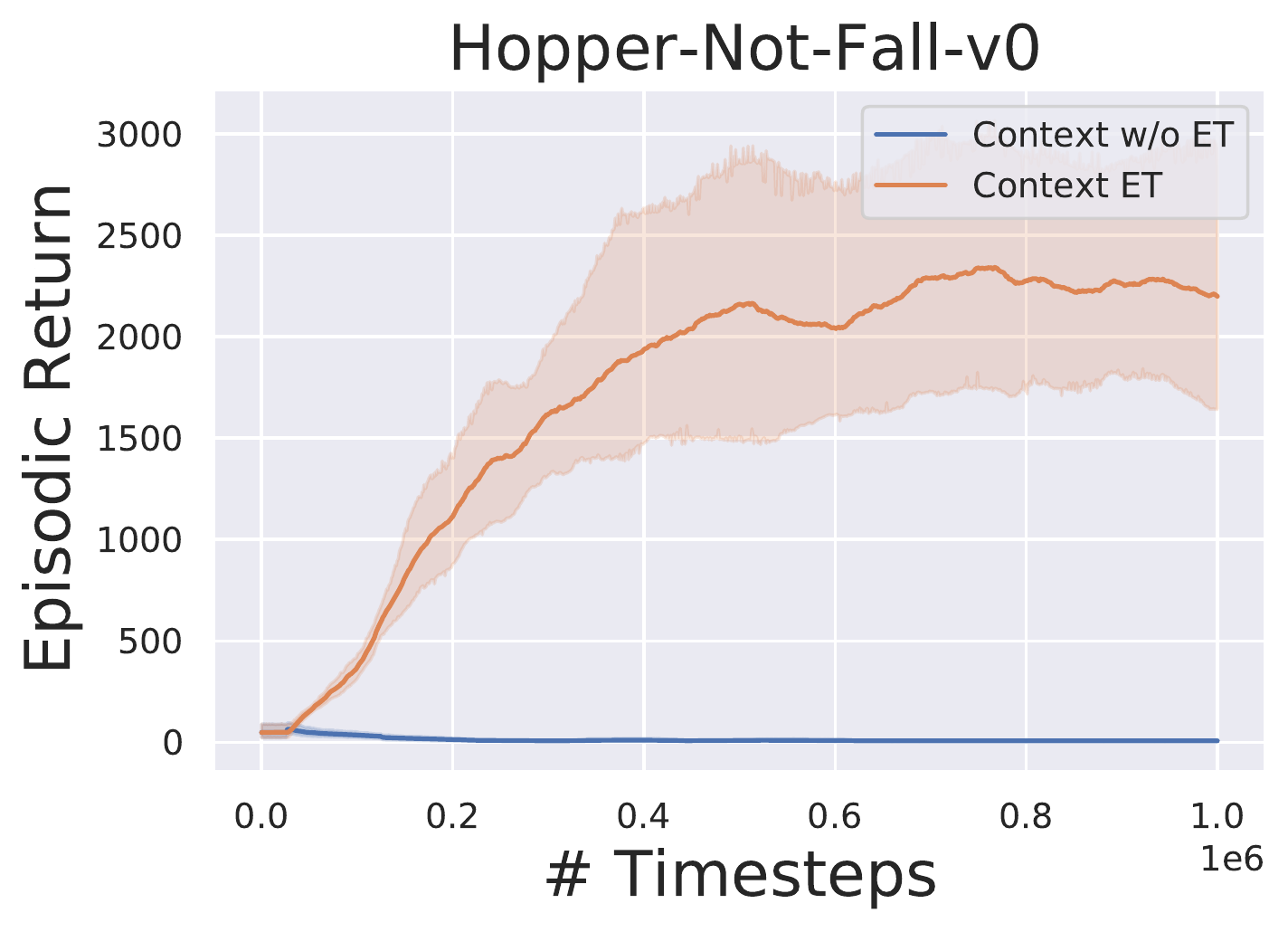}
		\end{minipage}%
		\begin{minipage}[htbp]{0.245\linewidth}
			\centering
			\includegraphics[width=1.0\linewidth]{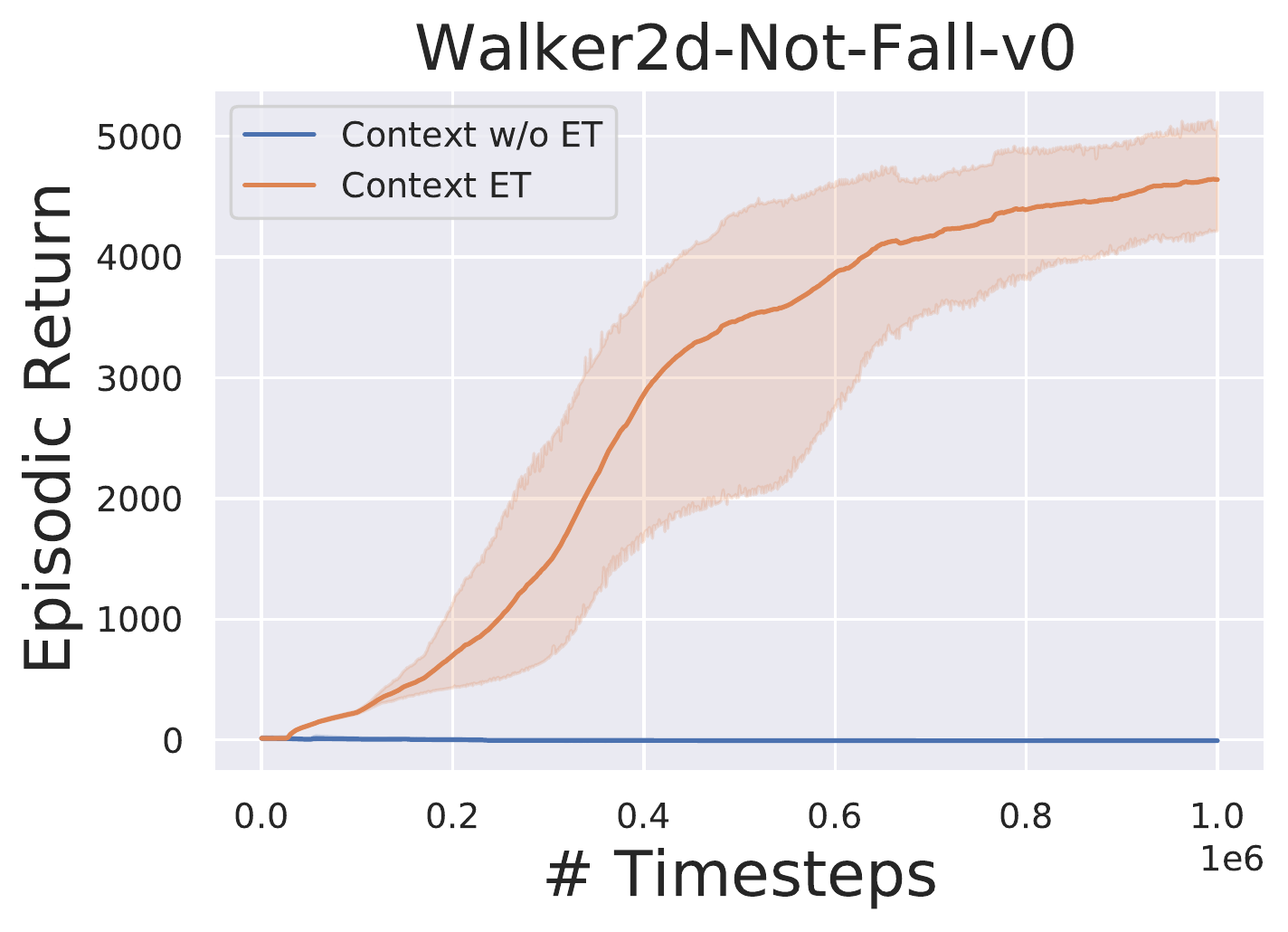}
		\end{minipage}%
		\begin{minipage}[htbp]{0.245\linewidth}
			\centering
			\includegraphics[width=1.0\linewidth]{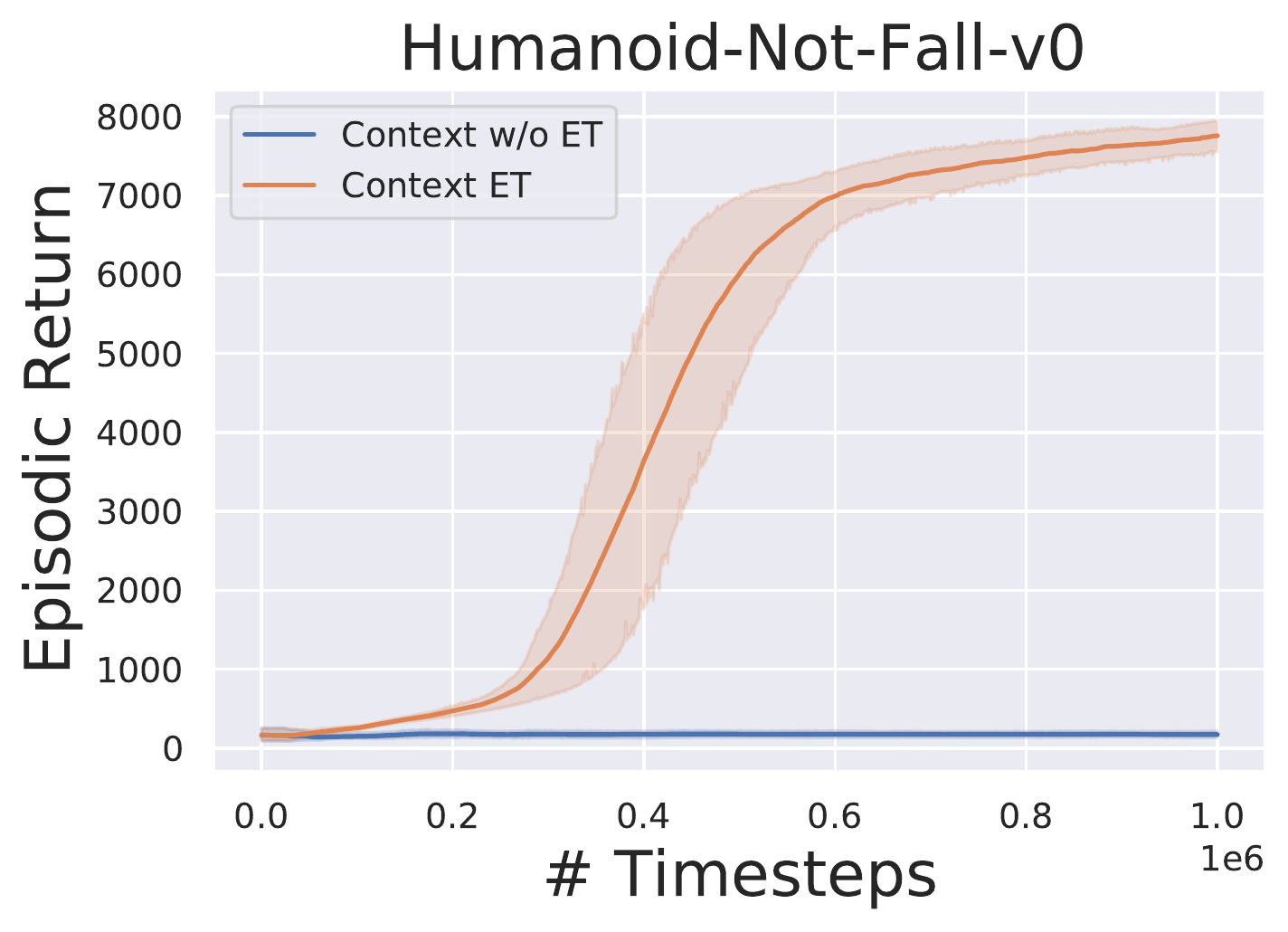}
		\end{minipage}%
		\begin{minipage}[htbp]{0.245\linewidth}
			\centering
			\includegraphics[width=1.0\linewidth]{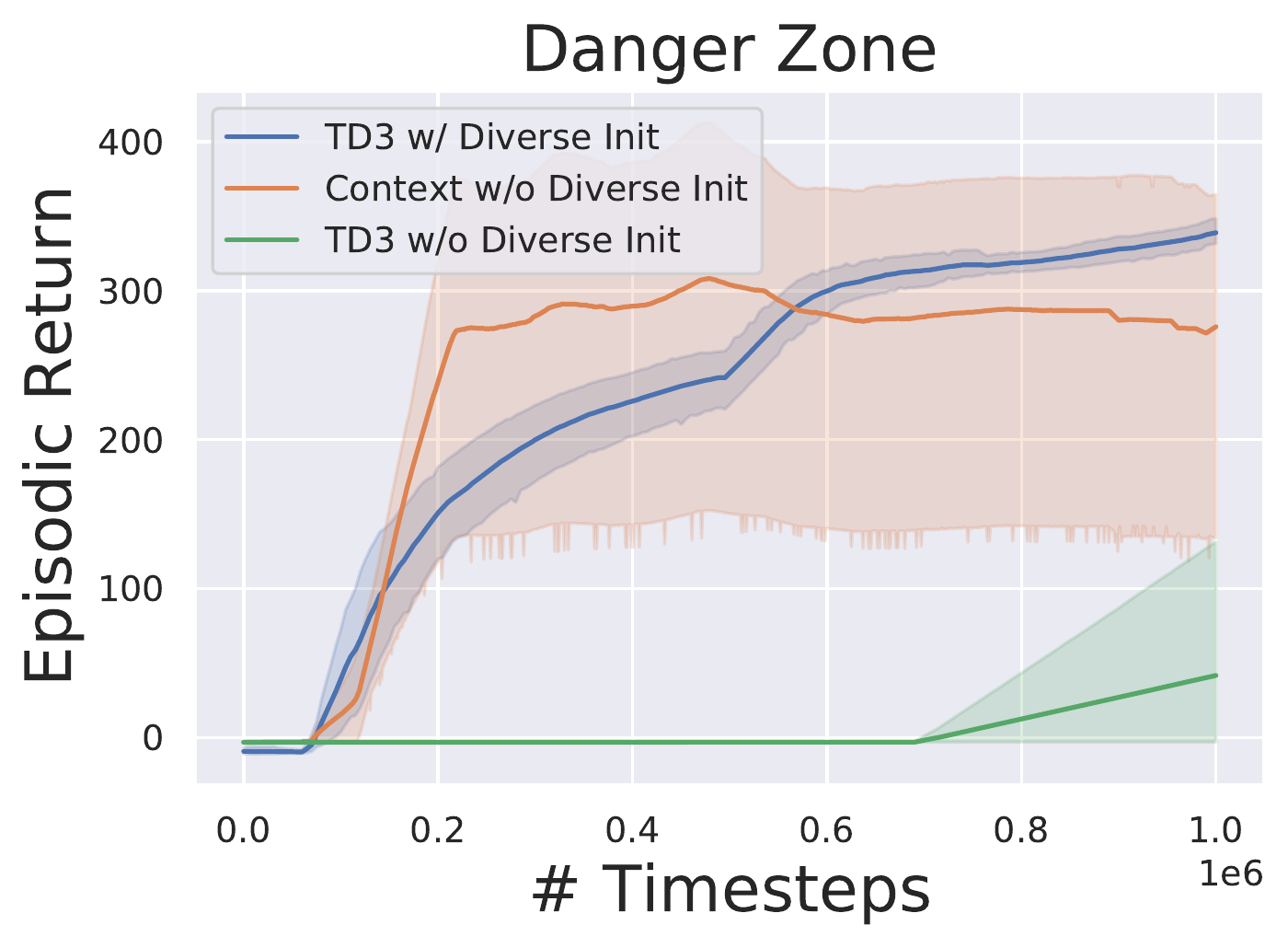}
		\end{minipage}%
\caption{The first three figures show learning curves of TD3 and Context TD3 with/without early termination trick in three MuJoCo locomotion tasks; The last figure shows that context model can remarkably improve learning efficiency when the state visitation is limited.}
\label{fig_loose_constraints}
\end{center}
\vskip -0.2in
\end{figure}

\subsection{Loose Constraints}
\label{exp_loose}

We evaluate our method on the MuJoCo locomotion benchmarks where early termination (ET) is previous applied as a default setting to benefit learning. This experiment shows that loose constraints like \textit{center of mass higher than a certain threshold} are crucial for sample-efficient learning as they greatly reduce the state space.

In this set of experiments, we remove the \textit{alive bonus} term in the reward during training, otherwise this term will be always the same (i.e., $1000$ for Hopper, Walker and $5000$ for Humanoid) and the rewards between environments with ET and without ET can not be compared fairly. In evaluation, the \textit{alive bonus} term is kept as default settings so that the asymptotic performances can be compared to previous agents trained in the vanilla environments to get a basic sense of what our policies have learned. The results are shown in Figure~\ref{fig_loose_constraints}: while both Context-TD3 and TD3 are able to learn locomotion skills when ET is applied, none of those methods get success when learning without ET. Moreover, Context-TD3 outperforms TD3 in terms of learning efficiency in all three environments.





%% file: tex/conclusion.tex
\section{Conclusion}
We address the safe exploration from the perspective of solving Early-Terminated MDP (ET-MDP).
Different from previous CMDP formulations where the constraints requires ad-hoc algorithm design, we propose an equivalent formulation of those tasks as ET-MDP, which lead to identical optimal value function as well as optimal policy to the previous CMDP formulation.
To better exploit the potential benefit of solving CMDPs with ET-MDP counterparts, we further introduce the context models to mitigate the limited state visitation problem in solving ET-MDP and improves the empirical learning efficiency in terms of better asymptotic return and lower cost in various safe RL benchmarks.

%% file: tex/appendix.tex
\onecolumn
\appendix

\section{Proof for Proposition1}
\label{prop_1_prf}
\begin{proof}
The value function of CMDP is defined in the feasible region $\Pi^c = \{\pi\in\Pi^c: \sum_{t=1}^H c(s_t,a_t)\le C, a_t=\pi(s_t), s_{t+1} = \mathcal{T}(s_t,a_t) \}$, where $\pi\in\Pi^{c}$ and
\begin{equation}
     V_c^{\pi}(s) = \sum_{t=1}^{H} r(s_t,a_t), \text{ where } \quad a_t=\pi(s_t), s_{t+1} = \mathcal{T}(s_t,a_t)
\end{equation}
The learning objective is to find $\pi\in\Pi^{c}$ such that
\begin{equation}
    V^*_c(s) = \max_{\pi\in \Pi^c} V_c^{\pi}(s).
\end{equation}
The value function for ET-MDP is defined similarly as normal MDP by 
\begin{equation}
    V_{ET}^{\pi}(s) = \sum_{t=1}^{H} r'(s_t, b_t, a_t), \text{ where }  a_t=\pi(s_t), s_{t+1} = \mathcal{T}'(s_t, b_t,a_t), b_{t+1} = b_t + c_t.
\end{equation}
For any $\pi\in\Pi^c$, the trajectories are the same in the ET-MDP and its counterpart. We have
$
    r'(s_t, b_t, a_t) = r(s_t, a_t)
$
for all $t \leq H$. Therefore, we have $V_c^{\pi} = V_{ET}^{\pi}$ for $\pi \in \Pi^c$.

The optimal value function of ET-MDP is defined over its optimal policy
\begin{equation}
     V^*_{ET}(s) = \max \{ \max_{\pi\in \Pi^c} V_c^{\pi}(s), \max_{\pi \not\in \Pi^c} \sum_{t=1}^{h_{\pi}\le H} r(s_t,a_t)+r_e\},
\end{equation}
where $h_{\pi}$ is the step at which the constraint is violated.
Therefore, $V^*_{ET}(s) = V^*_c(s)$ for sufficiently small $r_e$ and the optimal state values are achieved for the same optimal policy $\pi^*\in\Pi^c$
\end{proof}

\section{Detailed Pseudo-Code of the Proposed Method}
\label{detailed_algo}
\begin{algorithm}[h]
\caption{Context TD3 in ET-MDP}
\label{Algorithm2}
\begin{algorithmic}[1]
		\STATE Initialize critic networks $Q_{w_1}$, $Q_{w_2}$, actor network $\pi_{\theta}$
		\STATE Initialize context models $\mathcal{C}_{w_a},\mathcal{C}_{w_c}$ for the actor and critic networks separately with recurrent networks.
		\STATE Initialize target networks $w'_1 \leftarrow {w_1}$, $w'_2 \leftarrow {w_2}$, $\theta' \leftarrow {\theta}$
		\STATE Initialize replay buffer $\mathcal{B} = \{\}$
		\STATE Initialize a context queue $\mathcal{Z}_L$ with length $L$ by $\mathcal{Z}_L = [\boldsymbol{0}_s,\boldsymbol{0}_a,\boldsymbol{0}_r] \times L$, maintain a copy $\mathcal{Z}'_L \leftarrow \mathcal{Z}_L$
	    \FOR{$t = 1,2,...$}
	        \STATE Interact with environment and get transition tuple $(s,a,r,c,s')$, $r\leftarrow r+r_e$ if $c>0$.
	        \STATE Update context queue with $\mathcal{Z}_L$, append $(s,a,r)$, and store $(s,a,r,s',\mathcal{Z}'_L,\mathcal{Z}_L)$ in $\mathcal{B}$, update $\mathcal{Z}'_L\leftarrow\mathcal{Z}_L$
	        \STATE Sample a batch of transitions $\{(s,a,r,s',\mathcal{Z}'_L,\mathcal{Z}_L)\}$ from $\mathcal{B}$
	        \STATE Calculate context variable for actor and critic with $z_a = \mathcal{C}_{w_a}(\mathcal{Z}'_L), z_c = \mathcal{C}_{w_c}(\mathcal{Z}'_L)$, and context variable for calculating the next action and next value $z'_a = \mathcal{C}_{w_a}(\mathcal{Z}_L), z'_c = \mathcal{C}_{w_c}(\mathcal{Z}_L)$
	        \STATE Calculate perturbed next action by $\tilde{a}\leftarrow \pi_{\theta'}(s',z'_a) + \epsilon$, $\epsilon$ is sampled from a clipped Gaussian.
	        \STATE Calculate target critic value $y$ and update critic networks:\\
	          ~~$y \leftarrow r + \gamma \min_{i=1,2}Q_{w'_i}(s',\tilde{a},z'_c)$\\
	          ~~$w_i \leftarrow \arg\min_{w_i} \mathbf{MSE}(y,Q_{w_i}(s,a,z_c))$
	        \STATE Update $w_c$, the context model for critic through \\
	         ~~$w_c \leftarrow \arg\min_{w_c} \mathbf{MSE}(y,Q_{w_i}(s,a,\mathcal{C}_{w_c}(\mathcal{Z}'_L)))$
		    \STATE Update $\theta$ by the deterministic policy gradient, with learning rate $\eta$:\\
		    ~~$\theta\leftarrow \theta - \eta \nabla_a Q_{w_1}(s,a,z_c)|_{a=\pi_\theta(s,z_a)}\nabla_\theta \pi_\theta (s,z_a)$
		    \STATE Update $w_a$, the context model for actor according to Eqn.(\ref{eq_upd_cwa})
		    \STATE Update target networks, with $\tau \in (0,1)$: \\
		    ~~$w'_i \leftarrow \tau {w_i} + (1-\tau)w'_i$; $\theta' \leftarrow \tau {\theta} + (1-\tau)\theta'$
		    \STATE Break this episode if $c>0$.
		\ENDFOR
\end{algorithmic}
\end{algorithm}

\newpage
\section{Reproduction Checklist}
\paragraph{Network Structure}
Our implementation of Context TD3 is mainly based on the code of ~\cite{fujimoto2018addressing}. The hyper-parameters of TD3 are the same as the authors recommend in the paper. In our Context TD3, we also use 3-layer MLPs for both actor and critic networks (with 256 hidden units).

We find in our experiments using separated context networks that trained through gradients of actor and critic will benefit learning. Details of ablation study on the network structure are provided in Appendix~\ref{appd_sep_struc}

\paragraph{Value of {$r_e$}} In our analysis, the value of $r_e$ can be selected as any sufficiently small number. However selecting too small value may lead to over-conservative behavior. In our experiments reported in the main text, we use $r_e=-10$ for the maze environments and $r_e=-1$ for the other environments. Ablation studies on the selection $r_e$ are provided in Appendix~\ref{appd_r_e}.

\paragraph{Batch Size} In our experiments we follow ~\citet{fujimoto2018addressing} to use a mini-batch size of $256$. In PPO, CPO and PPO-Largrangian, we use a batch size of $1000$ and mini-batch size of $256$ for the short-horizon games (e.g., Maze, PointGather, both with $T\le32$), so that there are around $1000$ episodes in training. For the long-horizon games where $T\sim 1000$, we collect $10$ trajectories for each episode for better training stability~\cite{schulman2015trust,schulman2017proximal}.

\paragraph{Hardware and Training Time}
We experiment on a server with 8 TITAN X GPUs and 32 Intel(R) E5-2640 CPUs. Experiments take $0.5$ (the maze environment with 0.1M interactions) to $10$ hours (the safety-gym with 1M interactions) to run. The training of Context TD3 will introduce higher computation expense as additional context models need to be trained.

\section{Environments Details}
\label{appd_env_details}

\subsection{Maze Environment}
In the Maze environment, an agent is asked to navigate to the goal point without stepping into the lava. The input of the agent is the coordinate of current state, and permitted action is limited to $[-1,1]$. We generate four different level of tasks. In all experiments the size of the maze is set to be $16$, and episodic length is set to be $32$, which is two times of the side length. In each episode, the agent is initialized in the center of the maze. Stepping into the target position which is located at middle of right edge will result in a $+30$ reward, and stay in the position will continuously receive that reward. A tiny punishment of $-0.1$ is applied for each timestep otherwise. 

\begin{figure}[h]
\vskip 0.2in
\begin{center}
\begin{minipage}[htbp]{0.25\linewidth}
			\centering
			\includegraphics[width=1.0\linewidth]{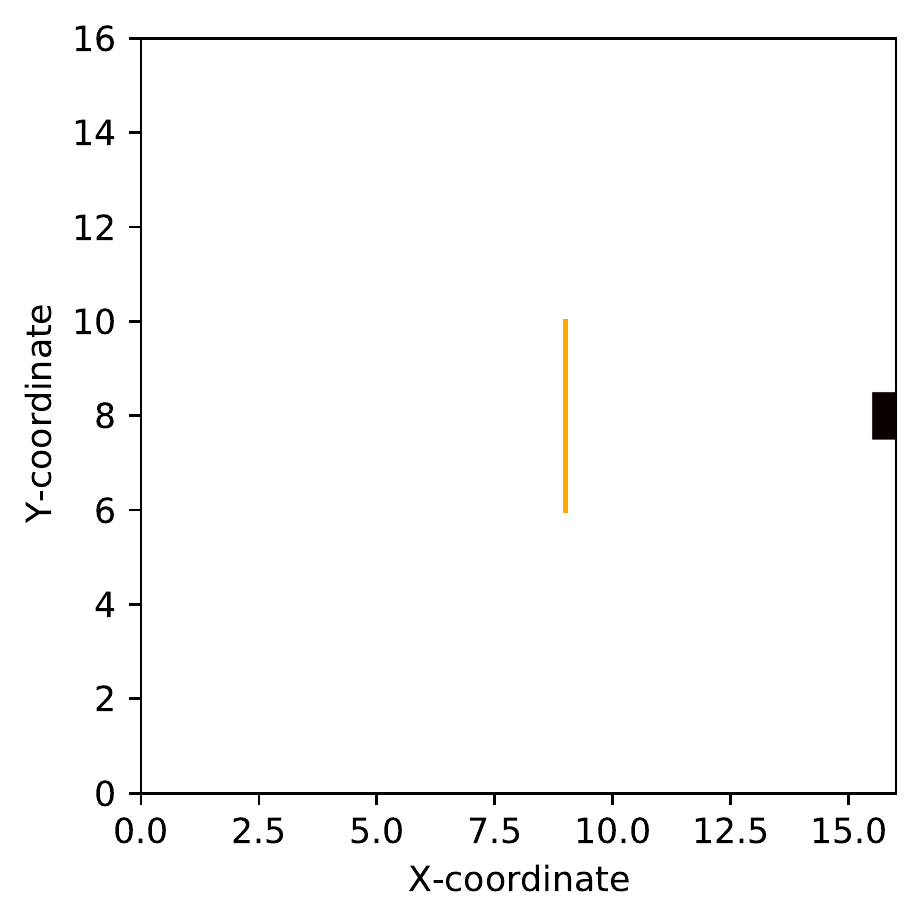}
		\end{minipage}%
		\begin{minipage}[htbp]{0.25\linewidth}
			\centering
			\includegraphics[width=1.0\linewidth]{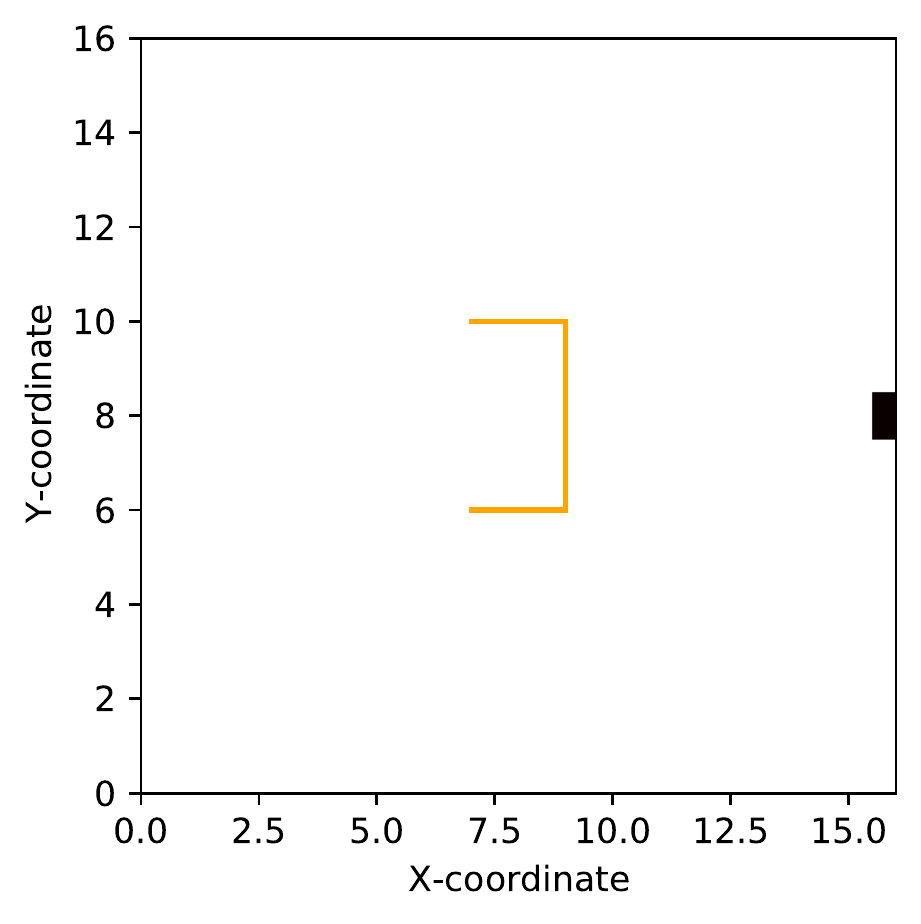}
		\end{minipage}%
		\begin{minipage}[htbp]{0.25\linewidth}
			\centering
			\includegraphics[width=1.0\linewidth]{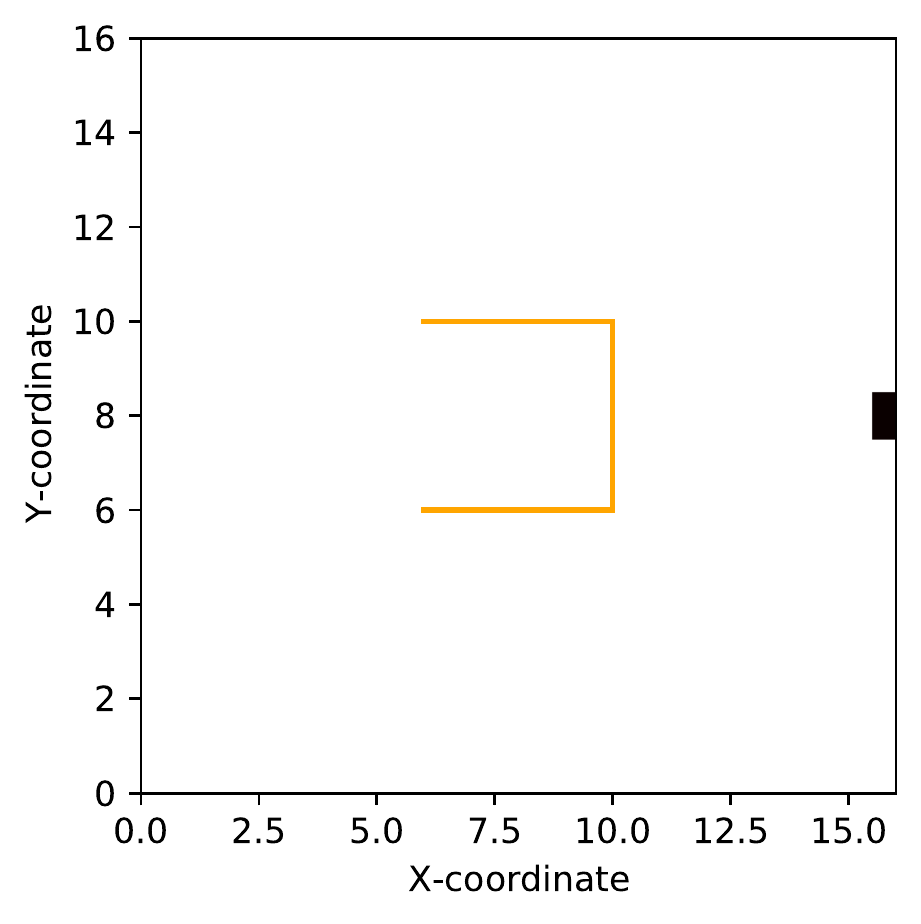}
		\end{minipage}%
		\begin{minipage}[htbp]{0.25\linewidth}
			\centering
			\includegraphics[width=1.0\linewidth]{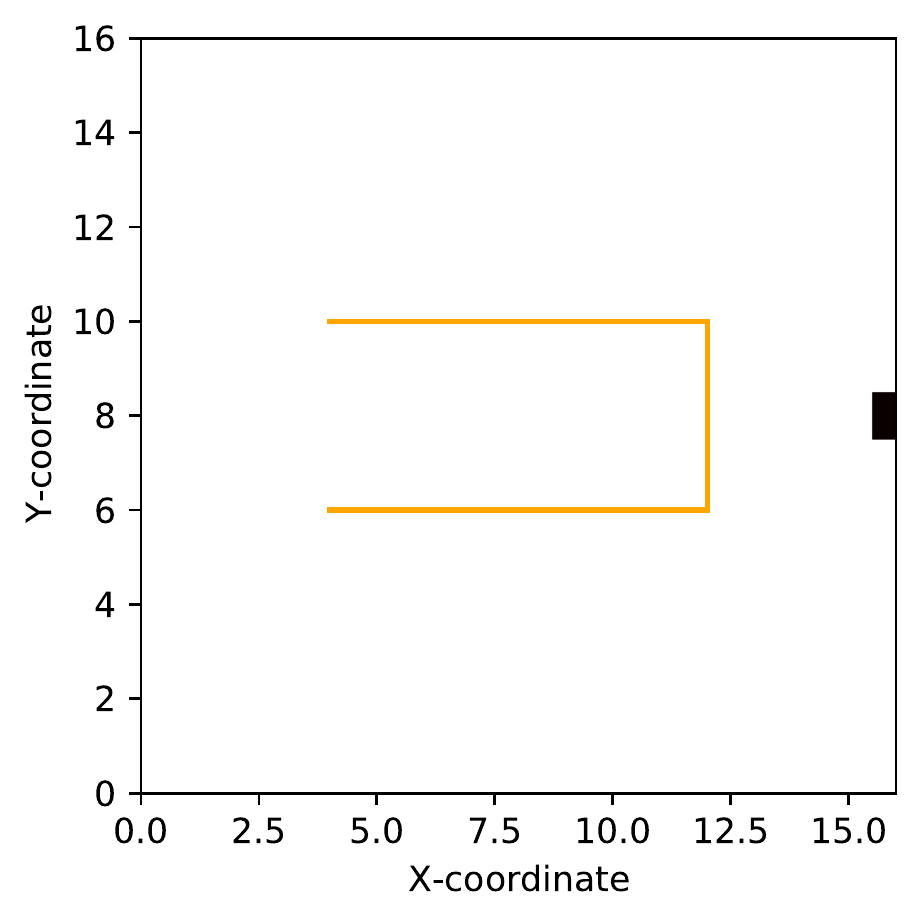}
		\end{minipage}%
\caption{The four environments with different constraint levels in our experiments. From left to right: Maze-Level-1, Maze-Level-2, Maze-Level-3, Maze-level-4. The regions with orange color are Lava region where the agent should not step into. For each game, the agent is initialized at center of the map, therefore the difficulty of finding a solution without violating the constraints becomes harder and harder from Level-1 to Level-4.}
\label{fig_maze_envs}
\end{center}
\vskip -0.2in
\end{figure}


\section{Additional Experiments}
\subsection{Sensitivity to Hyper-Parameter}
\subsubsection{Value of Historical Horizon}
\begin{figure}[h]
\vskip 0.2in
\begin{center}
\begin{minipage}[htbp]{0.25\linewidth}
			\centering
			\includegraphics[width=1.0\linewidth]{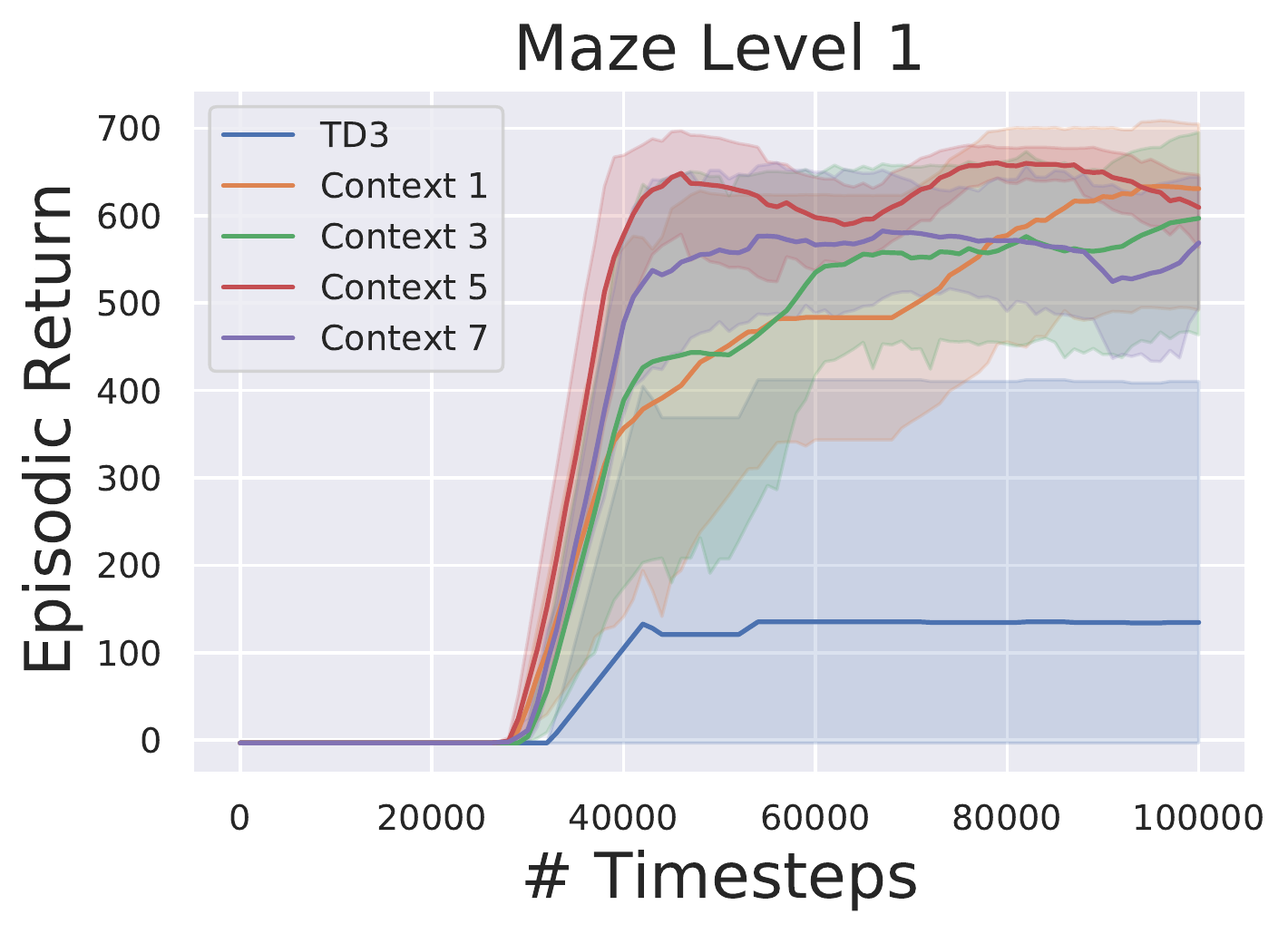}
		\end{minipage}%
		\begin{minipage}[htbp]{0.25\linewidth}
			\centering
			\includegraphics[width=1.0\linewidth]{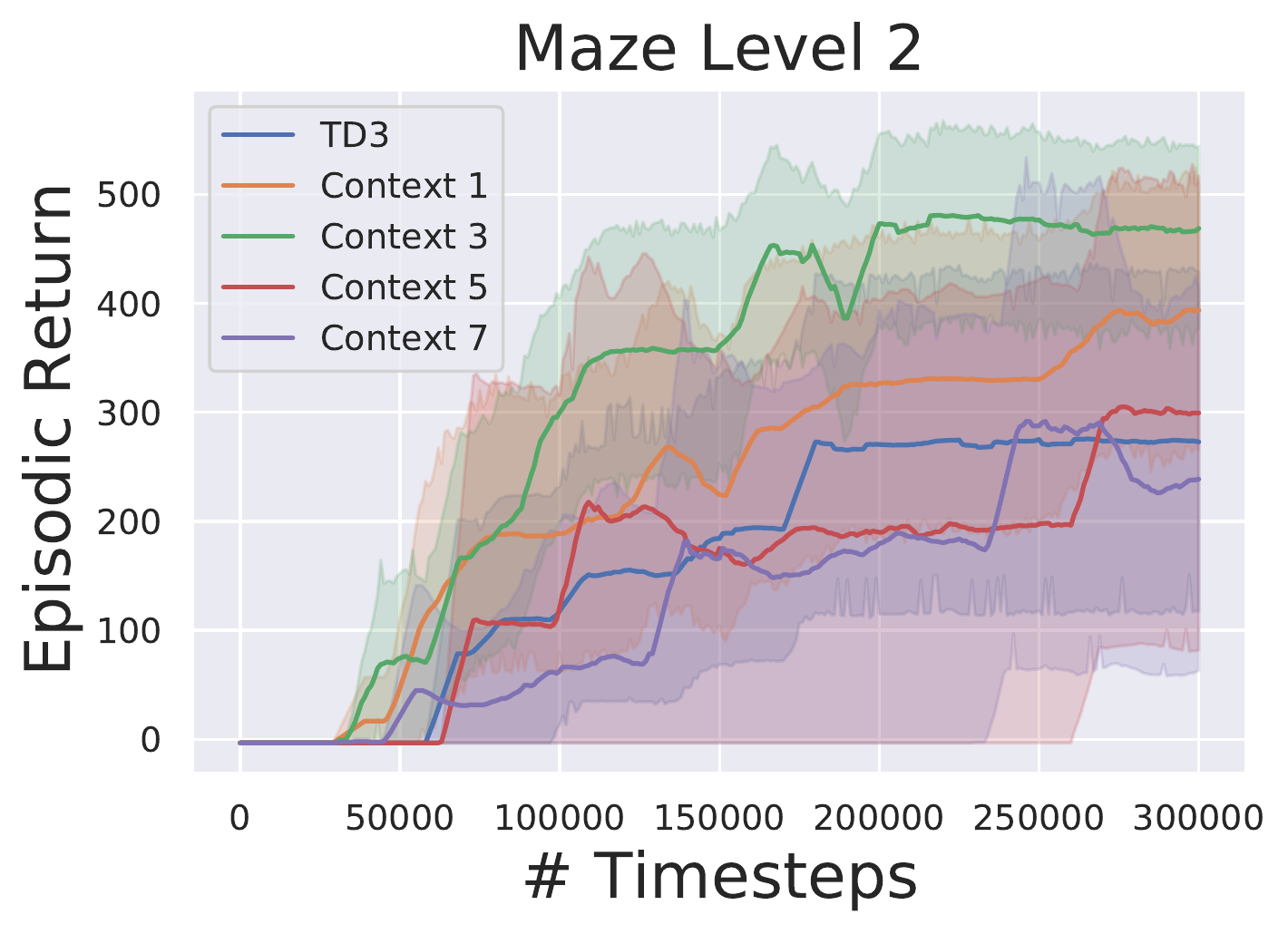}
		\end{minipage}%
		\begin{minipage}[htbp]{0.25\linewidth}
			\centering
			\includegraphics[width=1.0\linewidth]{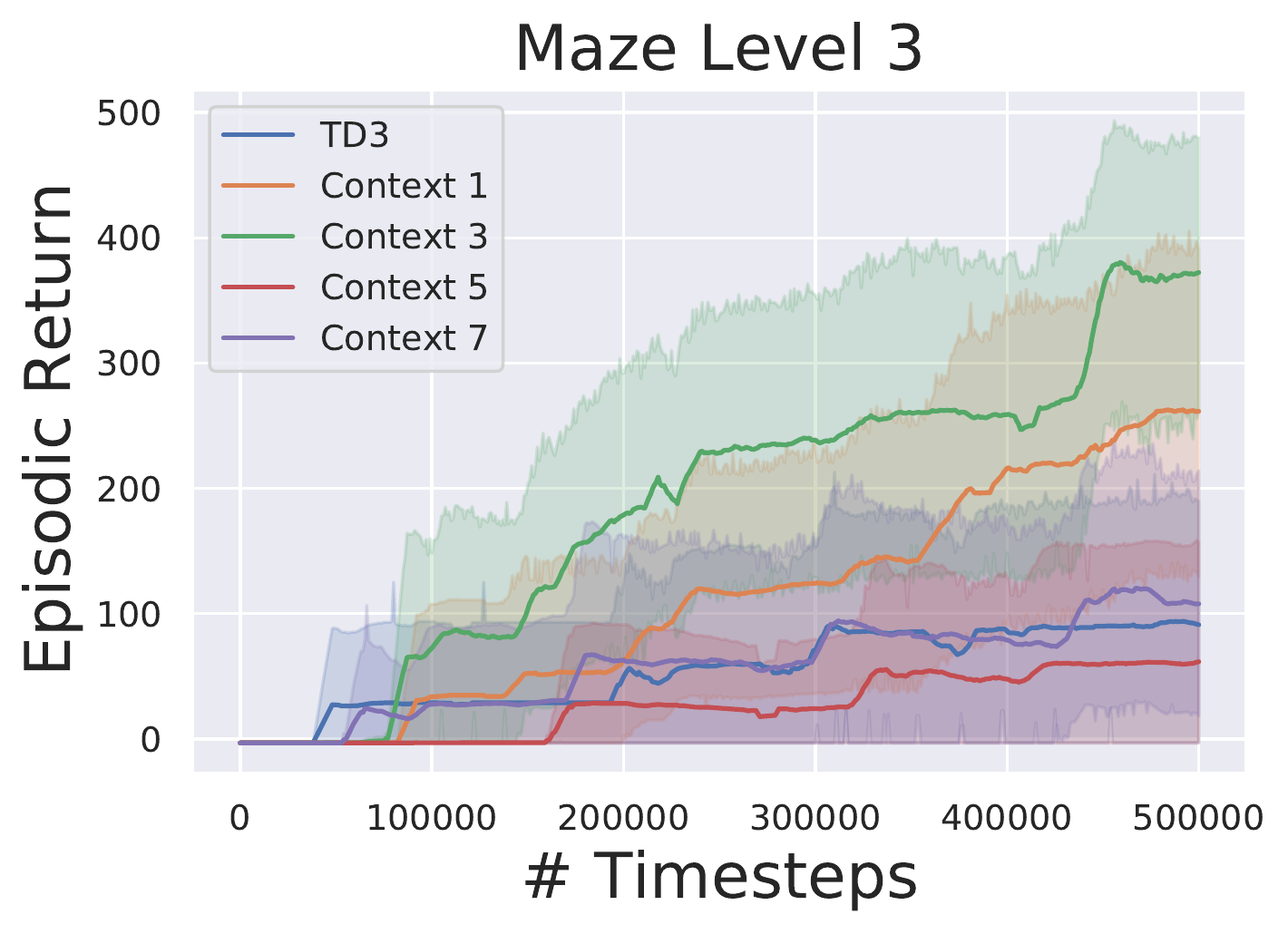}
		\end{minipage}%
		\begin{minipage}[htbp]{0.25\linewidth}
			\centering
			\includegraphics[width=1.0\linewidth]{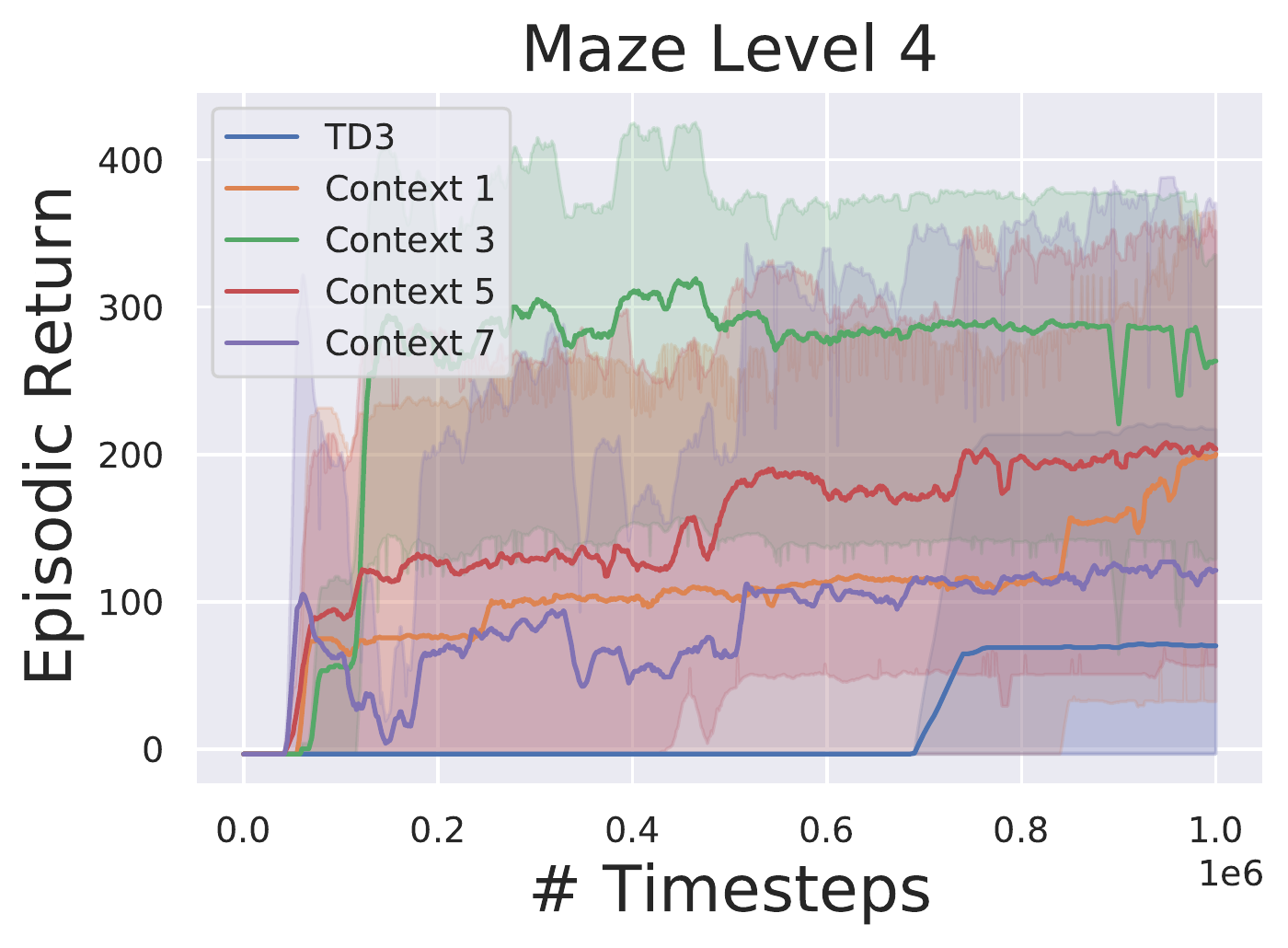}
		\end{minipage}%
\caption{Ablation studies on the selection of different length of historical horizon. All corresponding costs are zero under our ETMDP settings.}
\label{fig_hist_horizon}
\end{center}
\vskip -0.2in
\end{figure}

We experiment on the maze environments to show how the proposed method work with different length of historical horizon in the context model. Results are shown in Figure~\ref{fig_hist_horizon}. \textbf{Context $1$} means we only include the past state, action, reward in the computation of context variables, while \textbf{Context $7$} indicates the past $7$ steps of transitions are leveraged in generating the context variables. We find the context model with historical horizon $3$ achieve good performance in all level of the maze environments.

\subsubsection{Number of Hidden Units in GRUs}
\begin{figure}[h!]
\vskip 0.2in
\begin{center}
\begin{minipage}[htbp]{0.33\linewidth}
			\centering
			\includegraphics[width=1.0\linewidth]{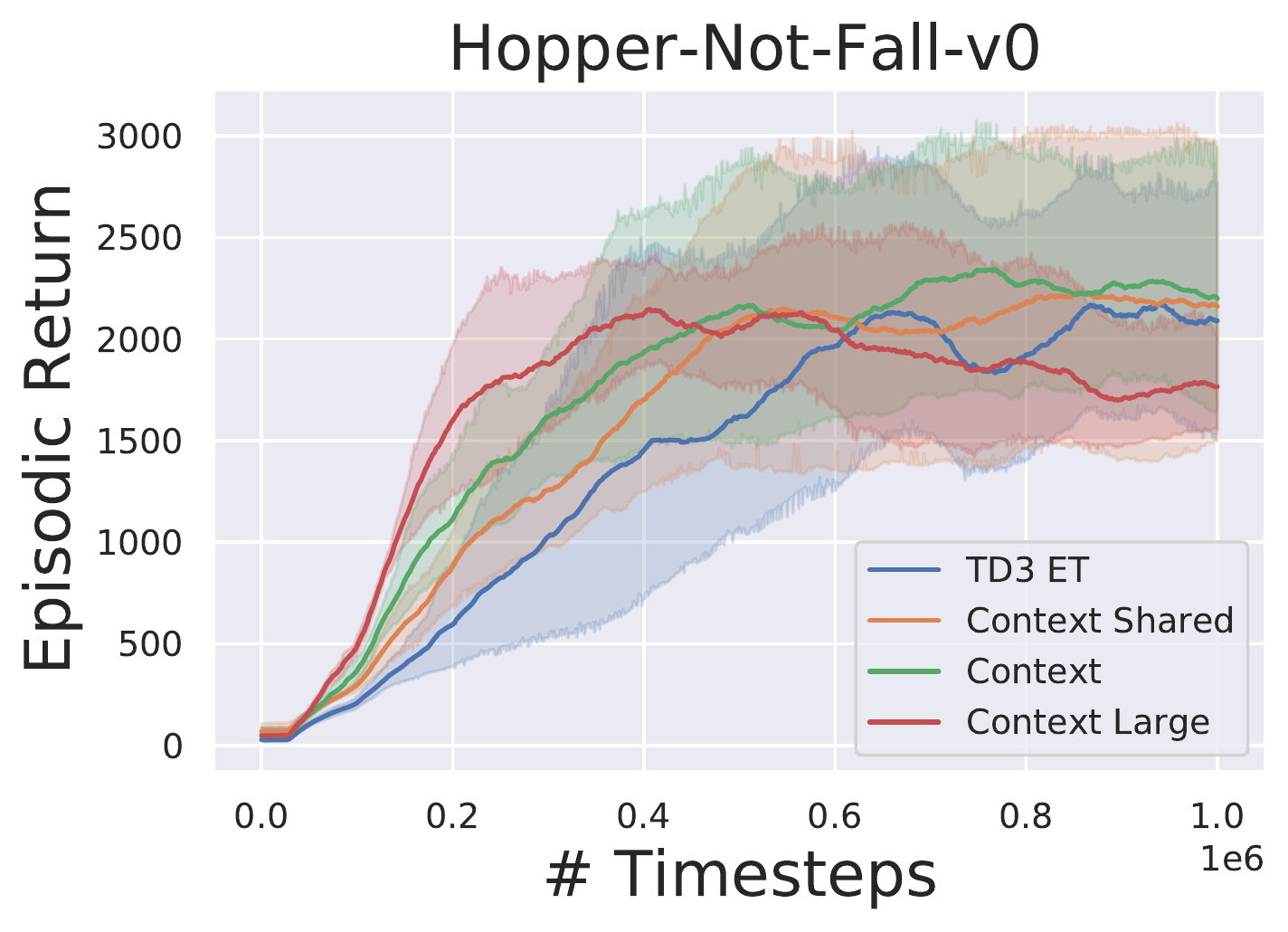}
		\end{minipage}%
		\begin{minipage}[htbp]{0.33\linewidth}
			\centering
			\includegraphics[width=1.0\linewidth]{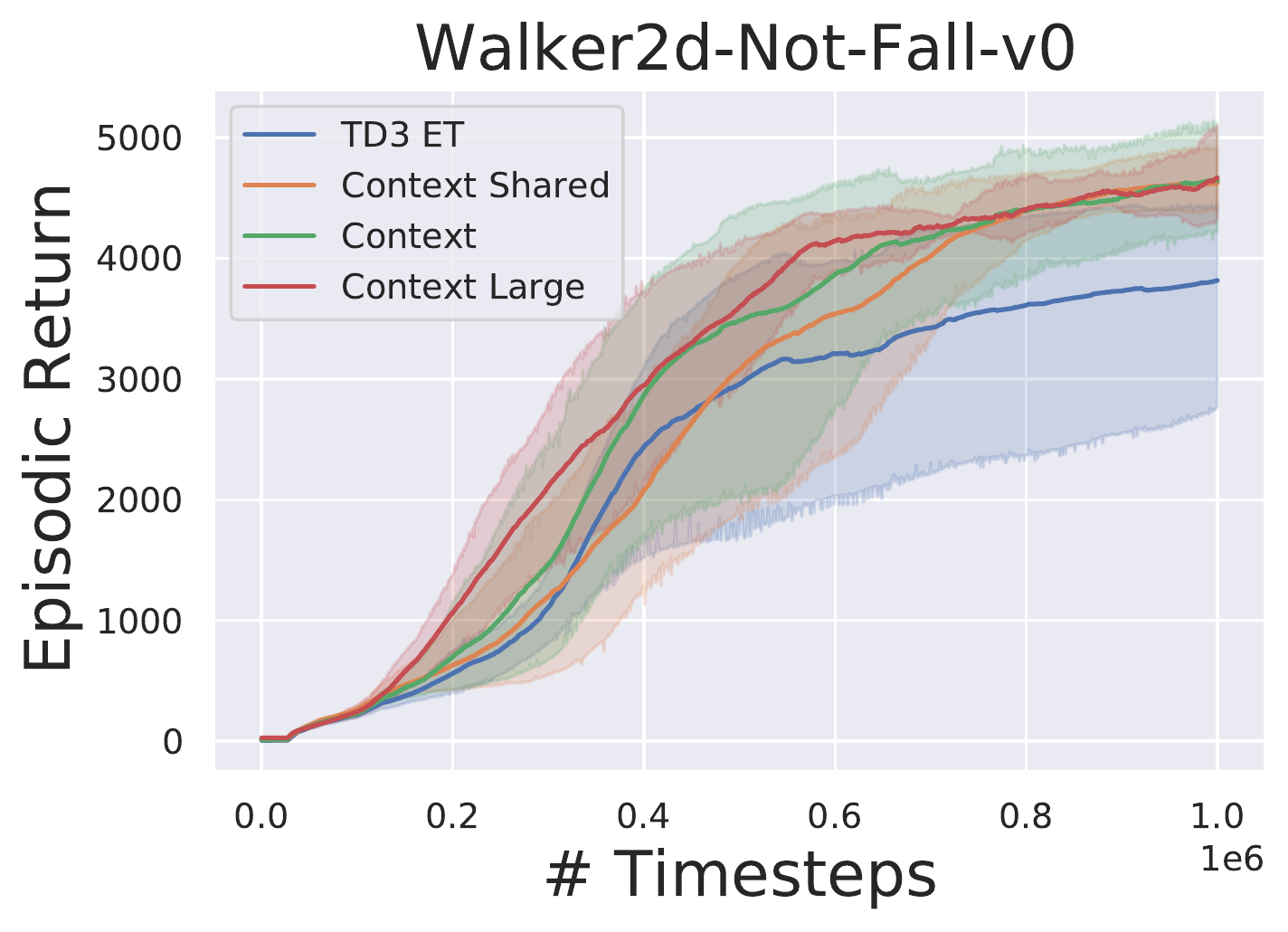}
		\end{minipage}%
		\begin{minipage}[htbp]{0.33\linewidth}
			\centering
			\includegraphics[width=1.0\linewidth]{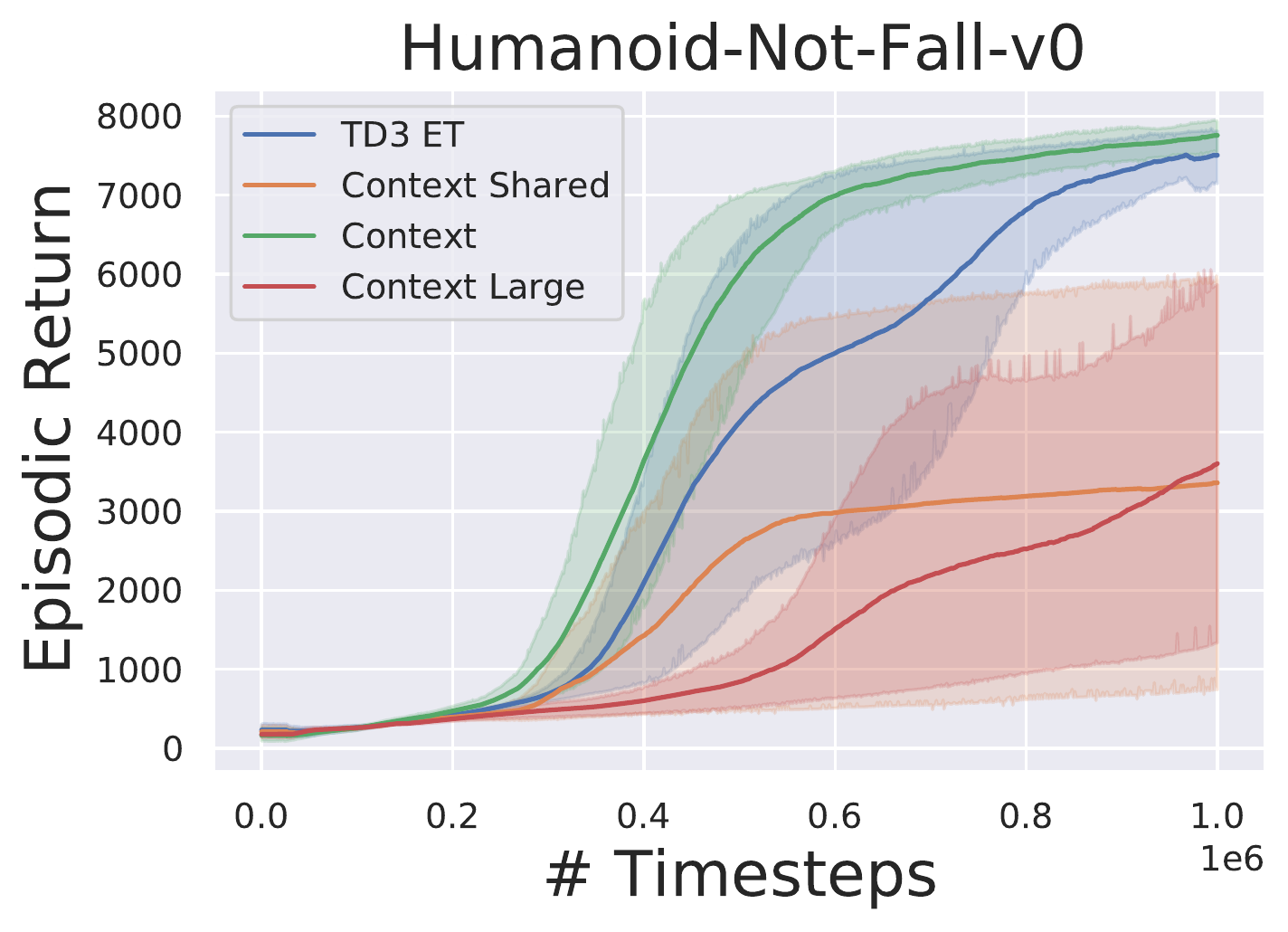}
		\end{minipage}%
\caption{Ablation studies on the number of hidden units used in GRU, and comparison on different selection of network structure: shared v.s. separated context model.}
\label{fig_hidden_units}
\end{center}
\vskip -0.2in
\end{figure}

We experiment on the selection of different number of hidden units used in GRUs. We compare the results with $30$ hidden units (reported in the main text, denoted as \textbf{Context} in Figure~\ref{fig_hidden_units}) with the results with $120$ hidden units (denoted as \textbf{Context Large} in Figure~\ref{fig_hidden_units})). We find using $30$ hidden units is enough to achieve improved performance, and in the same time balance the computational cost. And using too much hidden units may lead to reduction on learning efficiency (in the Humanoid-Not-Fall-v0 environment).

\subsubsection{Value of $r_e$}
\label{appd_r_e}
\begin{figure}[h]
\vskip 0.2in
\begin{center}
\begin{minipage}[htbp]{0.33\linewidth}
			\centering
			\includegraphics[width=1.0\linewidth]{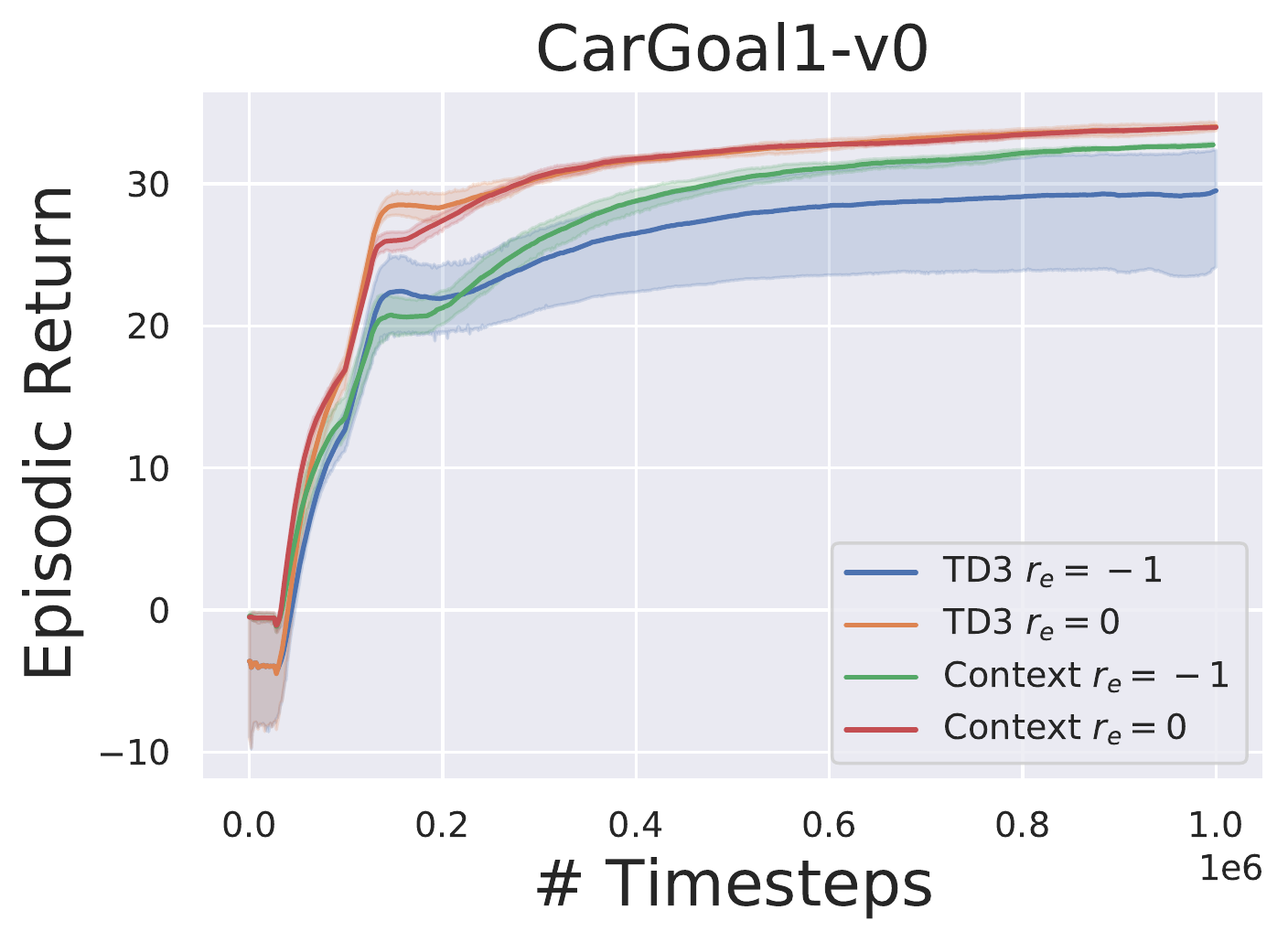}
		\end{minipage}%
		\begin{minipage}[htbp]{0.33\linewidth}
			\centering
			\includegraphics[width=1.0\linewidth]{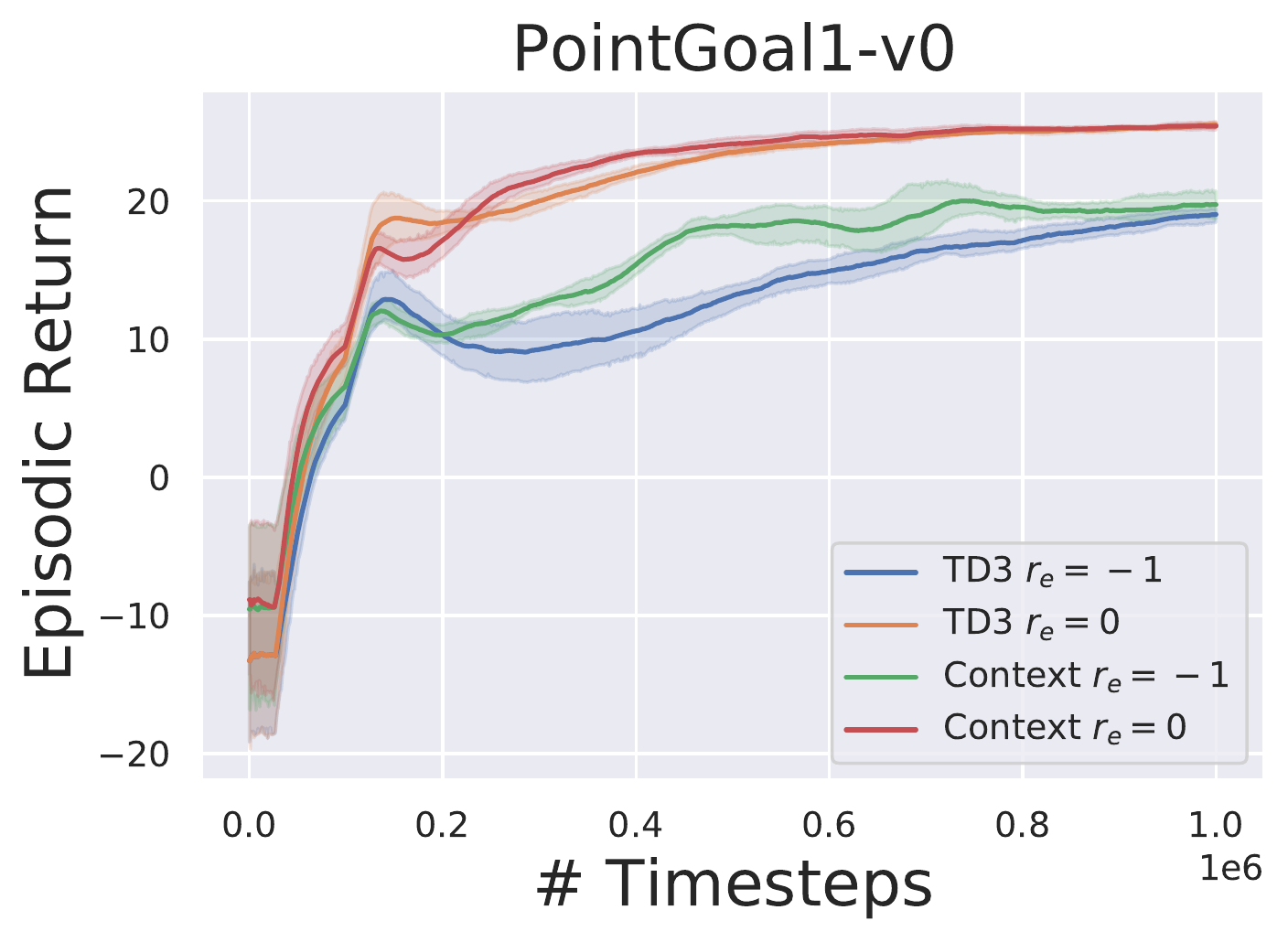}
		\end{minipage}%
		\begin{minipage}[htbp]{0.33\linewidth}
			\centering
			\includegraphics[width=1.0\linewidth]{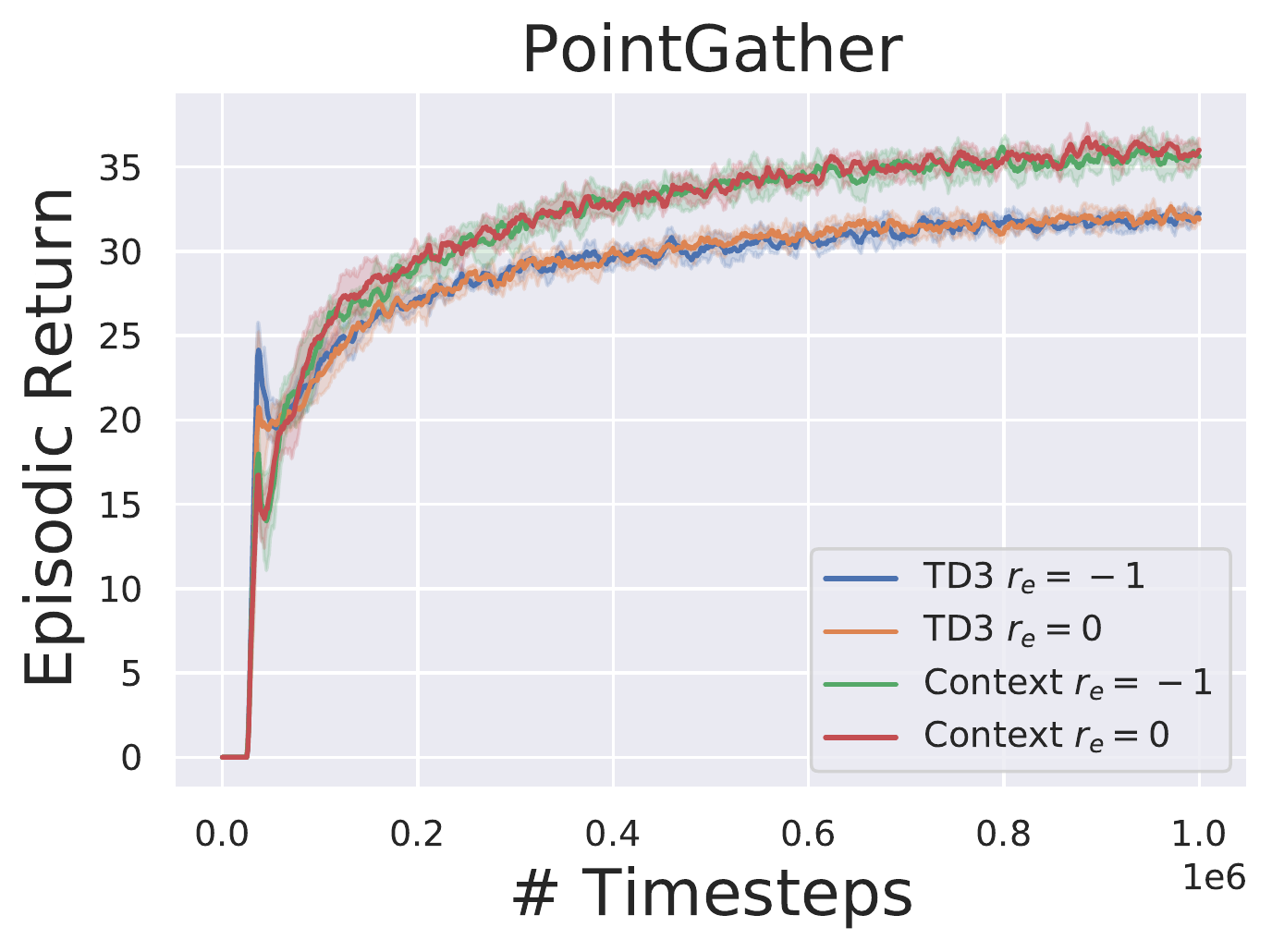}
		\end{minipage} \\%
		\begin{minipage}[htbp]{0.33\linewidth}
			\centering
			\includegraphics[width=1.0\linewidth]{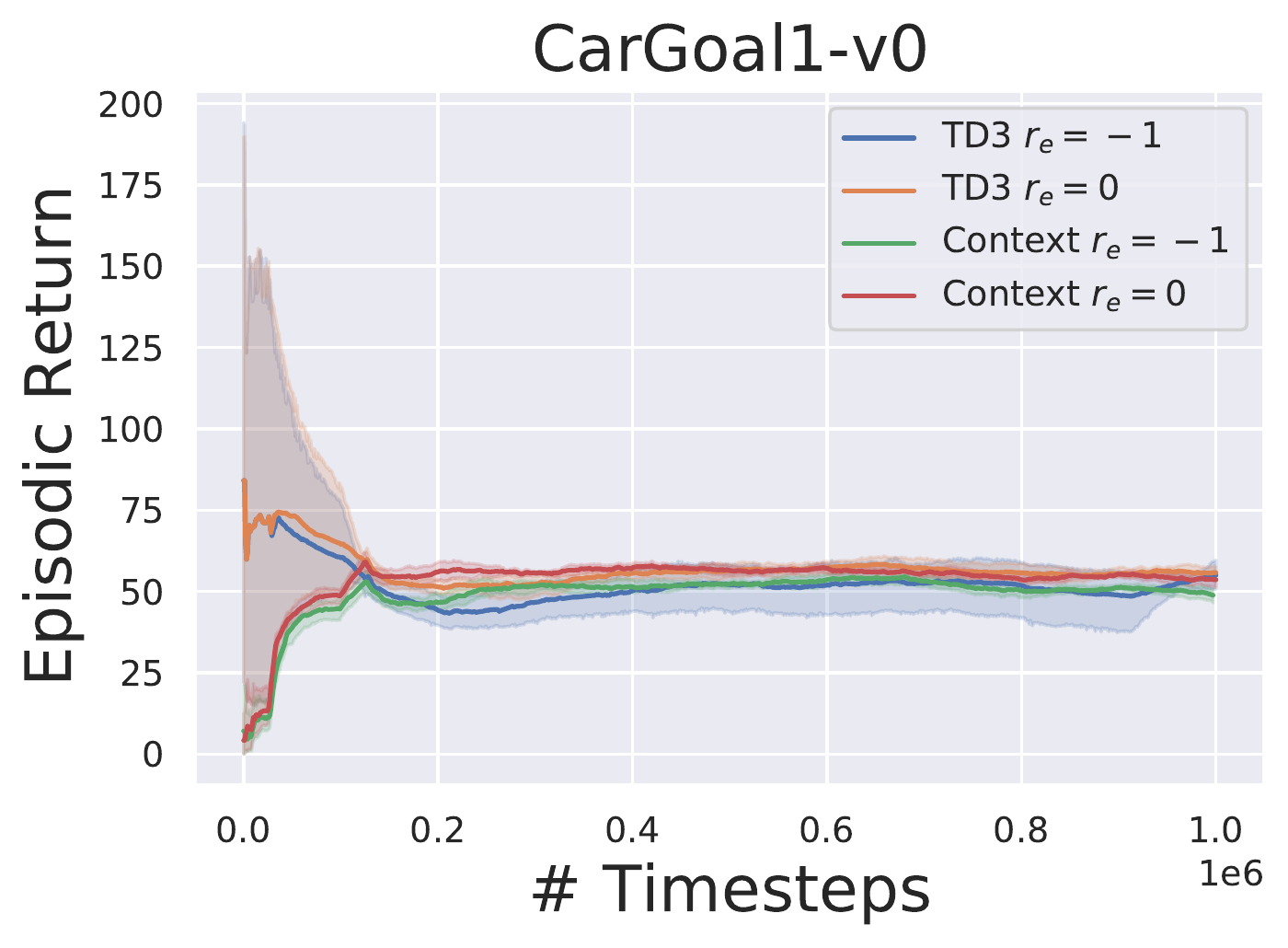}
		\end{minipage}%
		\begin{minipage}[htbp]{0.33\linewidth}
			\centering
			\includegraphics[width=1.0\linewidth]{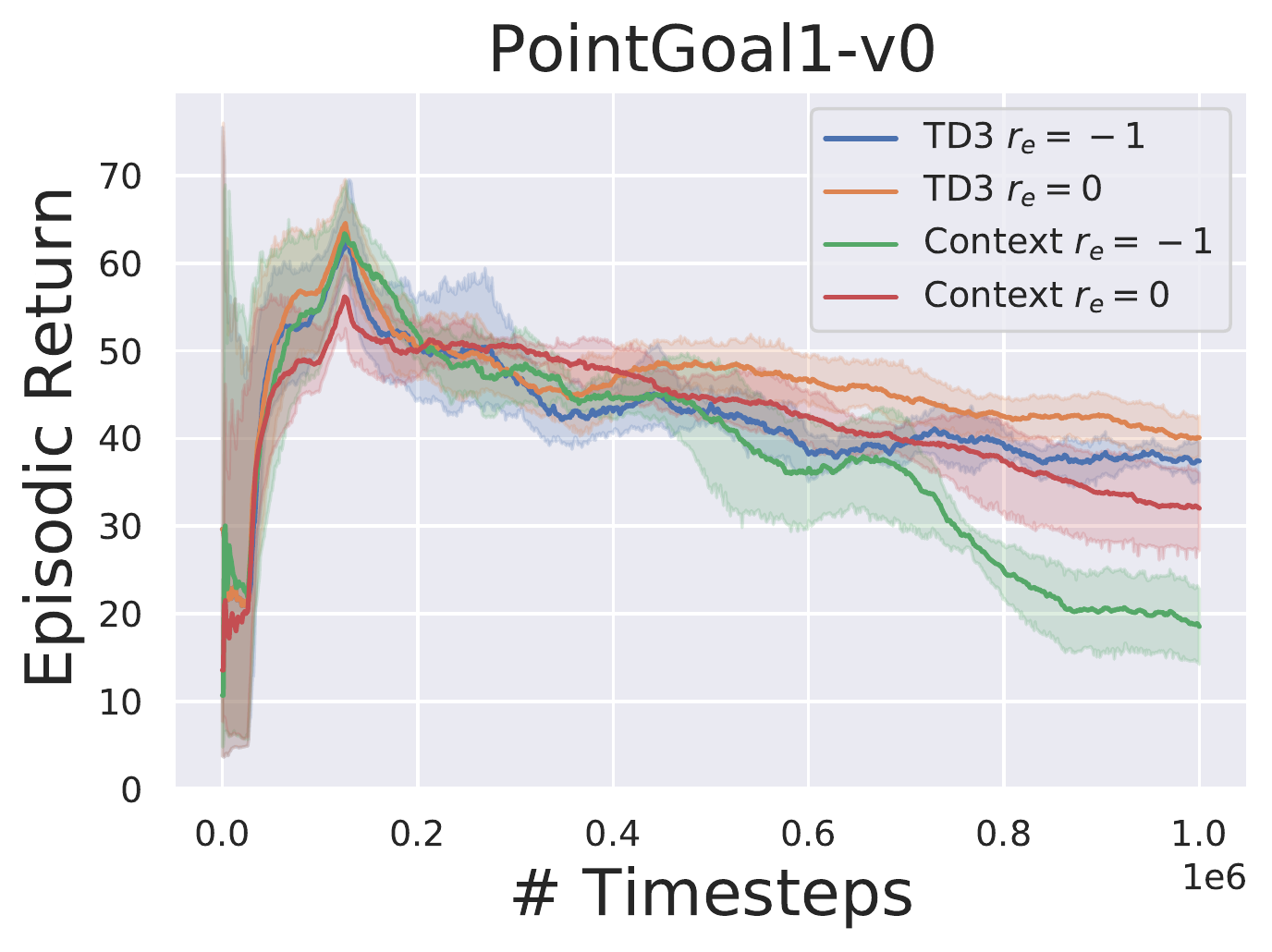}
		\end{minipage}%
		\begin{minipage}[htbp]{0.33\linewidth}
			\centering
			\includegraphics[width=1.0\linewidth]{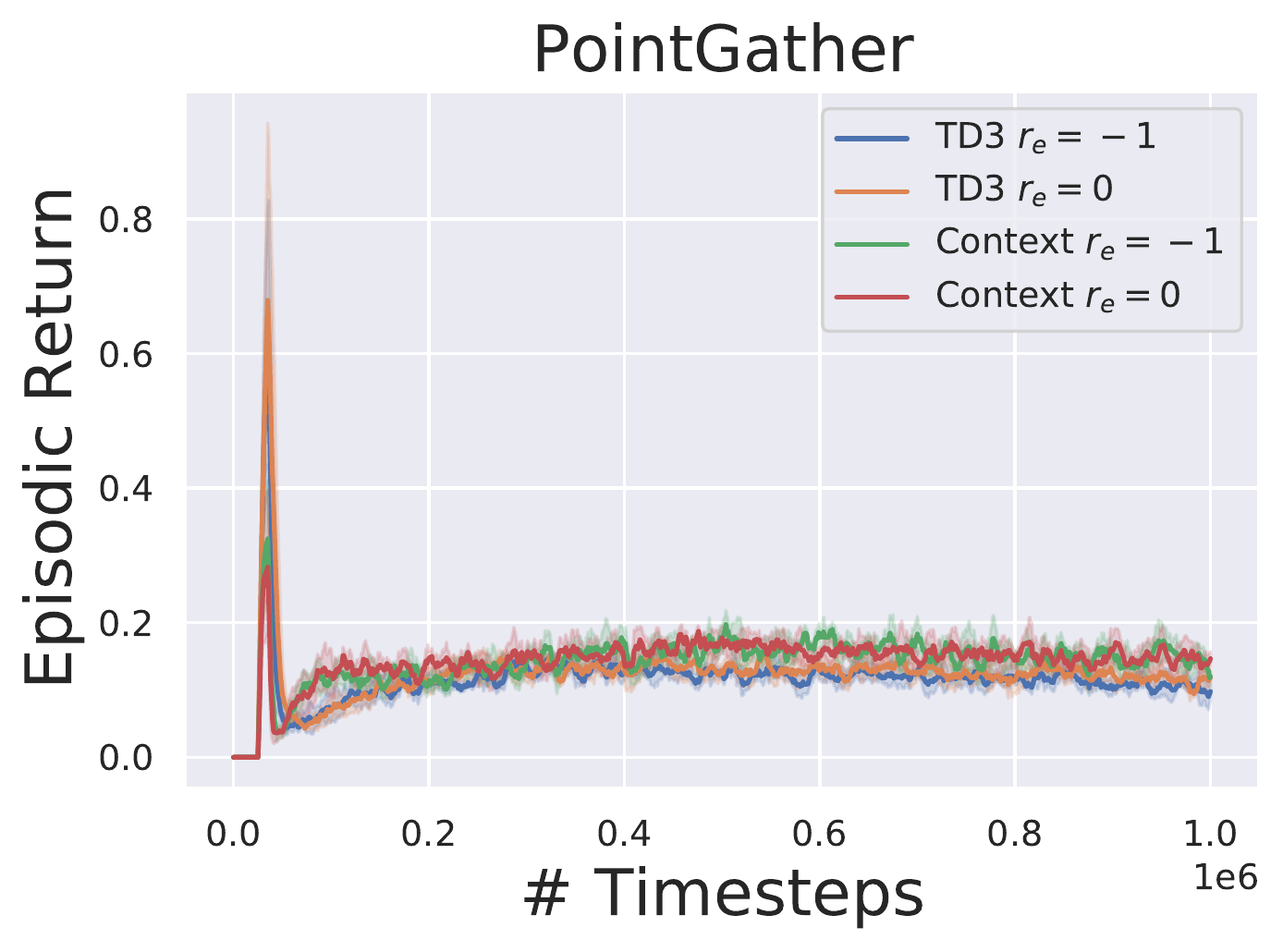}
		\end{minipage}
\caption{Ablation studies on the selection of $r_e$, the value of absorbing reward. The first line shows the episodic return curves of each methods in different environments while the second line shows the corresponding episodic costs. Using smaller $r_e$ will lead to more conservative behavior, i.e., slightly lower return and lower cost.}
\label{fig_r_e}
\end{center}
\vskip -0.2in
\end{figure}

We show experimental results on the selection of different value of the ending reward $r_e$ in this section. Figure~\ref{fig_r_e} shows the results on the CarGoal, PointGoal and PointGather environments. In both TD3 and Context TD3 working in ETMDP, smaller $r_e$'s result in more conservative policies that achieve lower cost and lower primal task reward.

\subsection{More Environments}
\subsubsection{Other Benchmarks}
\begin{figure}[h!]
\vskip 0.2in
\begin{center}
\begin{minipage}[htbp]{0.25\linewidth}
			\centering
			\includegraphics[width=1.0\linewidth]{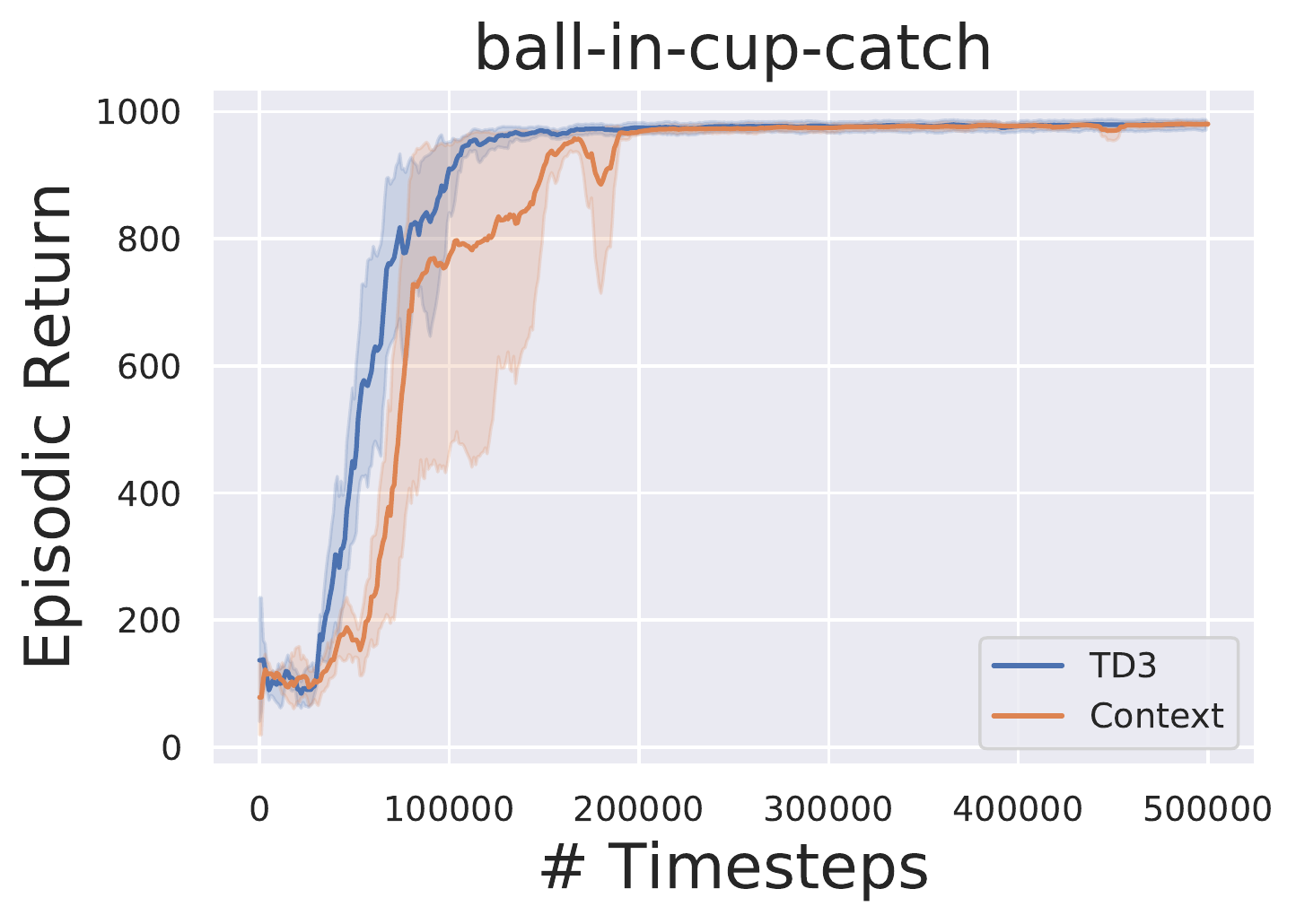}
		\end{minipage}%
		\begin{minipage}[htbp]{0.25\linewidth}
			\centering
			\includegraphics[width=1.0\linewidth]{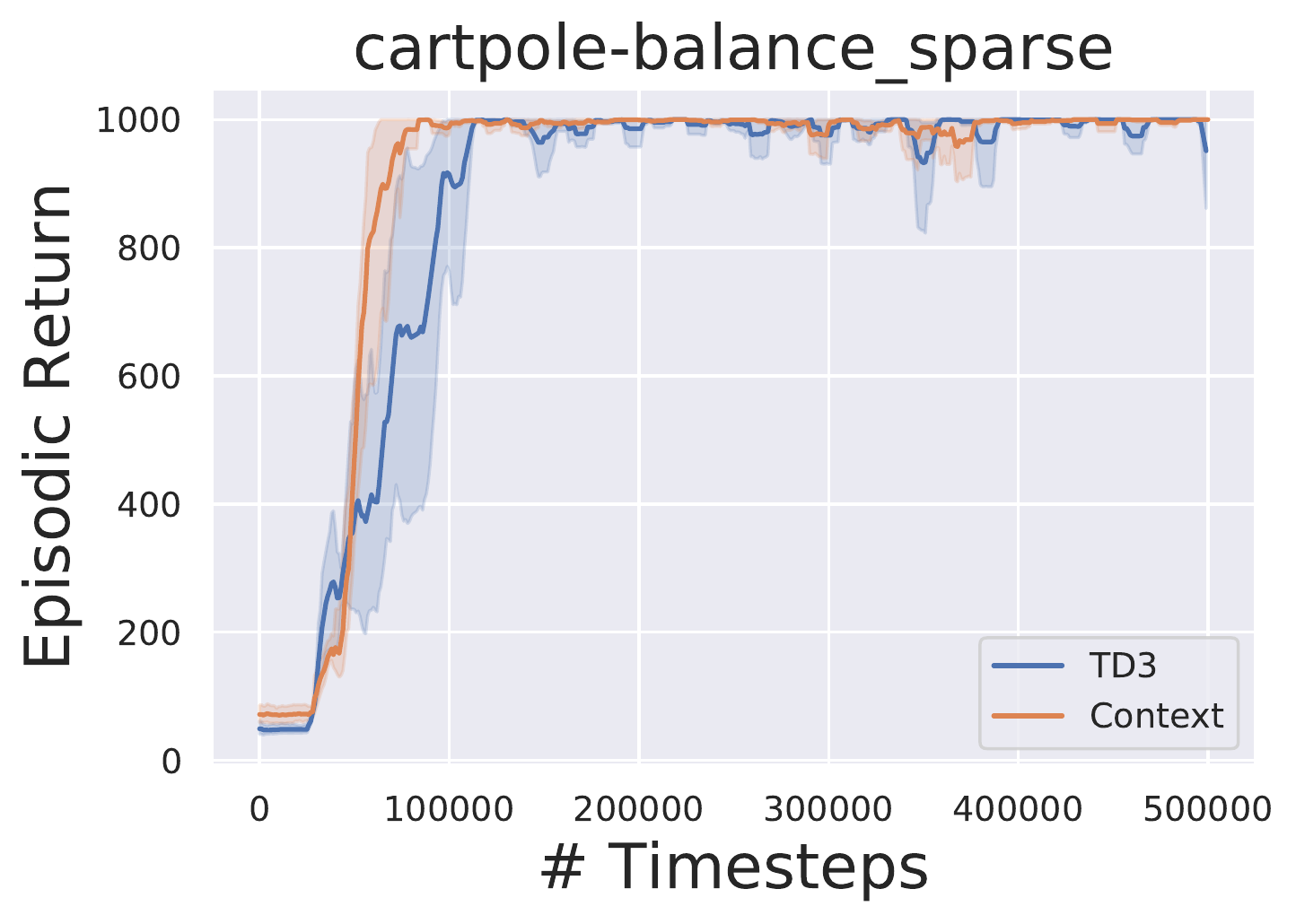}
		\end{minipage}%
		\begin{minipage}[htbp]{0.25\linewidth}
			\centering
			\includegraphics[width=1.0\linewidth]{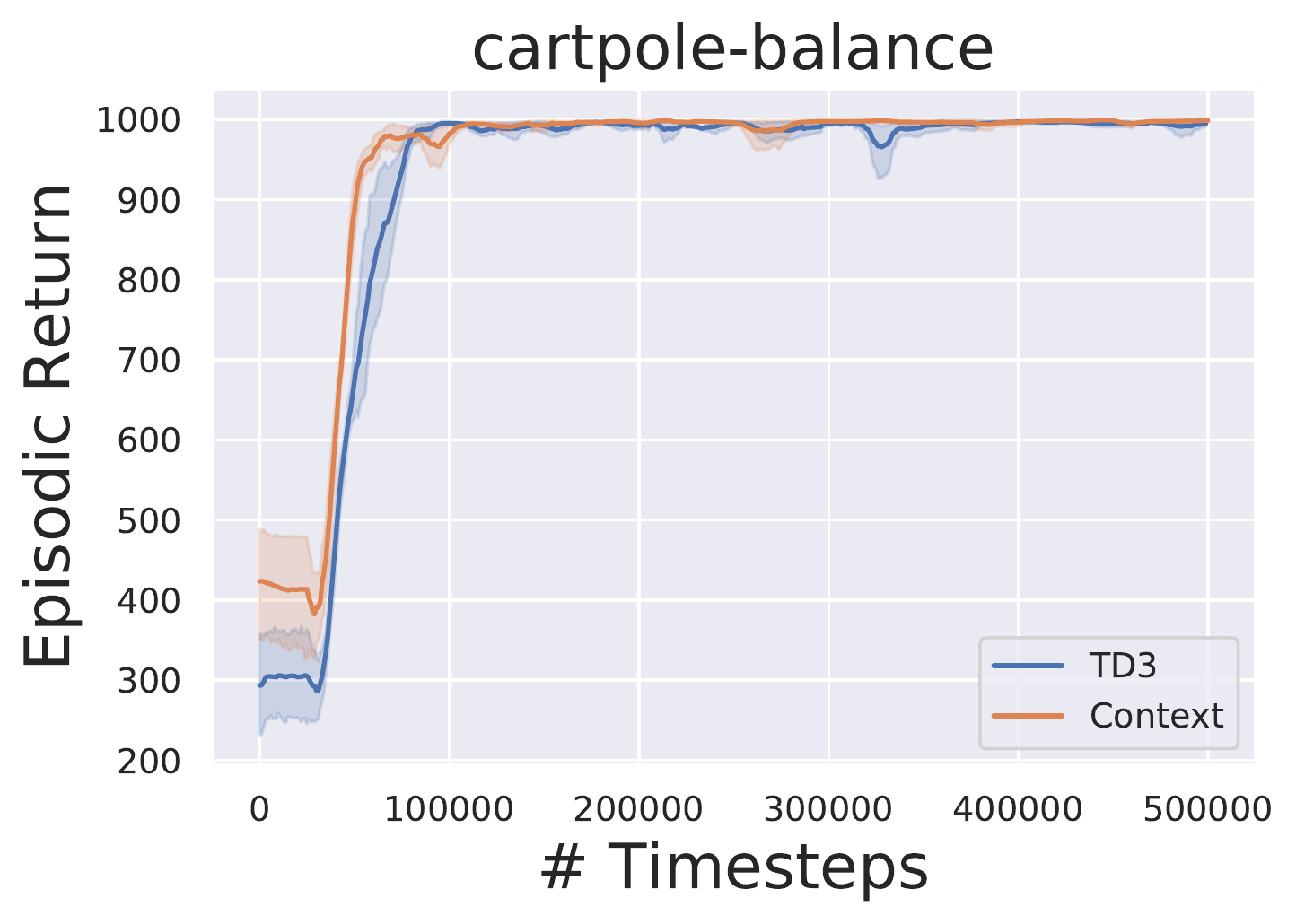}
		\end{minipage}%
		\begin{minipage}[htbp]{0.25\linewidth}
			\centering
			\includegraphics[width=1.0\linewidth]{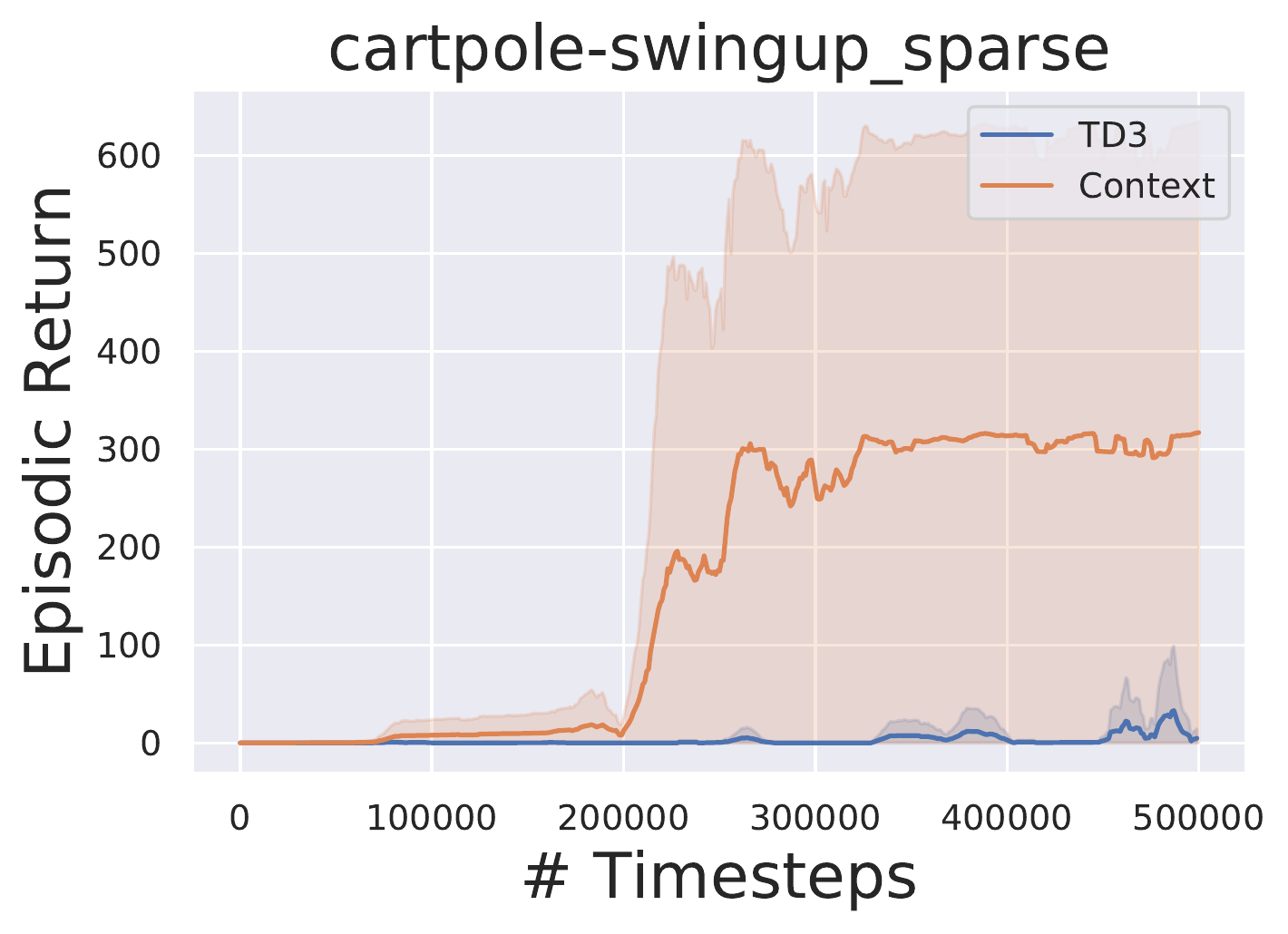}
		\end{minipage}\\ 
		\begin{minipage}[htbp]{0.25\linewidth}
		\centering
			\includegraphics[width=1.0\linewidth]{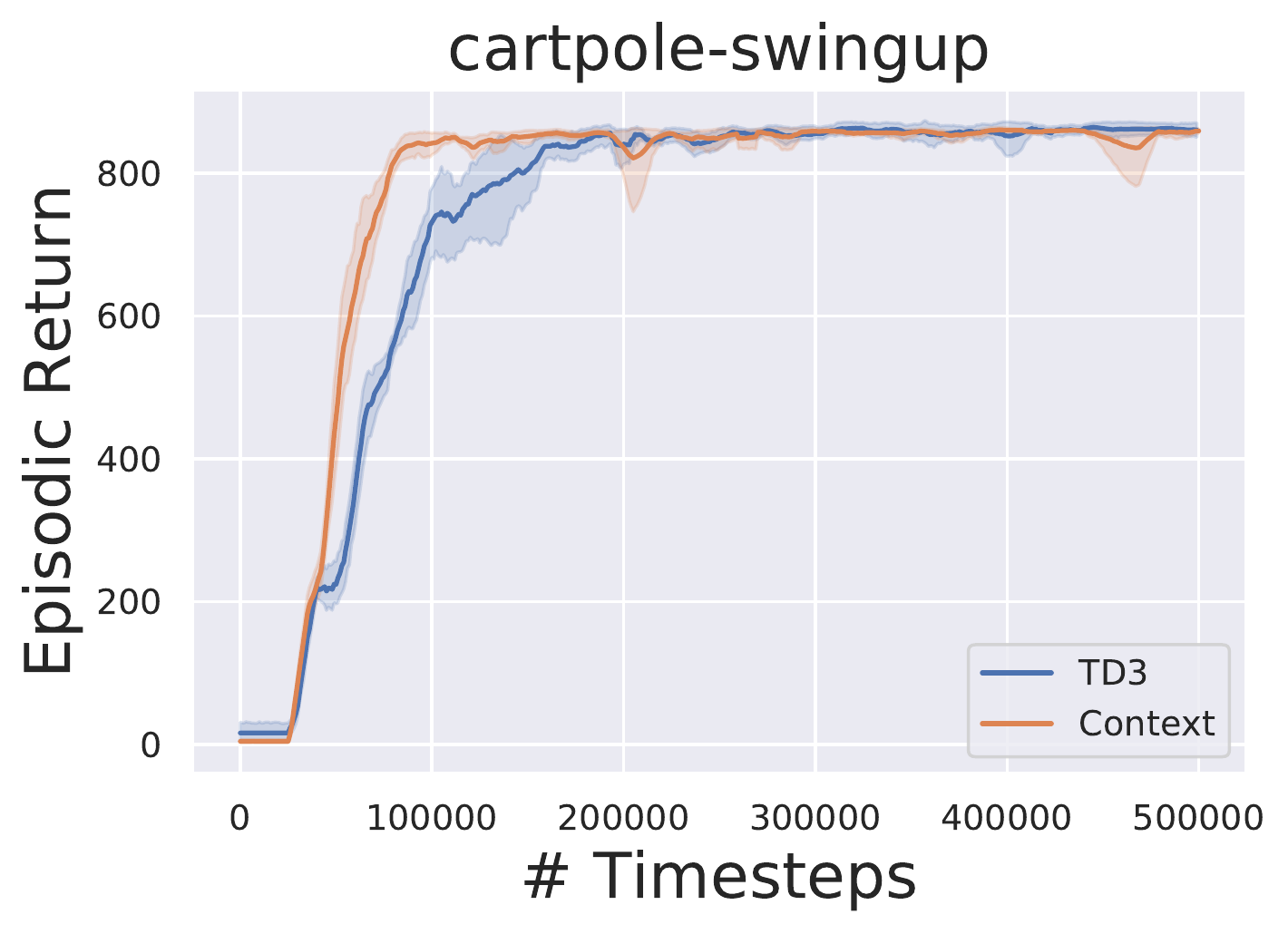}
		\end{minipage}%
		\begin{minipage}[htbp]{0.25\linewidth}
			\centering
			\includegraphics[width=1.0\linewidth]{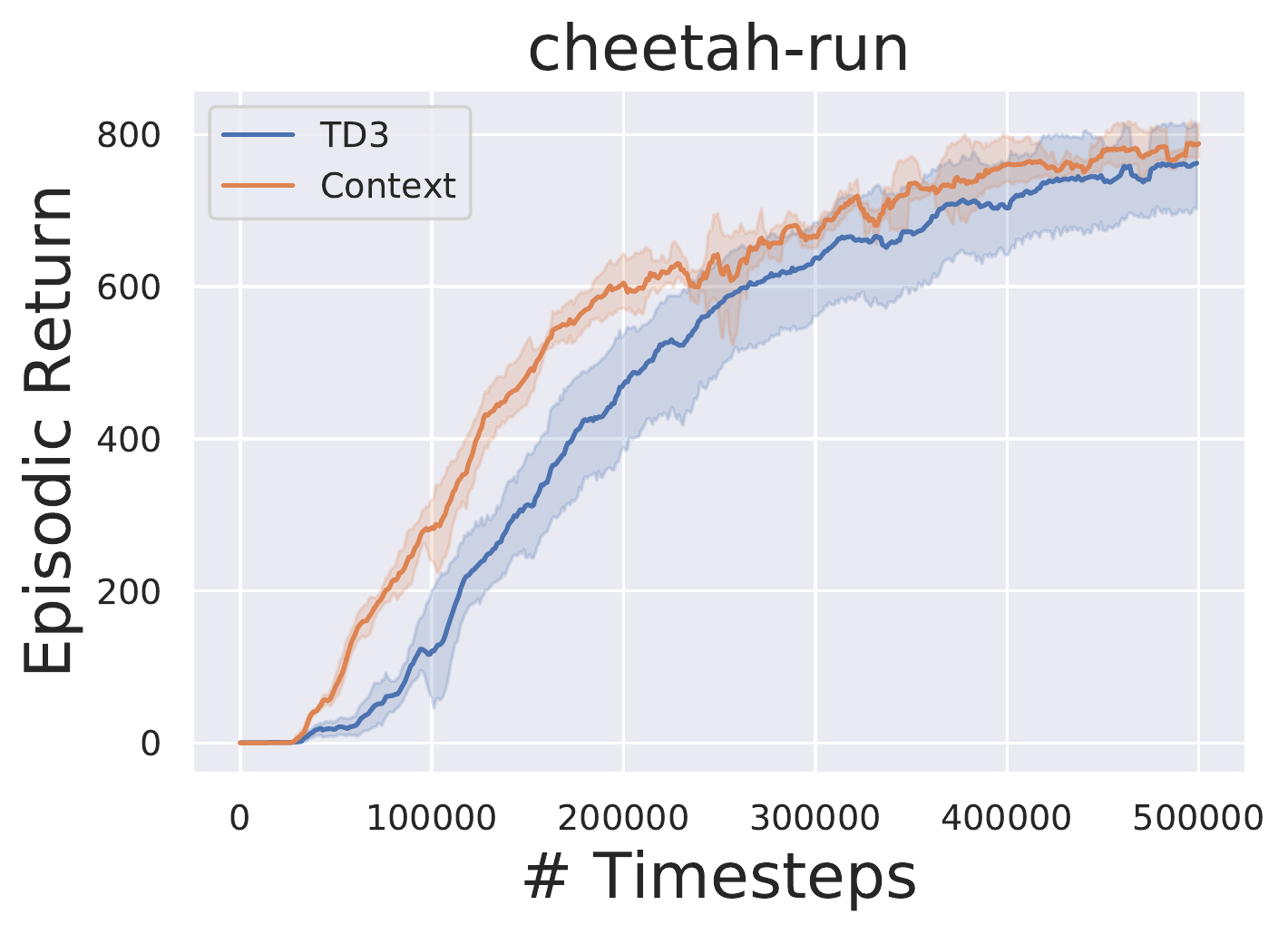}
		\end{minipage}%
		\begin{minipage}[htbp]{0.25\linewidth}
			\centering
			\includegraphics[width=1.0\linewidth]{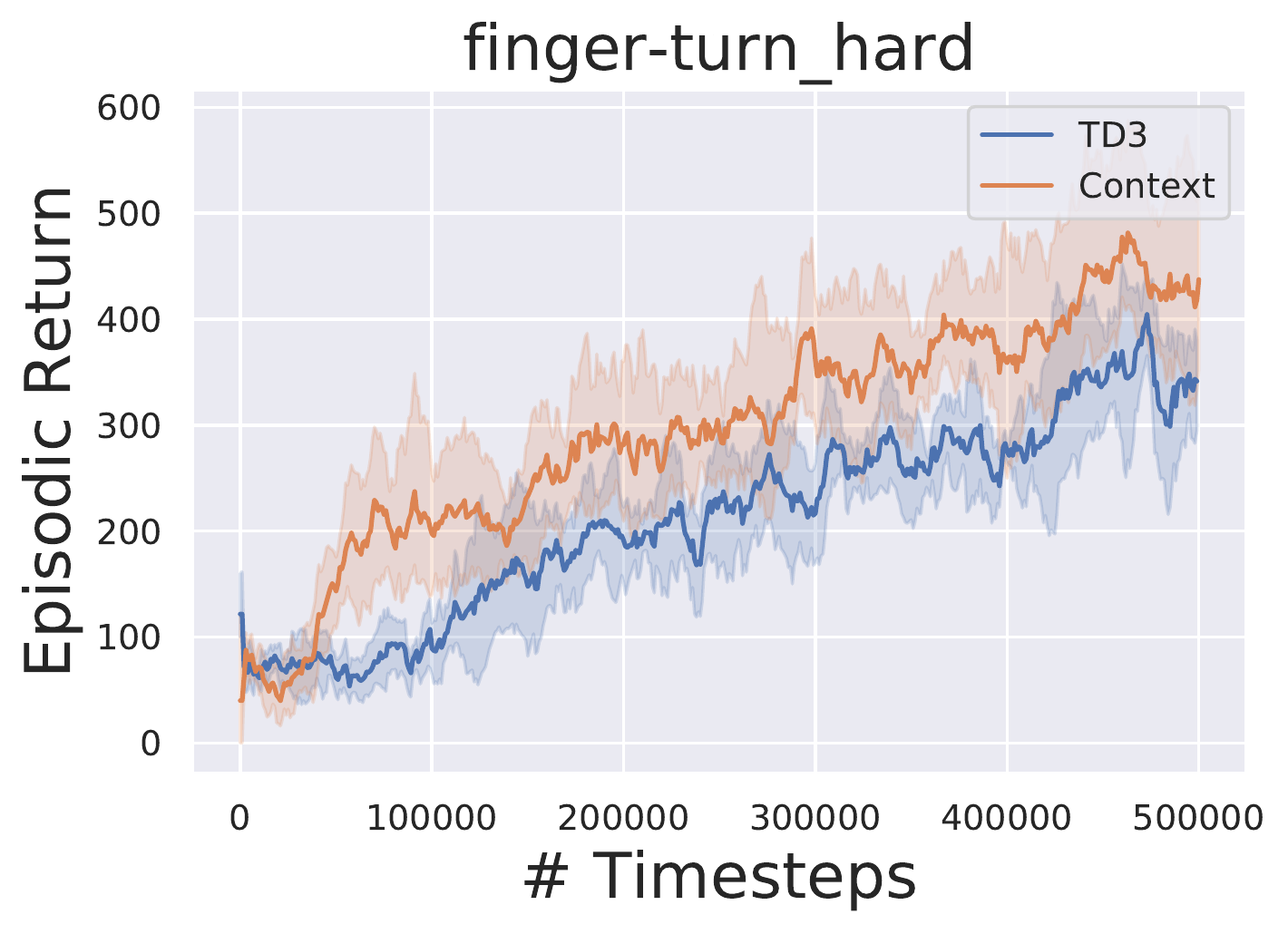}
		\end{minipage}%
		\begin{minipage}[htbp]{0.25\linewidth}
			\centering
			\includegraphics[width=1.0\linewidth]{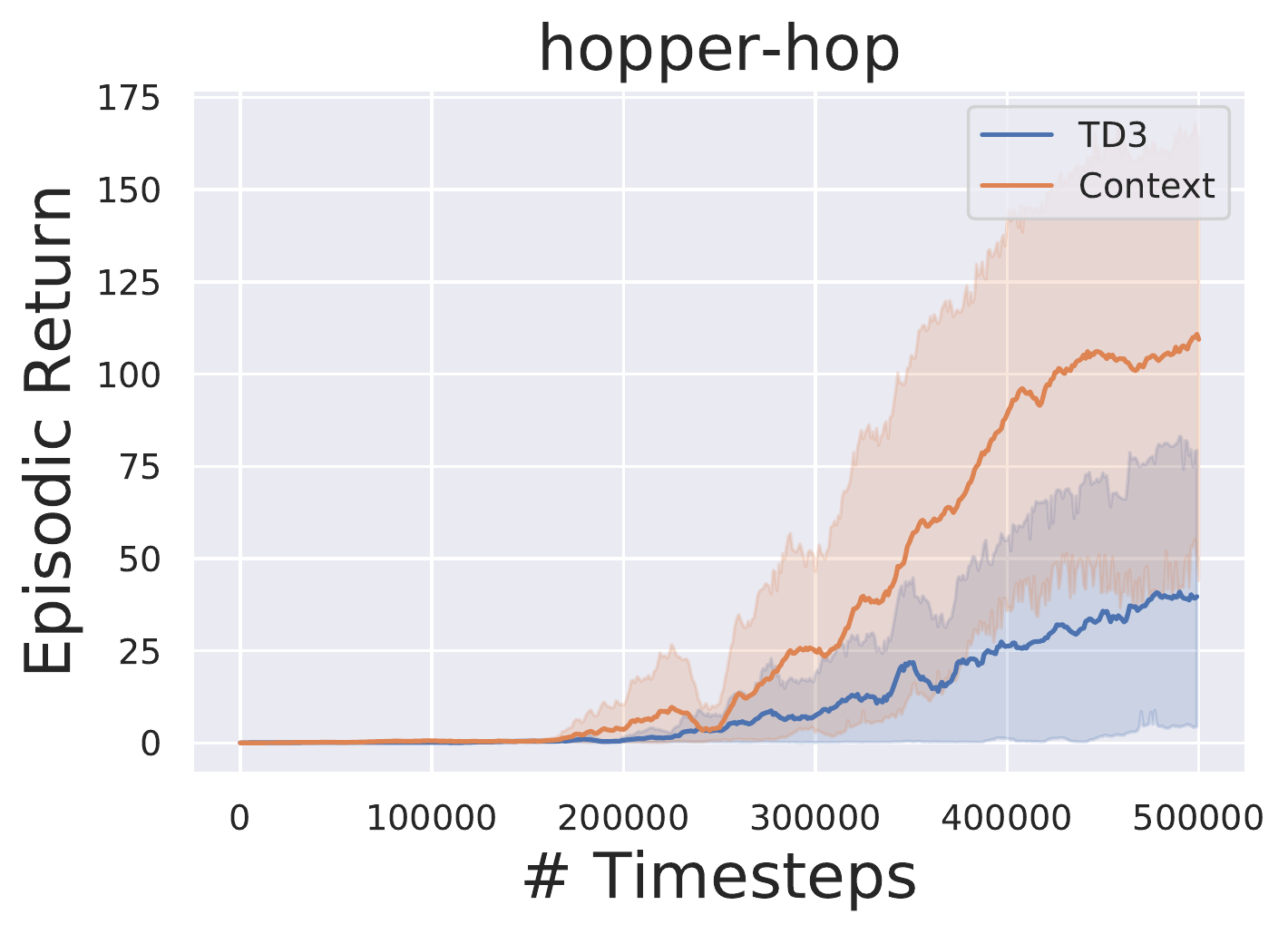}
		\end{minipage} \\ 
		\begin{minipage}[htbp]{0.25\linewidth}
		\centering
			\includegraphics[width=1.0\linewidth]{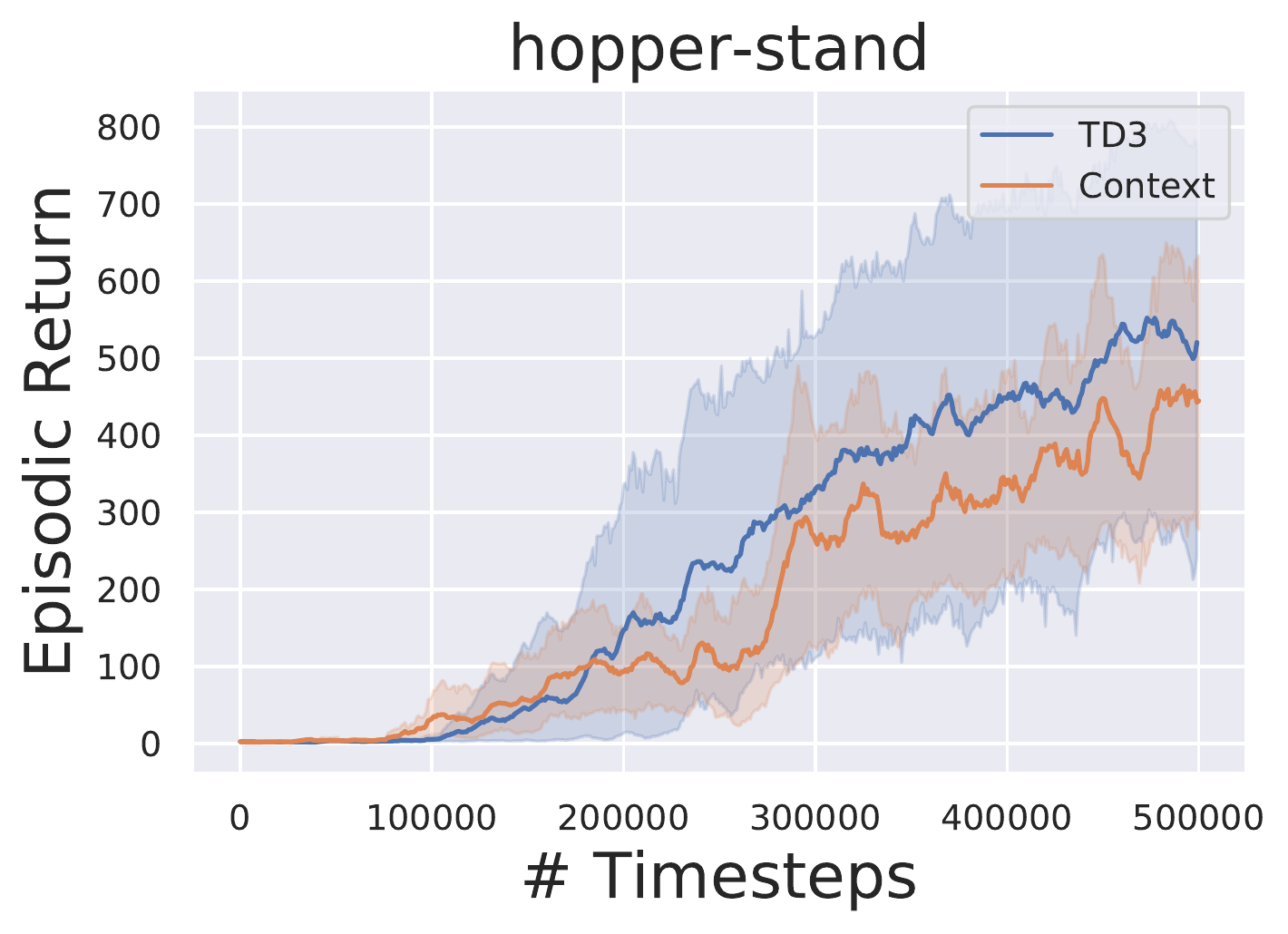}
		\end{minipage}%
		\begin{minipage}[htbp]{0.25\linewidth}
			\centering
			\includegraphics[width=1.0\linewidth]{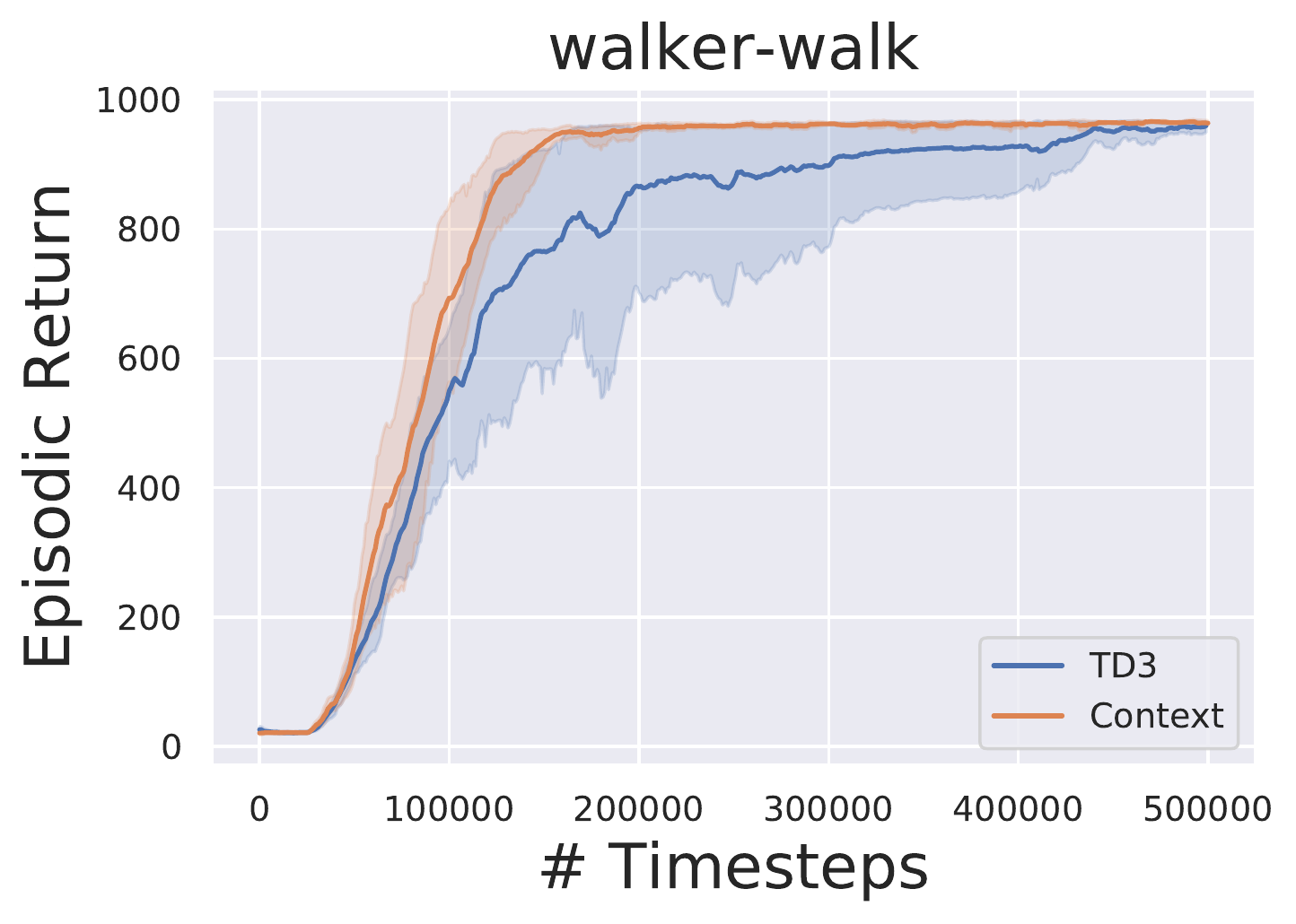}
		\end{minipage}%
		\begin{minipage}[htbp]{0.25\linewidth}
			\centering
			\includegraphics[width=1.0\linewidth]{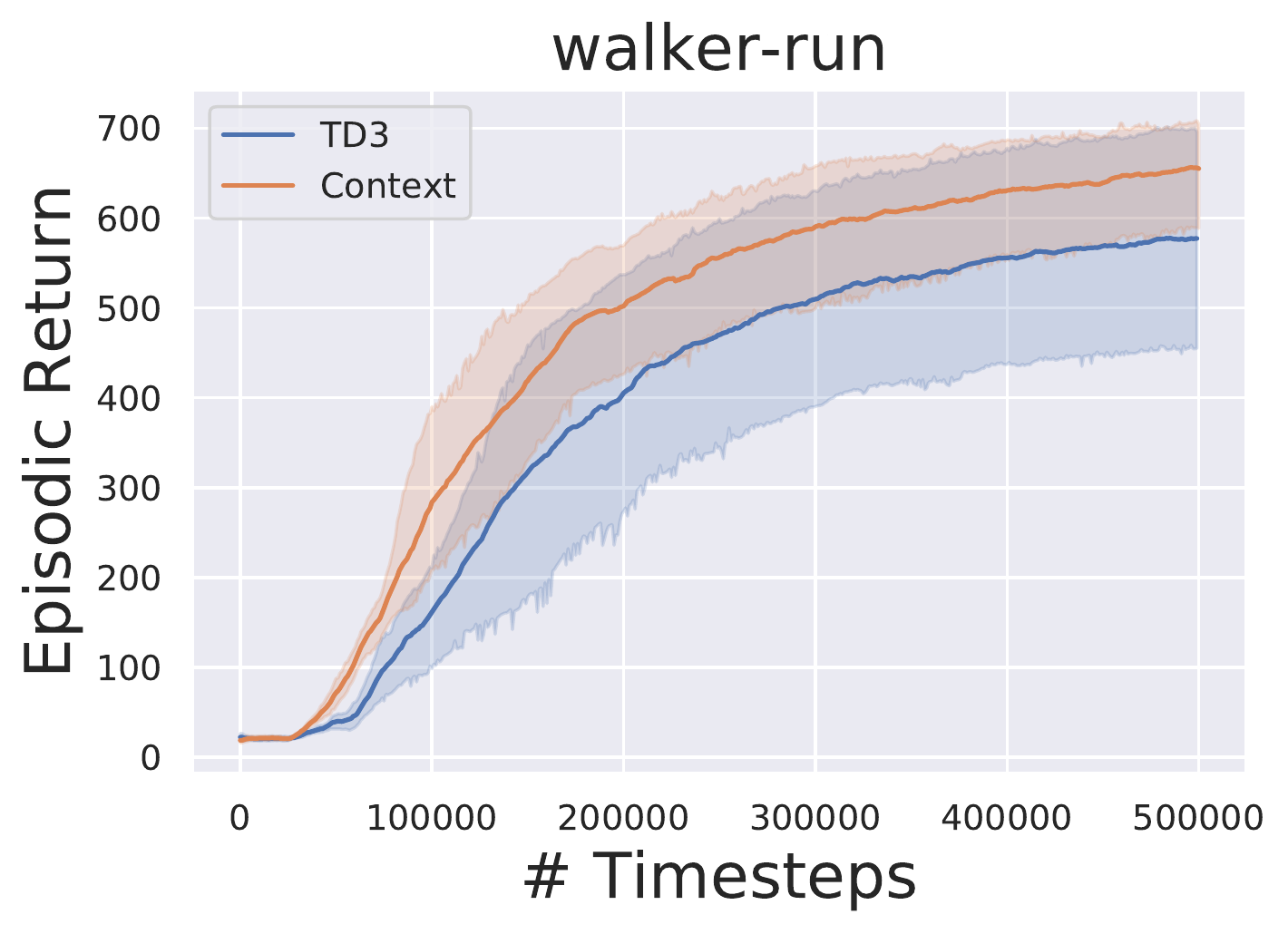}
		\end{minipage}%
		\begin{minipage}[htbp]{0.25\linewidth}
			\centering
			\includegraphics[width=1.0\linewidth]{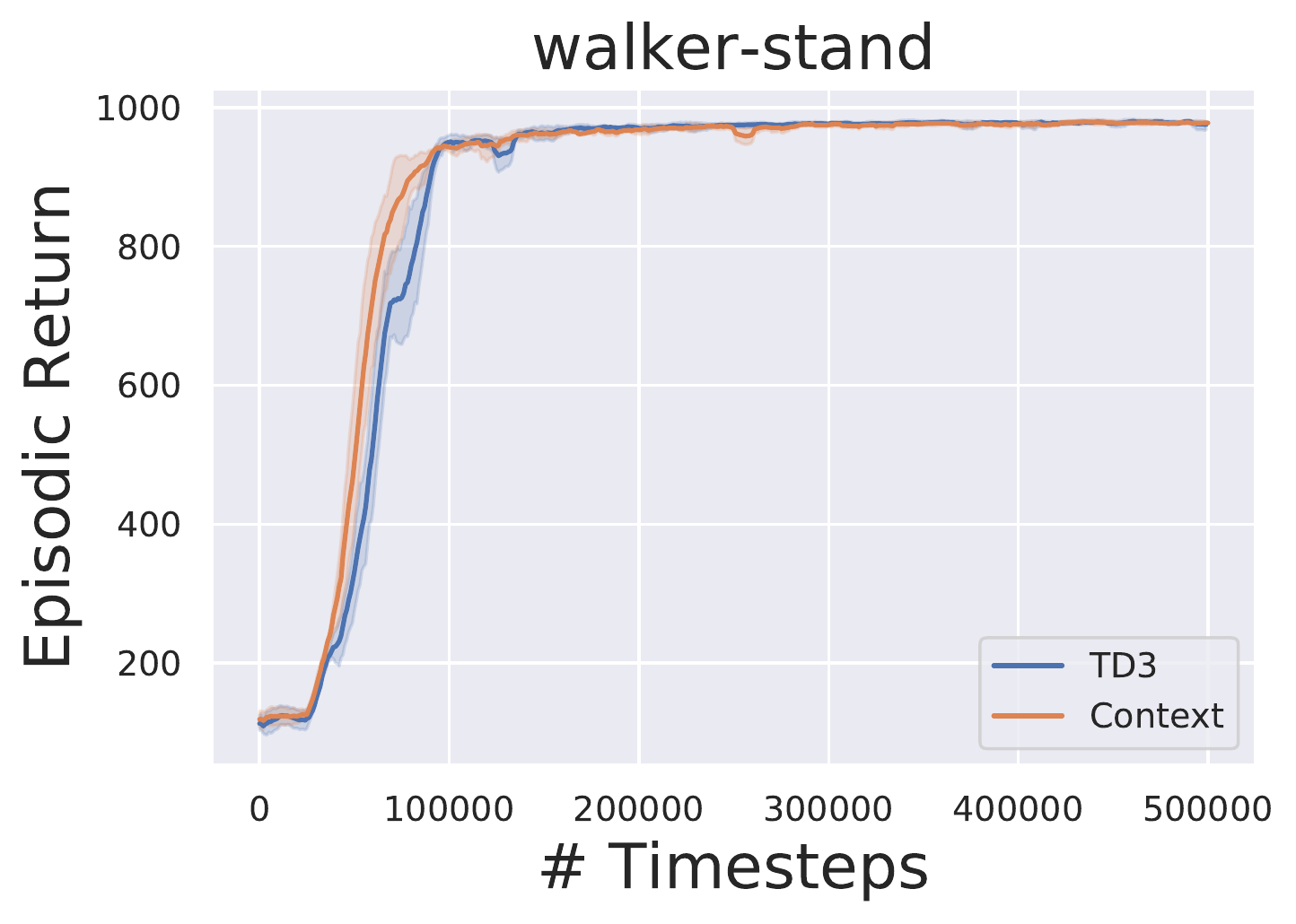}
		\end{minipage}\\ 
		\begin{minipage}[htbp]{0.25\linewidth}
			\centering
			\includegraphics[width=1.0\linewidth]{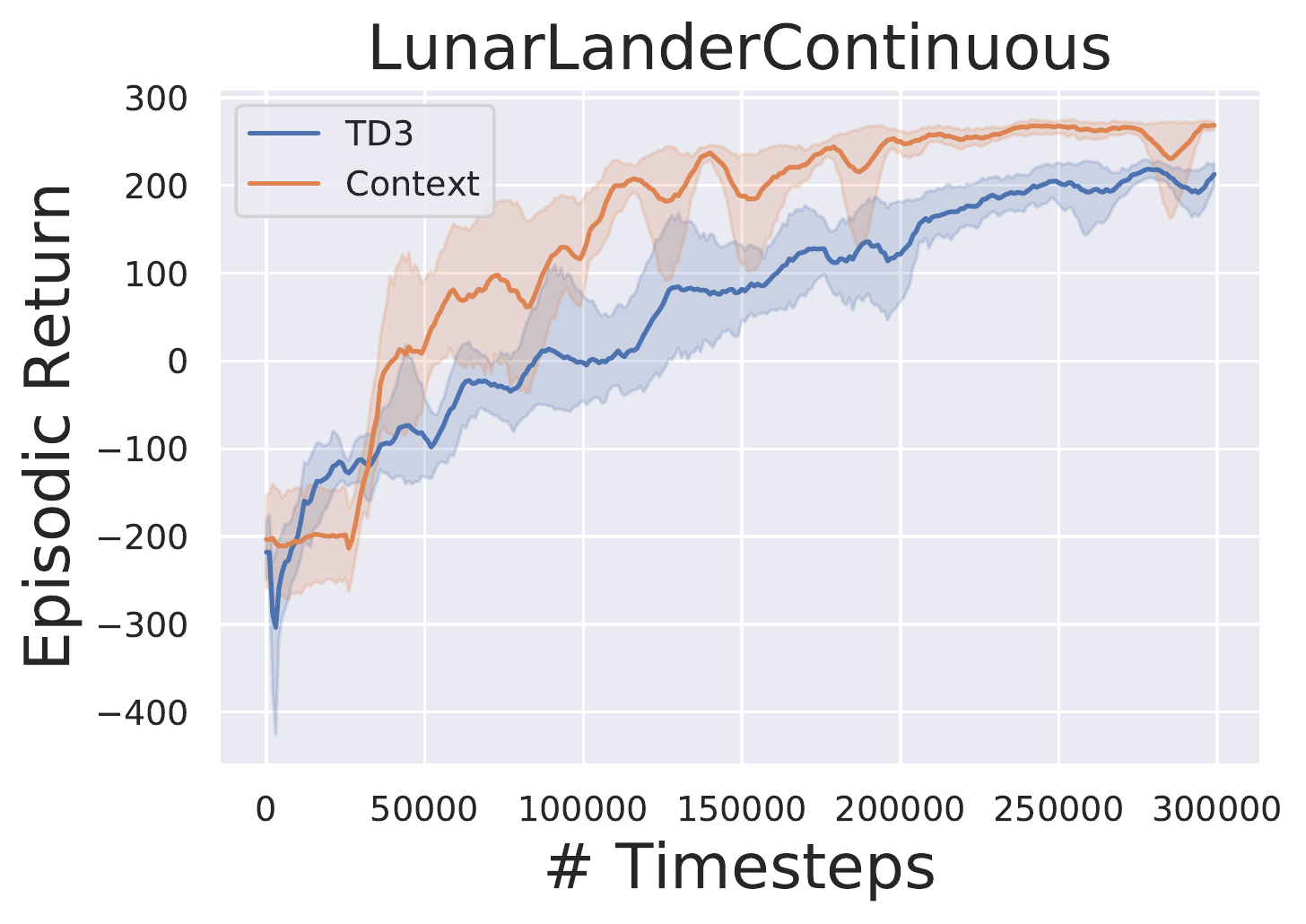}
		\end{minipage}%
		\begin{minipage}[htbp]{0.25\linewidth}
			\centering
			\includegraphics[width=1.0\linewidth]{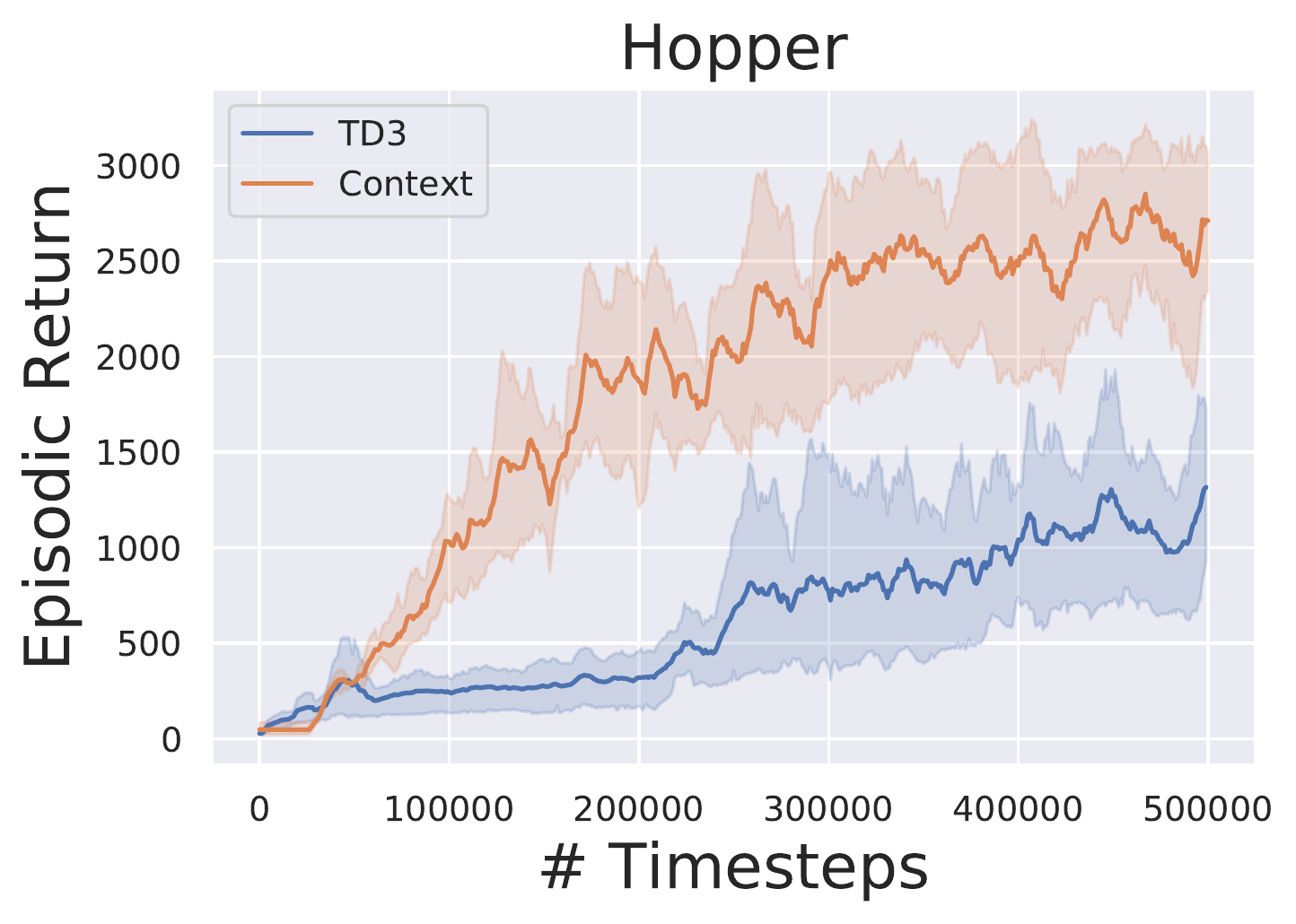}
		\end{minipage}%
		\begin{minipage}[htbp]{0.25\linewidth}
			\centering
			\includegraphics[width=1.0\linewidth]{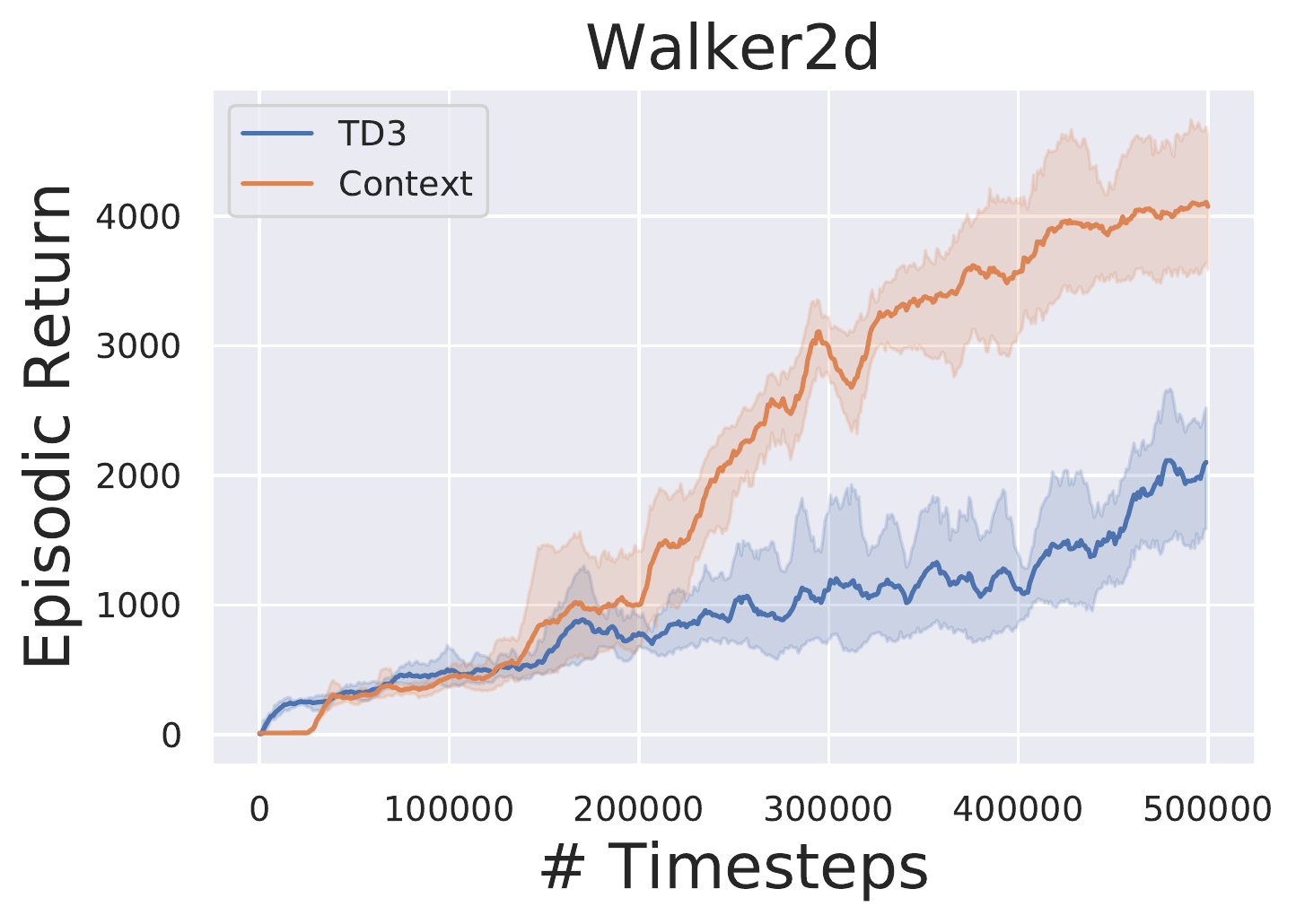}
		\end{minipage}%
		\begin{minipage}[htbp]{0.25\linewidth}
			\centering
			\includegraphics[width=1.0\linewidth]{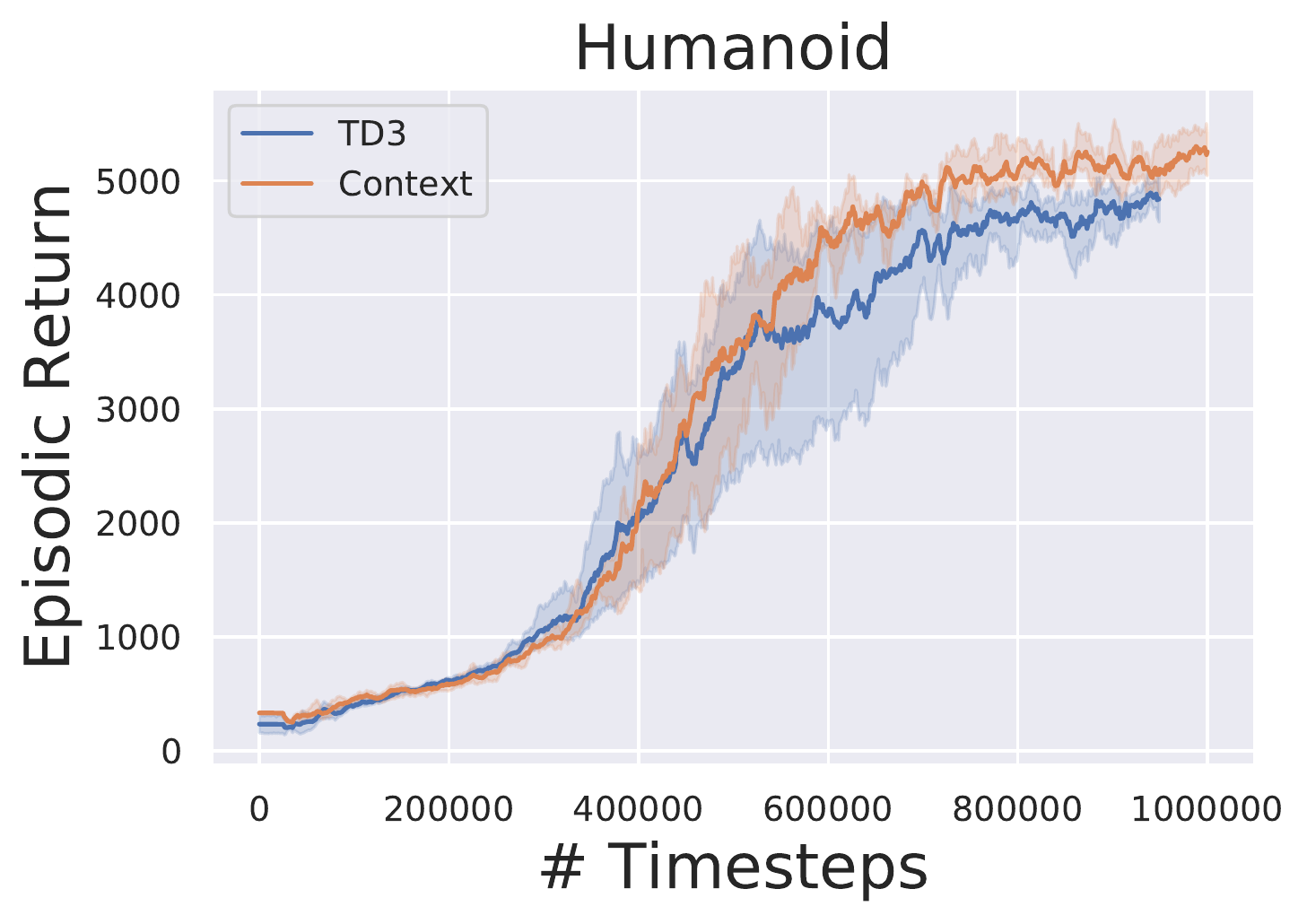}
		\end{minipage}%
\caption{Experiment results on the DeepMind Control Suite. In all 16 benchmark environments we experimented on, context models outperforms normal TD3 in most environments (13 out of 16), showing the superiority of context models in improving learning efficiency.}
\label{fig_more_exps}
\end{center}
\vskip -0.2in
\end{figure}

In this section we show our experiments on various MuJoCo and DeepMind Control benchmarks to show the superiority of the Context TD3 over the vanilla TD3 in sample-efficient learning in MDP tasks. Figure~\ref{fig_more_exps} shows the experiment results. In most environments, Context TD3 achieves better asymptotic performance while being able to converge faster. We use the same hyper-parameter of historical horizon $= 7$ in all experiments. Elaborated searching for hyper-parameters may result in even better performance.

\newpage
\subsection{Model Structure}
\label{appd_sep_struc}

\subsubsection{GRU v.s. Transformer}
\begin{figure}[h!]
\vskip 0.2in
\begin{center}
\begin{minipage}[htbp]{0.33\linewidth}
			\centering
			\includegraphics[width=1.0\linewidth]{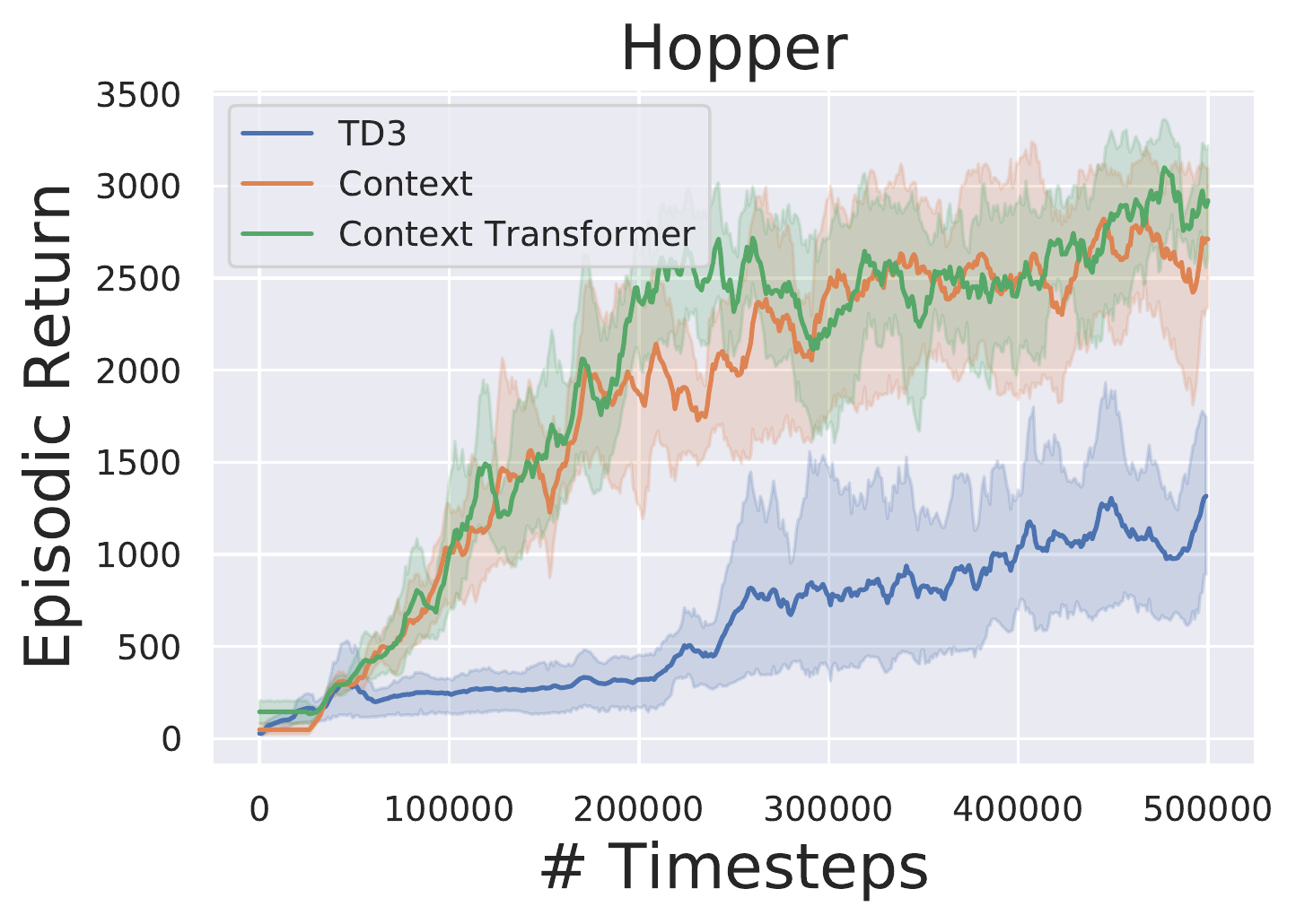}
		\end{minipage}%
		\begin{minipage}[htbp]{0.33\linewidth}
			\centering
			\includegraphics[width=1.0\linewidth]{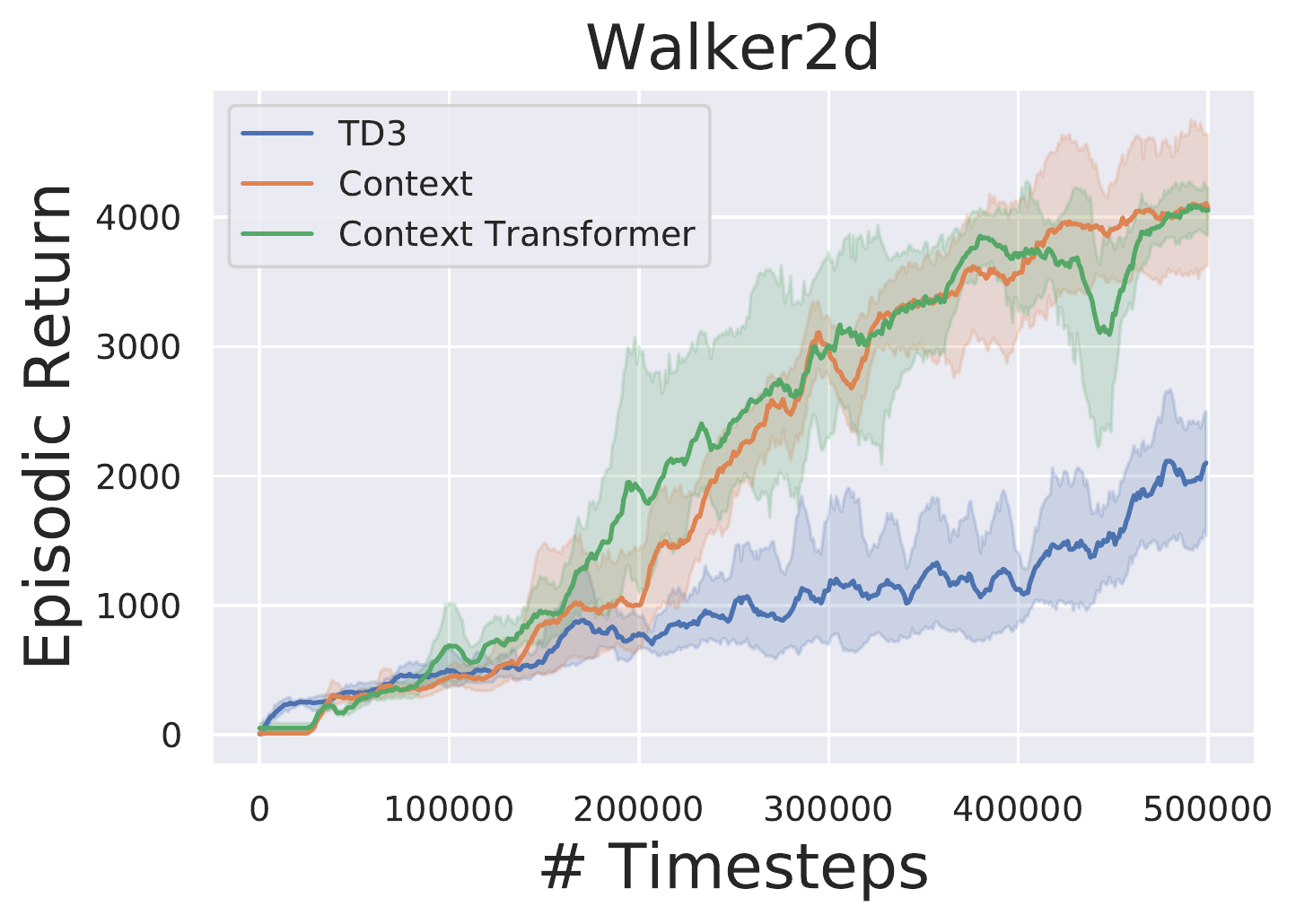}
		\end{minipage}%
\caption{Ablation studies on the selection of context models.}
\label{fig_tf}
\end{center}
\vskip -0.2in
\end{figure}

In this section we provide ablation studies on the choice of context models: we compare the results of Context models based on GRUs and based on recent advances of self-attention based models~\cite{vaswani2017attention}. The results are shown in Figure~\ref{fig_tf}, where we find leveraging the transformer models can not result in better performance.

\subsubsection{Shared v.s. Separated Context Variables}
In the work of \cite{fakoor2019meta}, the context model is trained only through the learning of critic networks. Differently, in our experiments we find training context models separately for the actor and critic can result in better performance. \textbf{Context Shared} in Figure~\ref{fig_hidden_units} denotes the results when the context model is shared by actor and critic as recommended in the Meta-RL literature ~\citep{fakoor2019meta}.

\section{Detailed Learning Curves}

\begin{figure}[h!]
\vskip 0.2in
\begin{center}
\begin{minipage}[htbp]{0.245\linewidth}
			\centering
			\includegraphics[width=1.0\linewidth]{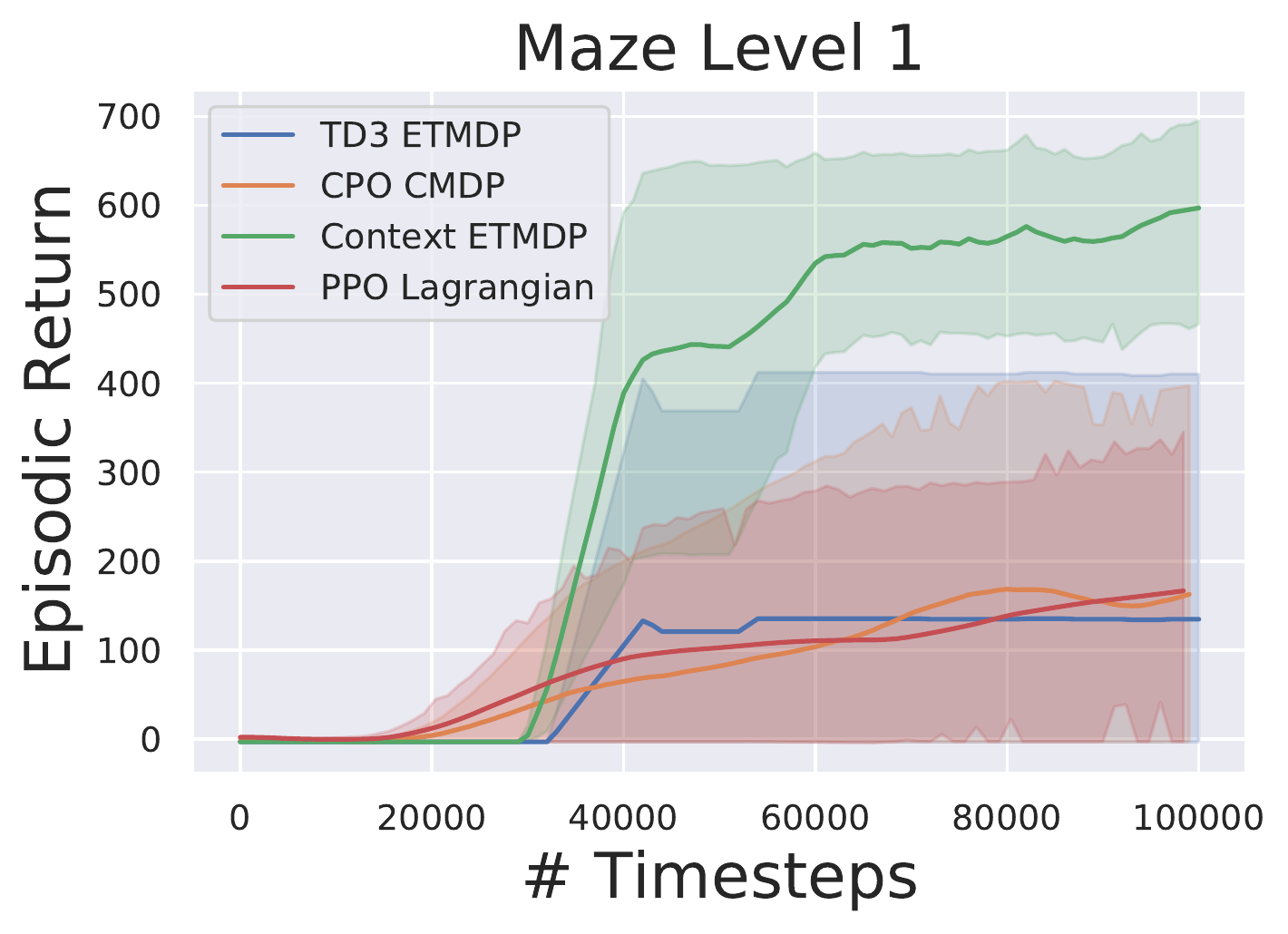}
		\end{minipage}%
		\begin{minipage}[htbp]{0.245\linewidth}
			\centering
			\includegraphics[width=1.0\linewidth]{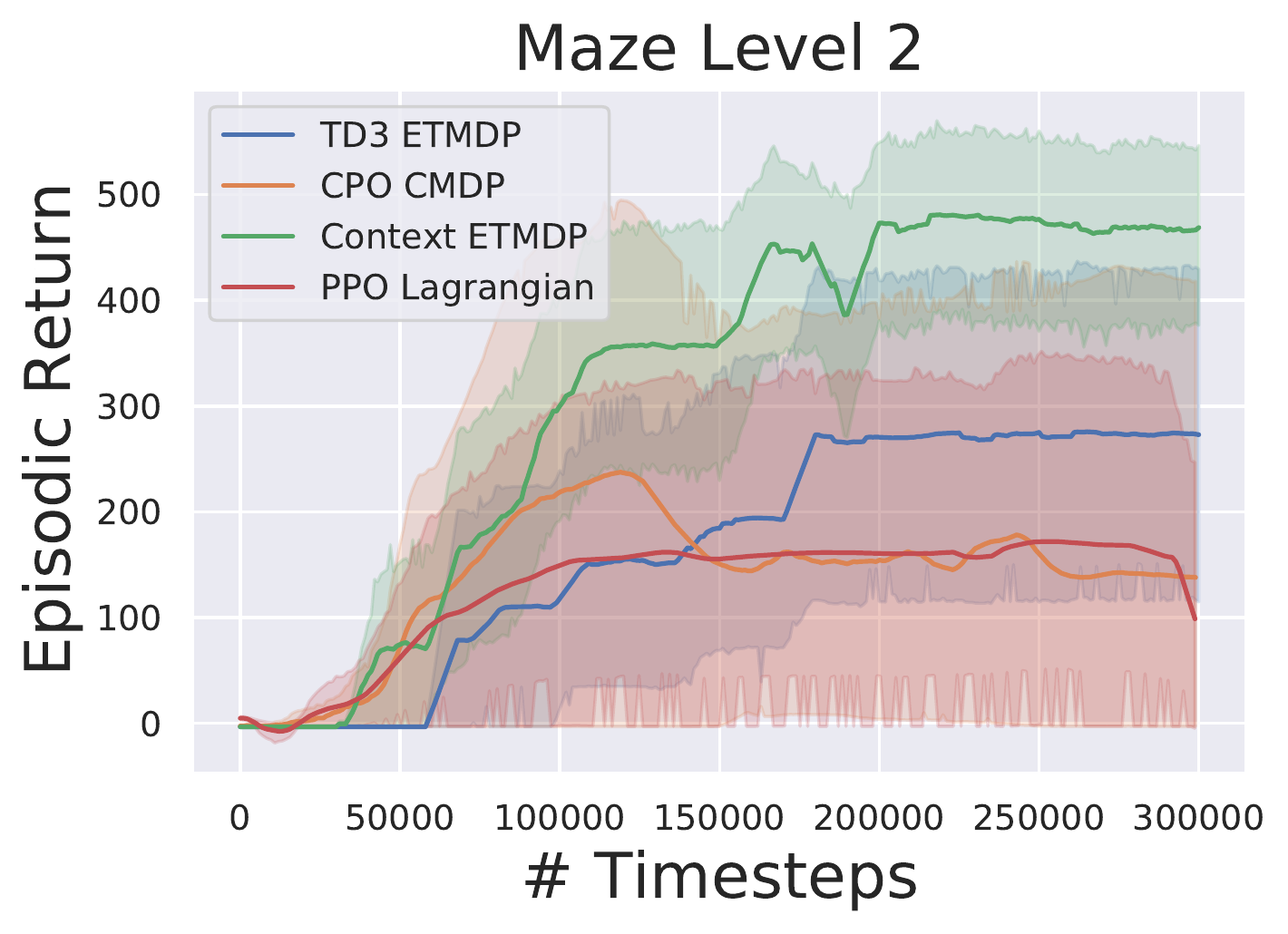}
		\end{minipage}%
		\begin{minipage}[htbp]{0.245\linewidth}
			\centering
			\includegraphics[width=1.0\linewidth]{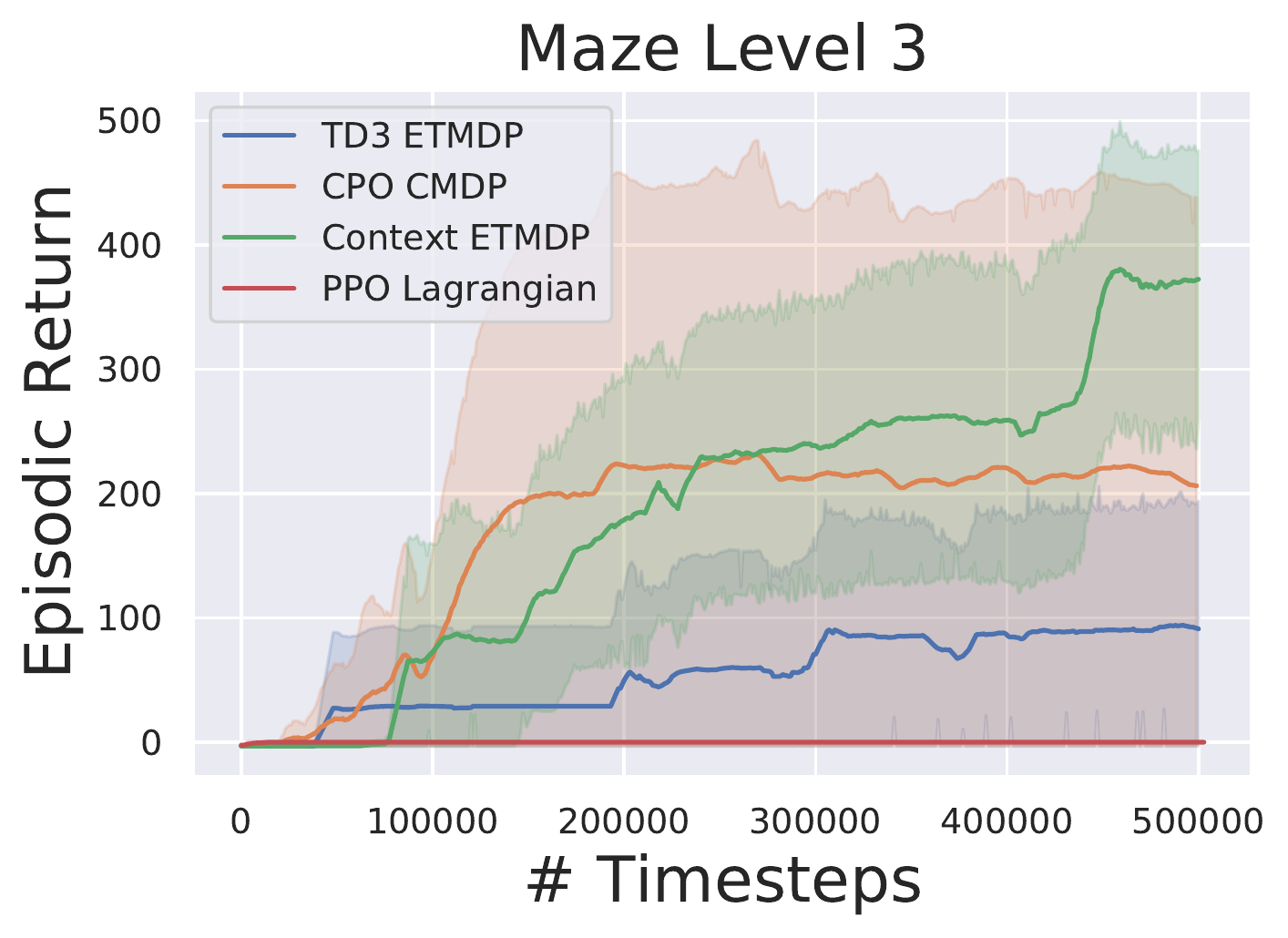}
		\end{minipage}%
		\begin{minipage}[htbp]{0.245\linewidth}
			\centering
			\includegraphics[width=1.0\linewidth]{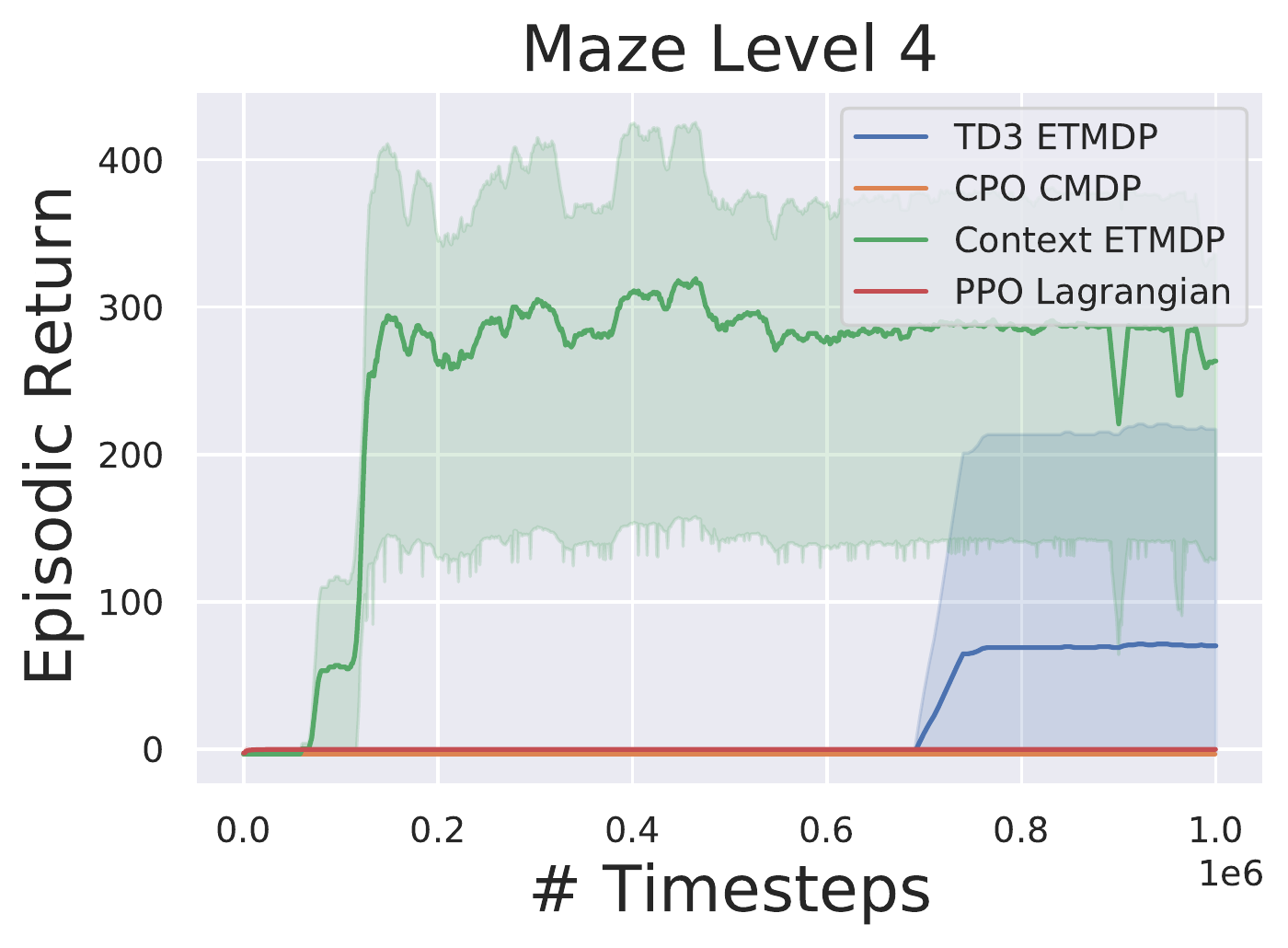}
		\end{minipage}%
\caption{Learning curves of the Maze environments. As the constraints are binary, any reward gained when the constraints are violated is not taken into consideration. Therefore only episodic return curves are shown in the figures. i.e., All of those rewards are gained \textit{without breaking the constraints}.}
\label{fig_main_maze}
\end{center}
\vskip -0.2in
\end{figure}

\section{Counterexample where Tightened Approximation Will Fail}
\label{counterexample}
\begin{figure}[t]
\vskip 0.2in
\begin{center}
\begin{minipage}[htbp]{0.33\linewidth}
			\centering
			\includegraphics[width=1.0\linewidth]{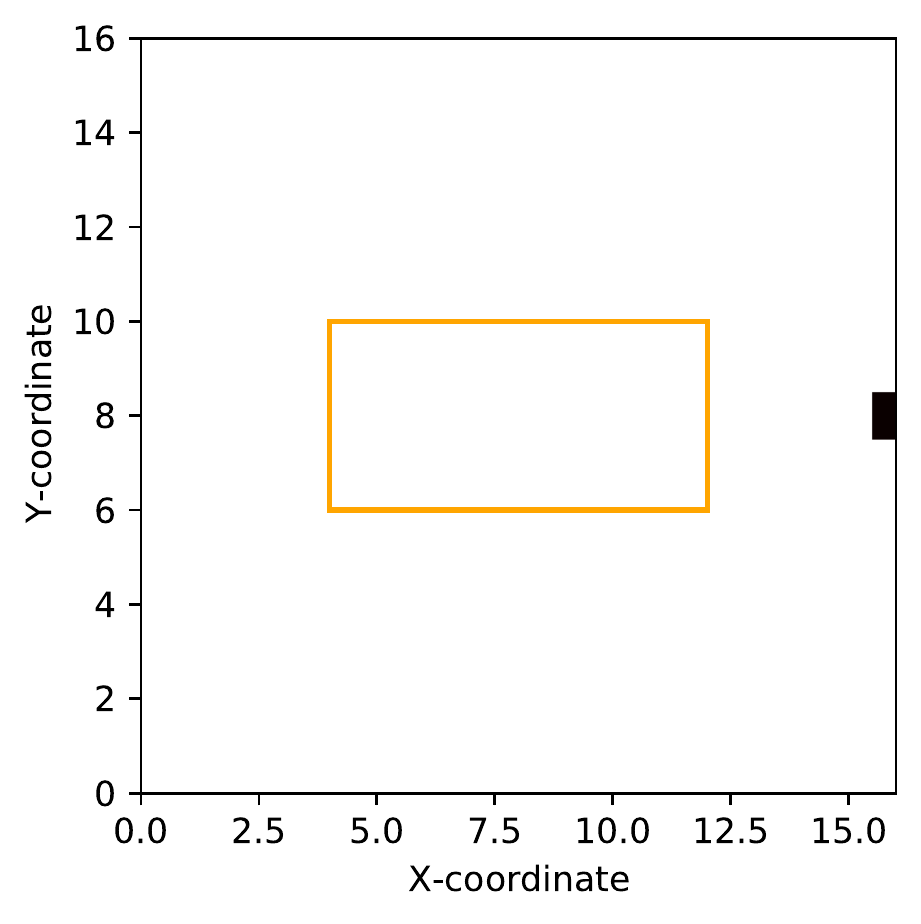}
		\end{minipage}%
\caption{Counterexample where the tightened approximation will fail.}
\label{fig_counter}
\end{center}
\vskip -0.2in
\end{figure}

Although in our work we find using the tightened approximation in CMDP tasks with budget constraint is able to achieve satisfactory performance, here are counterexamples that applying such an approximation in ETMDP can fail to solve the corresponding CMDP. Such a counterexample is not hard to construct, though we may not meet such a situation in practice, we expose the possibility in this section for future exploration. The basic idea is: when the task must be solved with the budget consumed partially, the approximation will fail to find a proper solution. In the example shown in Figure~\ref{fig_counter}, if the agent is initialized at center of the map, and the cost budget is $1$, permitting the agent to cross the constrained region for once, the tightened approximation will not lead to good policy as the agent can never learn how to behave outside the rectangle region. For those problems, the primal cost-aware design introduced in Section~\ref{sec_prac_issues} must be used. 

\newpage
\section{Qualitative Results}
We also include demo videos in the supplemental materials. Where the performance of agents trained with different algorithms in the PointGoal1-v0 and CarGoal1-v0 safe-navigation tasks are shown qualitatively.